\documentclass[10pt,twocolumn,letterpaper]{article}

\usepackage[pagenumbers]{cvpr}

\usepackage{lmodern}
\usepackage{relsize}
\usepackage{tabularray}
\usepackage{graphicx}
\usepackage{mwe}
\usepackage[normalem]{ulem}
\usepackage{tabularray}
\usepackage{graphicx}
\usepackage{array}
\usepackage{adjustbox}
\usepackage{float}
\usepackage{placeins}
\usepackage{pgfplots}
\usepackage{pgfplotstable}
\pgfplotsset{compat=1.18}
\usepgfplotslibrary{fillbetween}
\usetikzlibrary{intersections}

\definecolor{cvprblue}{rgb}{0.21,0.49,0.74}
\usepackage[pagebackref,breaklinks,colorlinks,allcolors=cvprblue]{hyperref}

\def\confYear{2026}
\newcommand{\model}{MAMMA\xspace}
\newcommand{\modellong}{Markerless Accurate Multi-person Motion Acquisition\xspace}
\newcommand{\net}{MammaNet\xspace}

\newcommand{\dataset}{MammaSyn\xspace}
\newcommand{\datasetEval}{MammaEval\xspace}

\newcommand{\supmat}{SupMat\xspace}
\newcommand{\supvid}{Supplementary Video\xspace}

\title{MAMMA: Markerless Accurate Multi-person Motion Acquisition}

\author{
Hanz Cuevas-Velasquez\textsuperscript{1},
Anastasios Yiannakidis\textsuperscript{1},
Soyong Shin\textsuperscript{2},
Giorgio Becherini\textsuperscript{1},\\
Markus Höschle\textsuperscript{1},
Joachim Tesch\textsuperscript{1},
Taylor Obersat\textsuperscript{1},
Tsvetelina Alexiadis\textsuperscript{1},\\
Eni Halilaj\textsuperscript{2},
Michael J.~Black\textsuperscript{1}
\\[3pt]
\textsuperscript{1}Max Planck Institute for Intelligent Systems, Germany
\textsuperscript{2}Carnegie Mellon University, USA
}

\begin{document}
\twocolumn[{%
	\renewcommand\twocolumn[1][]{#1}%
	\maketitle
	\begin{center}
		\newcommand{\teaserwidth}{\textwidth}
		\centerline{
			\includegraphics[width=0.85\teaserwidth,clip]{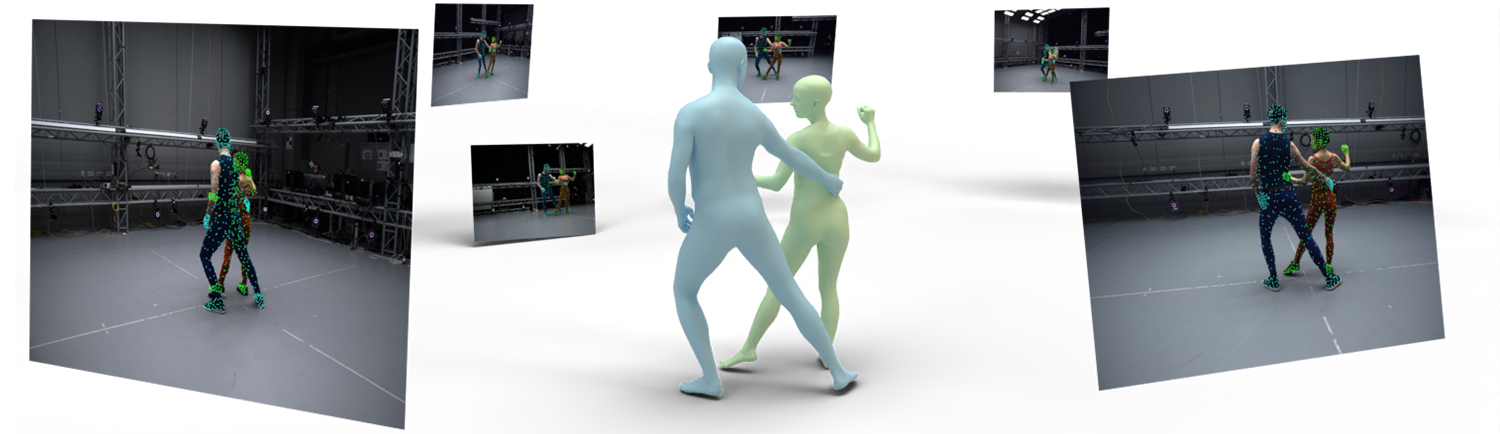}
		}
        \captionof{figure}{From dense surface landmarks to SMPL-X: \model accurately reconstructs pose and shape from synchronized multi-view videos.}
		\vspace{-0.2in}
		\label{fig:teaser}
	\end{center}%
}]

\maketitle
\begin{abstract}
We present \model, a markerless motion-capture pipeline that accurately recovers SMPL-X parameters from multi-view video.
Traditional motion-capture systems rely on physical markers.
Although they offer high accuracy, their requirements of specialized hardware, manual marker placement, and extensive post-processing make them costly and time-consuming.
Recent learning-based methods attempt to overcome these limitations, but most are designed for single-person capture, rely on sparse keypoints, or struggle with occlusions and physical interactions.
In this work, we introduce a method that predicts dense 2D contact-aware and visibility-aware surface landmarks conditioned on segmentation masks, enabling person-specific correspondence estimation even under heavy occlusion.
We employ a novel architecture that exploits learnable queries for each landmark.
We demonstrate that our approach can handle complex person--person interaction and offers greater accuracy than existing methods.
To train our network, we construct a large, synthetic multi-view dataset combining human motions from diverse sources, including extreme poses, hand motions, and close interactions.
Our dataset yields high-variability synthetic sequences with rich body contact and occlusion, and includes SMPL-X ground-truth annotations with dense 2D landmarks.
The result is a system capable of accurately capturing human motion without the need for markers.
Our approach offers competitive reconstruction quality compared to commercial marker-based motion-capture solutions, without the extensive manual cleanup.
Finally, we address the absence of common benchmarks for dense-landmark prediction and markerless motion capture by introducing two evaluation settings built from real multi-view sequences. \url{https://mamma.is.tue.mpg.de/}
\vspace{-.9cm}
\end{abstract}

\section{Introduction}
Traditional marker-based motion-capture (mocap) methods~\citep{vicon, optitrack, motionanalysis, qualisys} are widely used to capture 3D human movement for varied applications.
These systems rely on expensive, specialized hardware and optical markers that are time-consuming to apply.
Most require trained personnel to configure and operate, and involve laborious procedures to clean up noisy results, especially with complex poses and close multi-person interactions due to markers falling off or occlusions.
Close interactions are common in daily life, from hugging to interpersonal activities such as martial arts or dancing. Yet such challenging motions remain underrepresented in existing motion-capture datasets~\citep{amass}.

To address these issues, numerous attempts have been made to replace marker-based systems with markerless ones.
These capture a subject using multiple video cameras and apply various computer vision algorithms to track the 3D-body motion across frames.
While commercial markerless systems exist~\citep{captury,theia,moveai,moverse}, they remain expensive and are generally considered to be less accurate than marker-based systems (the ``gold standard'').

Given the rapid advances in computer vision, we ask whether it is possible to ``roll your own'' markerless system that achieves accuracy comparable to commercial marker-based systems without the tedious manual clean up.
Making such a system available for academic research would democratize motion capture and increase the scale at which complex motion could be captured.

Here we develop such a system, called \model (\modellong), and make it freely available for academic research purposes.
\model takes multi-camera video, and outputs SMPL-X~\citep{pavlakos2019expressive} bodies at each frame for every person.
We focus on the two-person case with close interactions where close contact is problematic for marker-based methods.

To achieve this, we introduce several key innovations and design choices.
We adopt a two-stage approach of estimating virtual markers in all camera views and then solving for the optimal human body that fits these markers.
Previous work has taken a similar approach, either estimating sparse anatomical markers like 2D joint locations~\citep{opencap} or dense surface markers~\citep{patel2024camerahmr,hewitt2024look} from every camera view.
We follow the latter direction and train a dense 2D-landmark detector.
However, landmark detectors are usually trained on images of people in isolation, making it difficult to guide the network to identify the correct person or correctly assign body parts when subjects have close-range interactions. Also, for real datasets it is hard to obtain fine-grained information like contact or visibility.

We address this in two ways.
First, we extend the synthetic human video dataset BEDLAM~\citep{black2023bedlam} by assembling close-interaction motions (e.g., dancing) and rendering realistic multi-view sequences.
Then, we define dense surface landmarks~\cite{patel2024camerahmr,hewitt2024look} and construct a new dataset, \dataset.

Second, rather than train the model to directly regress keypoints, we formulate a novel transformer-based dense landmark estimator, \net, which learns distinct embeddings for each keypoint, which we refer to as landmark queries.
We then train a transformer decoder that takes these queries together with the image patch embeddings and outputs landmark locations, together with uncertainty, visibility and contact information.
We find visibility and contact estimates to be critical for dealing with close person--person interaction and to prevent interpenetration.
We demonstrate that our approach is more robust than traditional landmark estimation.

Our method can also be optionally conditioned on additional information to help resolve ambiguities in the multi-person case.
In particular, we use SAM 2~\citep{sam2} to compute and track masks corresponding to the people, and condition the network on these.
This highlights a key advantage of vision-based systems over marker-based ones -- richer pixel-based supervisory signals can be incorporated to reduce ambiguity. We also assign the SAM2 tracking labels to each landmark prediction. This creates a robust geometric-temporal information for each person, which let's us match the people across views using symmetric epipolar distance as cost function.

From the estimated landmarks, uncertainty, visibility and contact, we fit SMPL-X by projecting landmark locations on the body mesh into each calibrated image, and optimizing the pose and shape parameters to minimize 2D reprojection error. Thanks to the accuracy of our landmarks, optimization does not rely on pose priors or pose initialization.
Our resulting system allows us to process complex movements, with a level of accuracy that is visually indistinguishable from that of a marker-based system (see \cref{fig:teaser}).

We perform extensive ablations to evaluate which design choices are important and evaluate performance on video datasets with both a single subject and two-person interaction.
The latter includes Harmony4D~\citep{khirodkar2024harmony4d}, CHI3D~\citep{fieraru2020three}, and Hi4D~\citep{yin2023hi4d}.
We find that our \model outperforms all academic methods.

However, a key question remains: how does the method measure up against a commercial, marker-based system such as Vicon?
To answer this, we perform a {\em unique quantitative evaluation} of our pipeline versus one that fits SMPL-X to Vicon markers using MoSh++~\citep{amass}.
We use held-out 3D markers to evaluate how well each method captures the 3D body shape and pose.
We find that the difference in held-out marker error between the methods is only 0.862mm and that resulting animations are visually indistinguishable.
This suggests that our markerless approach can be used as ``ground truth,'' replacing marker-based systems for SMPL-X capture.

\section{Related Work}
\noindent\textbf{Traditional Marker-Based Motion Capture.}
Commercial marker-based motion-capture systems, such as Vicon~\citep{vicon}, OptiTrack~\citep{optitrack}, Qualisys~\citep{qualisys}, Motion Analysis Corporation~\citep{motionanalysis}, and PhaseSpace~\citep{phasespace}, are widely used in biomechanics, animation, and virtual reality to capture precise 3D motion data.
These systems employ either passive or active markers to track movement, providing high-fidelity data.
However, placing markers, calibrating for both the environment and subject, and cleaning the data introduces substantial overhead in the capture process.
Captured sequences often require extensive manual post-processing to fix noisy, missing, or swapped markers, a task that can take from minutes to hours depending on the complexity of the motion.
In addition, extra processing is required to convert the raw capture data into the parameters of a 3D skeleton or parametric body model~\citep{loper2014mosh}.
Our pipeline removes the need for these manual steps by exploiting rich information present in the pixels to directly regress dense 2D surface landmarks, recovering the shape, pose, and translation parameters of the SMPL-X model from multi-view video of one or two subjects.

\noindent\textbf{Vision-Based Markerless Motion Capture.}
Several industry solutions for markerless motion capture have been proposed in recent years~\citep{captury,theia,moveai,moverse}.
However, with the exception of Moverse~\citep{moverse}, no such methods are capable of producing the parameters of a parametric body model such as SMPL~\citep{SMPL:2015}.
They also do not report accuracy on standard benchmarks, making it difficult to compare solutions or evaluate their accuracy.
Unfortunately these implementations are not open source.
Thus, here we focus on academic systems.

Markerless motion capture has a long history.
Many vision-based approaches focus on detecting sparse 2D human-body keypoints from monocular video sequences, either for individuals~\citep{wei2016cpm, pishchulin16cvpr} or crowds~\citep{8765346, alphapose}.
More-recent monocular methods adopt parametric body models such as SMPL~\citep{SMPL:2015} and SMPL-X~\citep{pavlakos2019expressive} as priors and regress the parameters of these models from a single view~\citep{hmrKanazawa17, goel2023humans, patel2024camerahmr};
these struggle with depth ambiguity and occlusion when people closely interact.
To capture accurate 3D human pose, multi-view systems triangulate 2D detections or learn cross-view feature representations~\citep{iskakov2019learnable, Srivastav_2024_CVPR, zhang2021direct, liao2024multiple}.
As public multi-view datasets~\citep{IonescuSminchisescu11, Joo_2017_TPAMI, mono-3dhp2017} typically annotate only sparse keypoints, such methods stop at the skeleton level, and require an additional model-fitting stage, necessitating strong priors, to obtain full surface parameters.
Instead, we predict a dense set of surface landmarks, rich enough to solve for both SMPL-X body and hand pose and shape, without heavy manual priors.
By conditioning the landmark regressor on an instance-segmentation mask, our proposed approach is able to cleanly separate overlapping subjects, remaining robust in scenes with close human interactions.

\noindent\textbf{Motion-Capture Datasets.}
Current progress in human pose and shape estimation has enabled the creation of a number of human-centric datasets.
Several of them provide single-person sequences with mesh pseudo-ground truth annotations~\citep{vonMarcard2018, kaufmann2023emdb, Huang:CVPR:2022, tripathi2023ipman, Sigal:IJCV:10b, ionescu2014human3, Trumble:BMVC:2017}.
Others, provide labeled sequences of human--object interaction~\citep{bhatnagar2022behave, fan2023arctic, hassan2019resolving, taheri2020grab}. There are datasets that provide human--human interactions \cite{Joo_2015_ICCV, mclean2025embody}, however these interactions are limited to simple actions where often both people are apart from each other.
Creating datasets with heavy interaction between people remains challenging due to frequent occlusions and such datasets are  restricted to laboratory environments~\citep{Joo_2017_TPAMI, singleshotmultiperson2018, belagiannis20143d, fieraru2020three, yin2023hi4d, khirodkar2024harmony4d}.

The development of synthetic datasets has played a crucial role in advancing research on human pose and shape estimation, particularly in addressing the limitations of real-world data in terms of scale, diversity, and annotation accuracy.
Synthetic data makes it possible to obtain large-scale, densely labeled samples under diverse conditions, free from the constraints of physical capture.
However, most existing synthetic datasets do not contain person--person interaction~\citep{Patel:CVPR:2021, black2023bedlam, tesch2025bedlam2, kocabas2024pace, hewitt2024look}. Among those that do~\cite{yin2024whac, fang2024capturing}, the range of activities and motion diversity is limited.

\section{Dataset}

\subsection{\model Synthetic Training Data}
We train \net exclusively using synthetic data.
To extend the coverage of existing data, targeting (i) multi-view coverage, (ii) extreme pose, (iii) high-fidelity hands, and (iv) two-person interaction scenarios, we introduce \dataset, an extension of BEDLAM~\citep{black2023bedlam}, containing 2.5M samples (crops).

\begin{figure}
\centering
\setlength{\tabcolsep}{0pt}
\renewcommand{\arraystretch}{0}
\resizebox{.92\linewidth}{!}{
\begin{tabular}{ccccc}
    \includegraphics[width=0.2\linewidth]{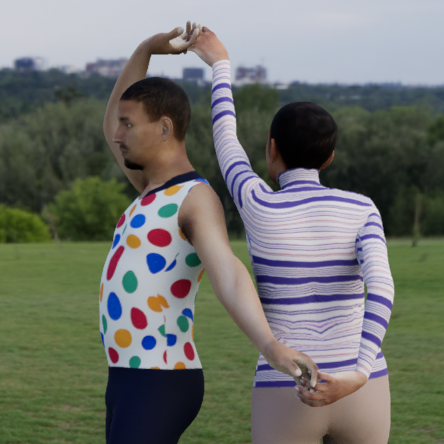} &
    \includegraphics[width=0.2\linewidth]{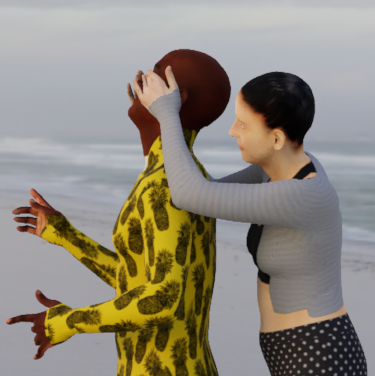} &
    \includegraphics[width=0.2\linewidth]{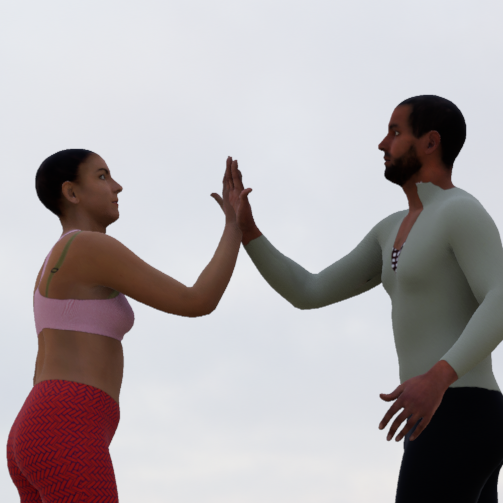} &
    \includegraphics[width=0.2\linewidth]{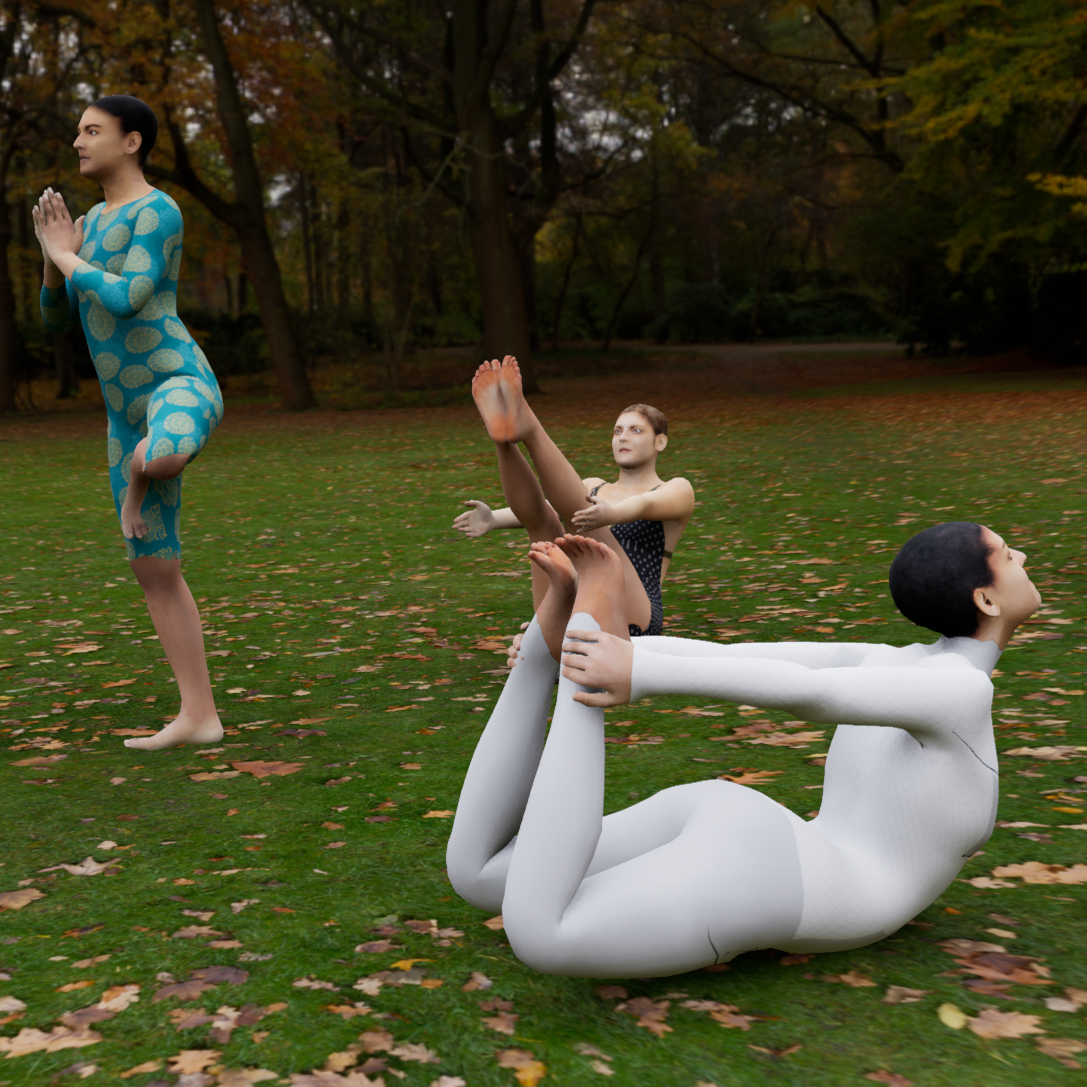} &
    \includegraphics[width=0.2\linewidth]{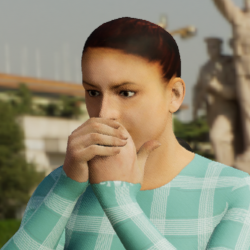} \\

    \includegraphics[width=0.2\linewidth]{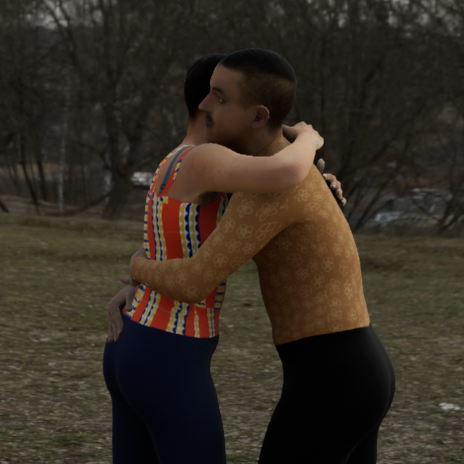} &
    \includegraphics[width=0.2\linewidth]{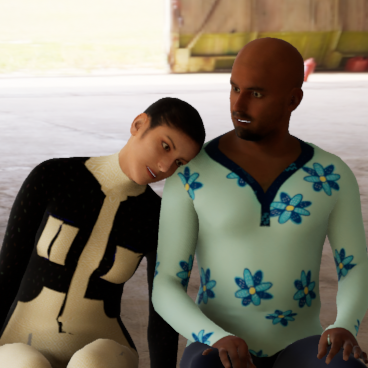} &
    \includegraphics[width=0.2\linewidth]{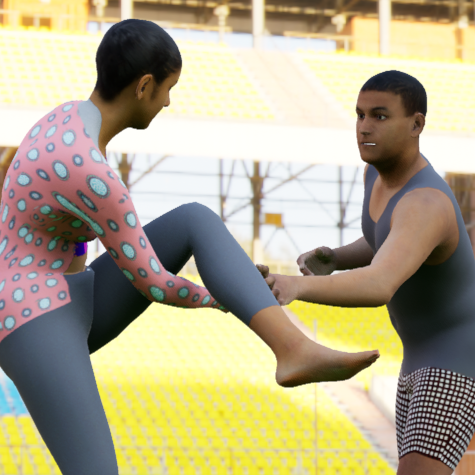} &
    \includegraphics[width=0.2\linewidth]{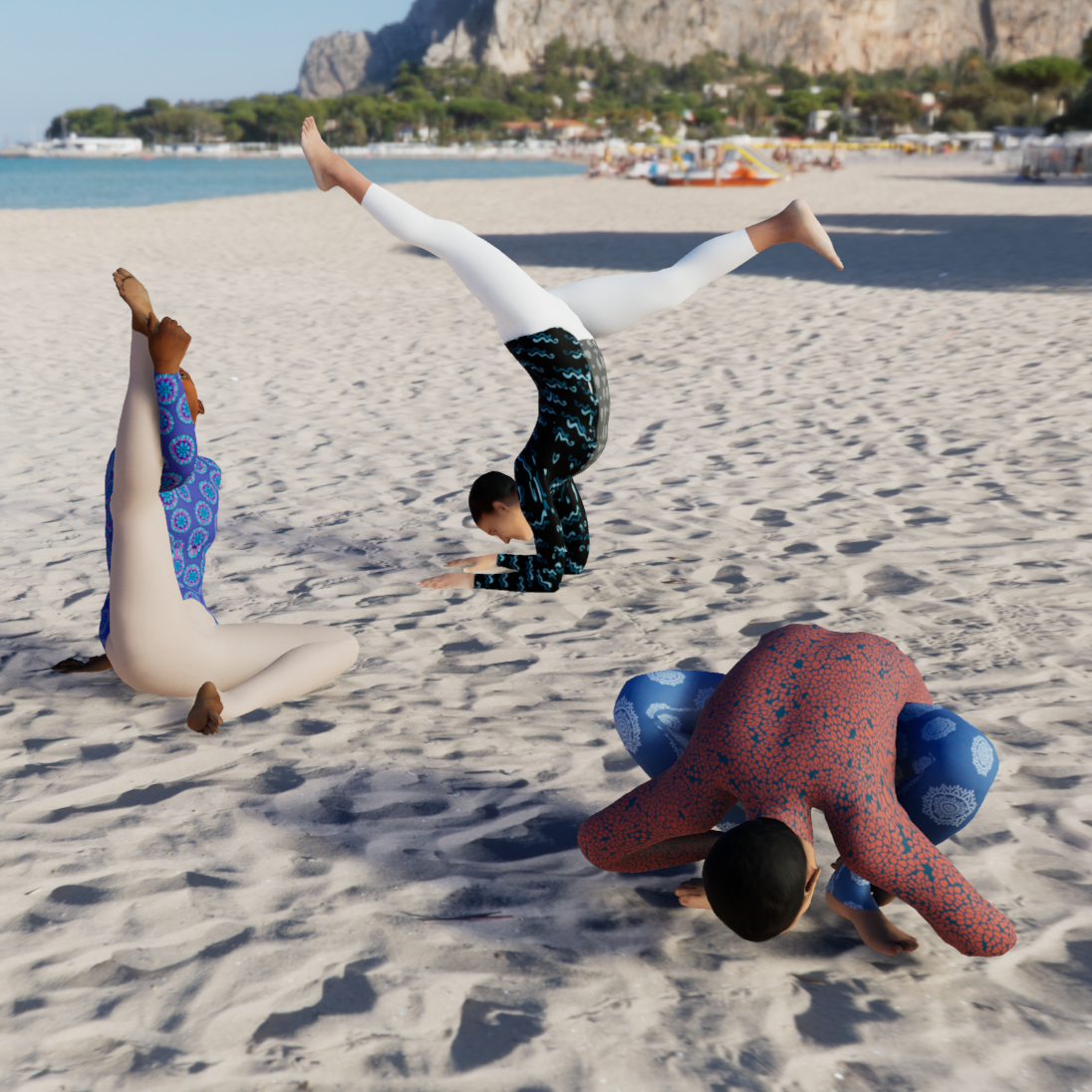} &
    \includegraphics[width=0.2\linewidth]{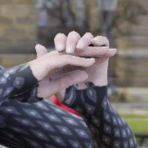} \\
\end{tabular}
}
\caption{Cropped samples from \dataset dataset.}
\label{fig:synthetic-data}
\end{figure}

We extend the original BEDLAM pipeline to a 32-camera multi-view configuration using a virtual camera setup.
We sample motion sequences from available datasets and capture two additional marker-based (Vicon) datasets (\cref{fig:synthetic-data}): (i) Latin-Dance (2 subjects, 10 dancing sequences), and (ii) Interacting Couples (2 subjects, 48 two-person interaction sequences).
We collect these to supply high-quality interaction sequences missing from BEDLAM.
We represent all the 3D humans in SMPL-X~\citep{pavlakos2019expressive} format. See \supmat for rendering details. We save per-subject segmentation masks, depth maps, and per-vertex visibility. We also generate per-vertex contact labels for floor and subject interactions by computing the Sign Distance Function (SDF) along with the surface normals. The resulting dataset consists of 955k images (\cref{fig:synthetic-data}).

\begin{figure}
\centerline{\includegraphics[width=1.0\linewidth]{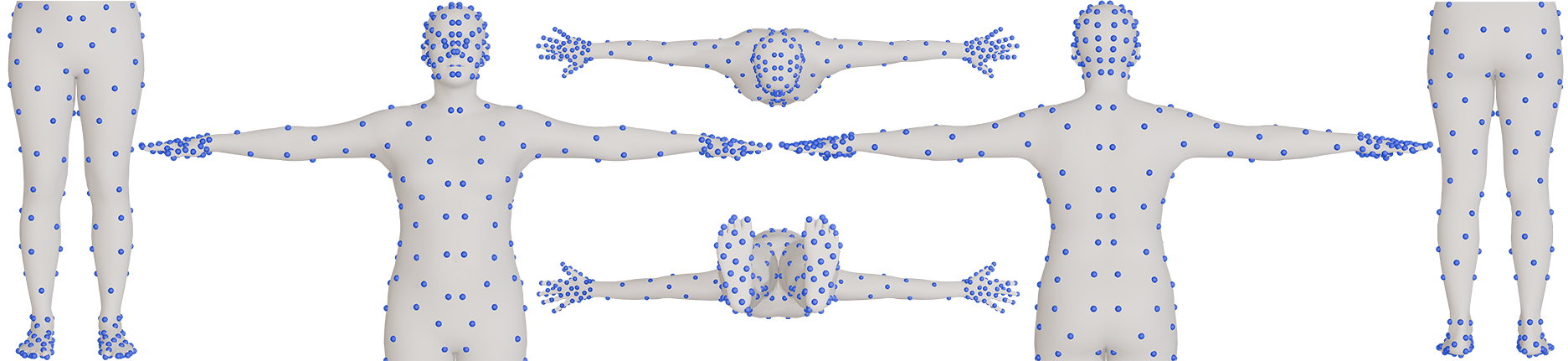}}
\caption{512 Landmarks sampled from the SMPL-X body.}
\label{fig:landmarks}
\end{figure}

For the dense-landmark ground truth, we sample 512 vertices from the SMPL-X body using Farthest Point Sampling (FPS), assigning more weight to vertices on the hands, feet, and head.
This provides the network more information for smaller and more articulated body regions, as seen in \cref{fig:landmarks}.
Our curated dataset can be categorized into three sub-datasets:

\noindent\textbf{\dataset-S (Single Person):}
We sample SMPL-X motion sequences from the BEDLAM data pool, which offers a wide range of motions and body shapes.
To broaden the pose distribution, we augment it with sequences from the MOYO training set.

\noindent\textbf{\dataset-I (Interactions):}
In addition to our own motion-capture recordings of Latin-dance routines and other two-person interactions, we include sequences from the Hi4D~\citep{yin2023hi4d}, Harmony4D~\citep{khirodkar2024harmony4d}, and Inter-X \citep{xu2024inter} training sets.
The resulting collection covers a diverse range of contact-rich interactions, from social exchanges to coupled dances.

\noindent\textbf{\dataset-H (Hands)}:
To improve the diversity of the sampled hand articulations we add SignAvatars~\citep{yu2024signavatars}. We also introduce hand motions from Interhand2.6M~\citep{Moon_2020_ECCV_InterHand2.6M}, which contains captured sequences of hand-to-hand interaction.
We merge these hand poses with the body parameters of BEDLAM by fitting SMPL-X to the MANO~\citep{mano} hand sequences (\cref{fig:synthetic-data}).

\subsection{\model Real Evaluation Data}

We captured three datasets to evaluate our method: \datasetEval-(S)ingles (16 views), \datasetEval-(D)ance (32 views), and \datasetEval-Extra (32 views).
All datasets were captured with a marker-based system (Vicon) synchronized with a multiview RGB system~\citep{ioi}.
Marker data was manually cleaned and then processed with MoSh++~\cite{amass} to obtain SMPL-X annotations paired with multi-view images.

Each dataset serves a different evaluation purpose. \datasetEval-Singles consists of 3 individuals in 22 single-subject sequences, including object interaction.
\datasetEval-Dance includes 18 West Coast Swing dance sequences performed by a professional dance couple, whose close partner interaction is intrinsic to the genre.
\datasetEval-Extra was acquired explicitly to benchmark our method against the Vicon-MoSh marker-based pipeline.
In addition to using a standard Vicon marker set (FrontWaist10Fingers), we add 37 additional markers on each of the 3 subjects.
These extra markers are not used in fitting SMPL-X with MoSh++, and provide independent ground-truth, enabling fair comparison between marker-based and markerless methods.
Subjects performed 4 sequences each, including walking, dancing, and basic sports motions. \supmat has more details regarding the marker-based capture pipeline and configuration.

As part of our contribution, we additionally release a new dance dataset captured solely using \model, covering Bachata, West Coast Swing, Breakdance and Ballroom. It consists of 9 subjects performing for approximately 1 hour in total, recorded at 30 FPS.

\section{Method}
\subsection{Dense Landmark Estimation}
\begin{figure*}[tbp]
\centerline{\includegraphics[width=.85\linewidth]{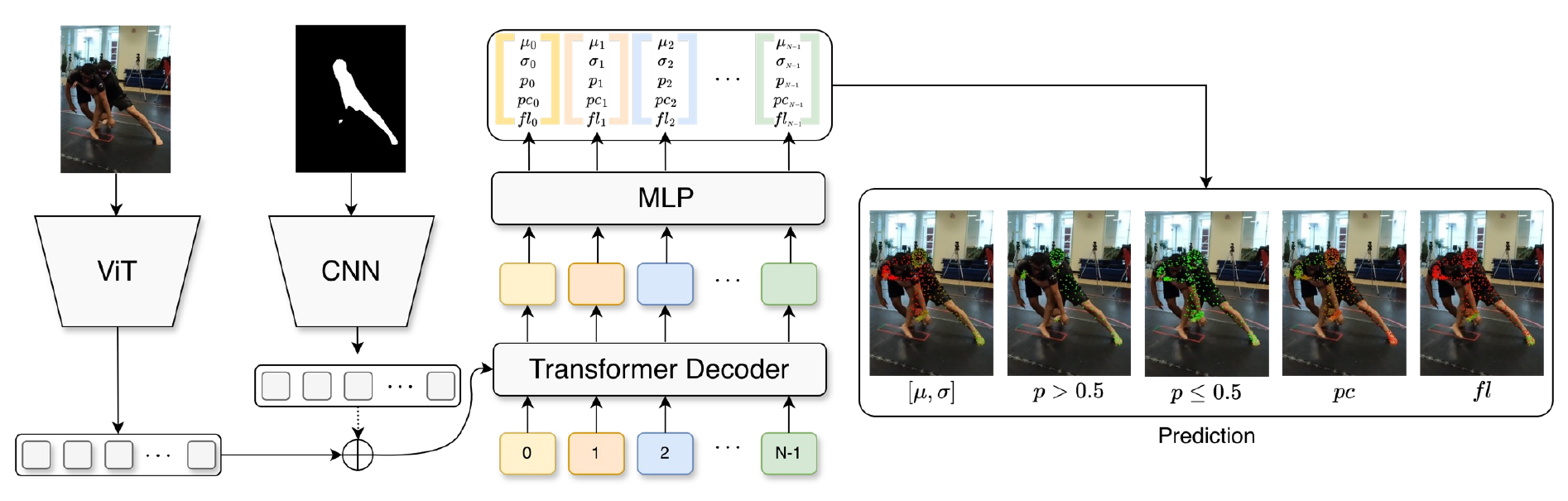}}
\caption{\net. The input to the network is the image and mask. It predicts per landmark visibility probability $p$ (green is visible, red not visible), landmark locations $\mu$, uncertainties $\sigma$ (red means highly uncertain), person--person $pc$ and floor contact $fl$ probabilities (red means no contact and green contact).}
\label{fig:network}
\vspace{-0.15in}
\end{figure*}

Our novel dense 2D landmark estimator, \net, extracts image features using ViT-Base and includes an additional CNN to process the mask (\cref{fig:network}).
It decodes $N=512$ surface landmarks using a Transformer decoder.
Unlike CameraHMR~\citep{patel2024camerahmr}, which uses a single learnable embedding input to estimate the entire set of landmarks, $\mathbf{L}$, we learn $N$ embeddings (landmark queries), with one corresponding to each landmark.
By designing the network to decode landmarks from individual queries, cross-attention layers learn to match each landmark to the most relevant image patches.
At the same time, the self-attention layers in the decoder are able to learn the pairwise correlations between landmarks.
We design the network so that the image and mask features are encoded to the same space, where we combine them with element-wise summation to integrate mask conditioning~\citep{kirillov2023segment}.
We find the proposed network architecture enables the model to generalize
to challenging poses (see \cref{fig:correlation_net_comp}).

For each landmark index $i\in\{0, ..., N-1\}$, the network predicts pixel coordinates $\mu_i =[x_i, y_i]$ and the associated uncertainty $\sigma_i$~\citep{wood20223d, hewitt2024look, patel2024camerahmr}.
We observed that the predicted uncertainties are often high on rigid regions (e.g. torso) that are not visible by the camera.
Due to this, we additionally task the network with estimating the visibility probability $p_i$ for each landmark.
We also introduce a person--person contact $pc_i$ and a floor contact $fl_i$ prediction layers.
To train the dense landmark detector, we use a Gaussian negative log likelihood loss. For the visibility probability, we use binary cross entropy. For the contact probabilities, we use focal loss~\cite{lin2017focal}, because there are more landmarks that do not have contact. All the losses have per-landmark weights $\lambda^*$. We denote $\mu_i' =[x_i', y_i']$ and $p'_i \in \{0, 1\}$ as ground truth landmark and visibility, respectively.

\subsection{Model Fitting}

Our method fits the SMPL-X neutral body, $\mathcal{M}(\boldsymbol{\beta}, \boldsymbol{\theta}, \boldsymbol{t})$, to a multi-view sequence of one or two people using L-BFGS~\citep{liu1989limited}.
We initialize all the SMPL-X bodies at the 3D point that best minimizes the distance between the rays projected from the camera centers.
Unlike other methods~\citep{hewitt2024look, patel2024camerahmr}, we do not initialize the pose and shape of the person with any regression method, as our surface landmarks carry enough information about the person and the scene. We assume the calibration parameters are given.
We use 16 shape coefficients ($\boldsymbol\beta$).

Our optimization process is as follows.
The first step involves finding the translation and rotation of the bodies by minimizing the reprojection error:
\begin{equation}
E_{\text{ldmks}} =
\frac{1}{C}\sum_{t,c,l}
\rho\!\Bigg(
\tfrac{\lVert \boldsymbol{\mu}_{t,c,l}
- \Pi(\mathbf{V}_{t,l}, \mathbf{Q}_c) \rVert}{\sigma_{t,c,l}}
\Bigg) p_{t,c,l}
\label{eq:reprojection_loss}
\end{equation}
where $E_{\text{ldmks}}$ is the energy term that penalizes the reprojection error between $\textbf{L}$ and the projected SMPL-X vertices $\textbf{V}_{t,l}$ using camera parameters $\textbf{Q}_c$ and
where $[\boldsymbol{\mu}_{tcl},\sigma_{tcl},p_{tcl}]$ is the predicted 2D location, uncertainty and visibility probability for the $l^{th}$ landmark. $\rho$ is the robust Geman-McClure function \cite{GemanMcClure1987}.
The second stage optimizes the pose, shape, and translation parameters with $E_{\text{ldmks}}$ and an L2 regularizer on the body priors $E_{\text{shape}}$.
In the third stage we use the same energies but update the uncertainty values. Sometimes, the network is not confident about some predictions, even when the location of the landmark is correct. Thus, we use the reprojection error $e_i$ as evidence to weight down the uncertainties, $\sigma'_i = \sigma_i \cdot \min\!\left(\max\!\left(\frac{e_i}{\tau},\, 0\right),\, 1\right)$, where $\tau=10$px. This is possible because the second stage result is close to local optimum. We also penalize large joint accelerations $E_{\text{temp}}$.

As final stage we optimize the contact. Assuming that each per-view network output provides an approximately unbiased and calibrated estimate of the local contact probability, conditioned on that view, we average the contact probabilities to estimate the expected contact probability. We do this for foot and person--person contact predictions and use them to minimize $E_{\text{cont}} = E_{\text{p}}+E_{\text{c}}$. where $E_{\text{p}} = \frac{1}{N} \sum_{i=1}^{N} \big[\,\min\big(0,\ \text{SDF}_{\text{other}}(\mathbf{v}_i) + \delta\big)\,\big]^2$ is a repulsion term that penalizes all the vertices that are inside the body of the other person. $\delta$ allows for limited interpenetration to account for soft-tissue deformation. $E_{\text{c}}= \frac{1}{N}\sum_{i=1}^{N}\,p_i\,\big[\max\big(0,\,\text{SDF}_{\text{other}}(\mathbf{v}_i)\big)\big]^{2}$ is a contact term that attracts the points that are above the surface, this term is weighted by the predicted contact probabilities and their corresponding vertices. SDF is the Sign Distance Function from~\cite{Mueller:CVPR:2021}. We re-implemented the SDF with a custom CUDA kernel for full GPU acceleration.

The final energy function is $E(\boldsymbol{\Phi};\mathbf{Q}, \mathbf{L}) = E_{\text{ldmks}}+E_{\text{shape}} + E_{\text{temp}} + E_{\text{cont}}$.
$\boldsymbol{\Phi} = \{ \boldsymbol{\beta}, \boldsymbol{\theta}, \boldsymbol{t} \}$ are the SMPL-X parameters, $\mathbf{Q}$ represents the camera parameters, and abusing notation $\mathbf{L}$ represents the predicted landmarks, uncertainty, visibility and contact probabilities.

\subsection{Multiview Correspondence}

Accurate subject identification is essential within each camera view and across multiple views, for convergence in the model fitting. We first obtain the masks with SAM2~\cite{sam2} which can be initialized once in several ways depending on the setting: either with automatic subject detection using a Grounding LLM or a bounding-box detector, or manually with a few clicks on the the subjects in a single frame.

We use SAM2 segmentation propagation as a way to track a person across the sequence. For each tracked person, we predict their 2D dense landmarks per frame $f \in \left\{ 0,\dots,F-1 \right\}$. Therefore, for each view $v \in \left\{ 0,\dots,V-1 \right\}$, each person $p \in \left\{ 0,\dots,P-1 \right\}$ has $\mathbf{x} \in \mathbb{R}^{F \times N \times 2}$ 2D landmarks. If the person is not visible in some frames, we assign a dummy value.

For each tracked person, we predict 2D dense landmarks for every frame
$f \in \{0,\dots,F-1\}$, view $v \in \{0,\dots,V-1\}$, and person
$p \in \{0,\dots,P-1\}$, resulting in
$\mathbf{x}_{v,p} \in \mathbb{R}^{F \times N \times 2}$.
If the person is not visible in a given frame, the corresponding value is ignored.

All detected people across all views are compared pairwise to establish cross-view correspondences. Given two sets of landmarks $\mathbf{x}_{a},\mathbf{x}_{b}$ from two views, their geometric affinity is defined as $A_g(\mathbf{x}_a,\mathbf{x}_b)= \exp(-D_g / \lambda)\in[0,1]$ where $\lambda$ is a temperature parameter and $D_g$ (\cref{eq:geometric_consistency_formula}) is the symmetric epipolar distance~\cite{dong2019fast}. To avoid possible wrong predictions from non-visible landmarks, we only compute the affinity between the landmarks that are visible by both views.
\begingroup
\begin{equation}
\label{eq:geometric_consistency_formula}
D_g = \frac{1}{2FN}
\sum_{i=1}^{FN}
\Big(
d(\mathbf{x}^{i}_{b},\,\mathbf{F}_{ba}\mathbf{x}^{i}_{a})
+
d(\mathbf{x}^{i}_{a},\,\mathbf{F}_{ab}\mathbf{x}^{i}_{b})
\Big)
\end{equation}

\endgroup

$\mathbf{F}_{ba}$ is the fundamental matrix between two views, $d(\mathbf{x},\mathbf{l})$ denotes the point–line distance in pixel coordinates. If fewer than a minimum number of landmarks are visible or if $D_g$ exceeds a threshold, the affinity is set to zero. The pairwise affinities between all detections in views $a$ and $b$ form the matrix $\mathbf{A}_{ab}\in[0,1]^{M_a\times M_b}$, where $M_a$ and $M_b$ denote the number of detected persons in each view. Optimal one-to-one matches are obtained by solving a Hungarian assignment on the cost matrix $1-\mathbf{A}_{ab}$. All high-affinity matches are then linked across all view pairs to form a cycle-consistent correspondence graph, whose connected components represent consistent multi-view groups corresponding to the same person across views.

\section{Experiments}
\subsection{2D Landmark Network Comparison}
We compare \net with two existing dense-landmark models: CameraHMR~\citep{patel2024camerahmr}, which uses the same backbone as ours and a transformer decoder with a single learnable token and output for all the landmarks.
The second model is a re-implementation of \citet{hewitt2024look}; their code is not publicly available.
For fair comparison, we remove the pose and shape heads and only use the landmark head.
We refer to the resulting network as Look-Ma*.

The input resolution is set to $512\times384$, consistent with the transformer-based models~\cite{xu2022vitpose}. As the other models do not predict contact, we trained a version of our model that only predicts uncertainty and visibility. For implementation details see \supmat.

All models are trained on cropped images from BEDLAM and evaluated on real datasets.
For single-person evaluation, we use RICH~\citep{Huang:CVPR:2022} and MOYO~\citep{tripathi2023ipman}.
For two-person interaction, we evaluate on Harmony4D~\citep{khirodkar2024harmony4d} and CHI3D~\citep{fieraru2020three}.
We also report results on our datasets: \datasetEval-S and \datasetEval-D.

\cref{tab:network_ablation_single} and \cref{tab:network_ablation_double} show the average 2D error between the ground truth 2D landmarks and the predicted landmarks across images for single and two-person datasets.
\cref{tab:results_without_vtemplate} reports standard 3D errors after fitting the model to the predicted markers; i.e.\ Mean Per-Joint Position Error (MPJPE) and Per-Vertex Error (PVE).
For both 2D marker prediction and 3D fitting, \net is more accurate than prior methods.

\begin{figure*}[ht]
\centering
\resizebox{.97\linewidth}{!}{%
\setlength{\tabcolsep}{1pt}
\begin{tabular}{@{}cccccc@{}}
    Input image & GT & Look-Ma* & CameraHMR  & \net  & \net +masks  \\
    \includegraphics[width=.20\linewidth]{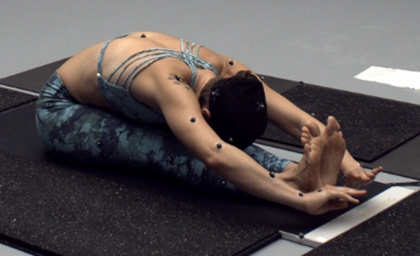} &
    \includegraphics[width=.20\linewidth]{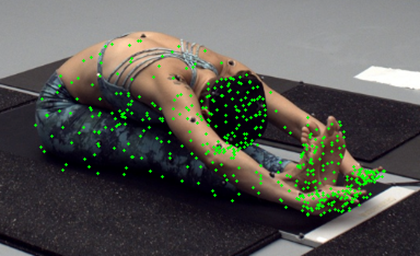} &
    \includegraphics[width=.20\linewidth]{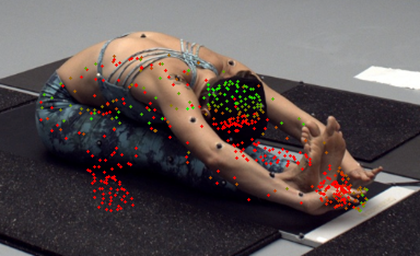} &
    \includegraphics[width=.20\linewidth]{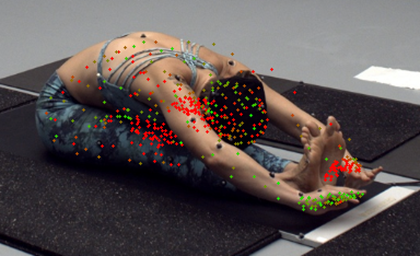} &
    \includegraphics[width=.20\linewidth]{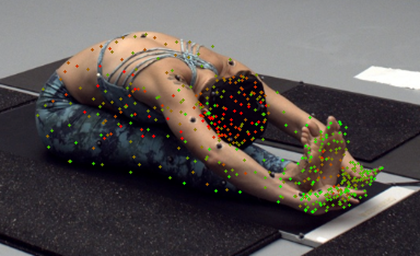} &
    \includegraphics[width=.20\linewidth]{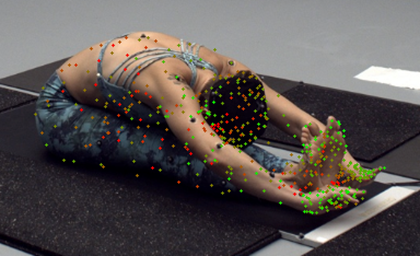}
    \\
\end{tabular}
}
\caption{Comparison on extreme poses. Ground-truth landmarks are shown in green. For each prediction, landmarks are color-coded: red indicates higher pixel error, green indicates lower pixel error. We compare networks trained on BEDLAM.}
\label{fig:correlation_net_comp}
\vspace{-0.15in}
\end{figure*}

Our network can also generalize to complex and unseen poses, unlike CameraHMR and Look-Ma*, and predict the limbs of the person more accurately.
\cref{fig:correlation_net_comp} and \cref{tab:network_ablation_single} shows that \net correctly estimates the landmarks from the yoga dataset (MOYO) even though it was never trained on such poses.

We additionally train a variant of our network conditioned on segmentation masks. During evaluation, we use masks obtained using SAM2. While masks can provide a marginal benefit to the performance of the network in the single-person case (\cref{tab:network_ablation_single}), their main impact is on two-person interactions (\cref{tab:network_ablation_double}), where masks help to disambiguate the target individual.  For \cref{tab:network_ablation_double}, we restrict the evaluation to images where the Intersection-over-Union (IoU) between two people is greater than 0.5. See \supmat for examples.

\begin{table}
\centering
\caption{Dense landmark evaluation on single person datasets. Mean 2D Euclidean distance error (in pixels) between GT and predicted landmarks. Bold is the most accurate and underline is the most accurate without mask.}
\resizebox{0.9\linewidth}{!}{%
\begin{tblr}{
  row{1} = {c},
  cell{2}{2} = {c},
  cell{2}{3} = {c},
  cell{2}{4} = {c},
  cell{3}{2} = {c},
  cell{3}{3} = {c},
  cell{3}{4} = {c},
  cell{4}{2} = {c},
  cell{4}{3} = {c},
  cell{4}{4} = {c},
  cell{5}{2} = {c},
  cell{5}{3} = {c},
  cell{5}{4} = {c},
  hline{1-2,5-6} = {-}{},
}
Model              & RICH                  & MOYO           & \datasetEval-S \\
Look-Ma*           & 13.26                 & 22.43          &  10.25  \\
CameraHMR~         & 8.84                  & 12.53          &  6.32  \\
\net(ours)             & \textbf{\uline{8.55}} & \uline{11.40}  &  \textbf{\uline{6.09}}  \\
\net+masks SAM2 (ours) & 8.83                  & \textbf{11.04} &  6.16
\end{tblr}
}
\label{tab:network_ablation_single}
\end{table}

\begin{table}
\centering
\caption{Two-person datasets dense landmark evaluation. We use the images where IOU $>$ 0.5 between two people.}
\resizebox{0.9\linewidth}{!}{%
\begin{tblr}{
  row{1} = {c},
  cell{2}{2} = {c},
  cell{2}{3} = {c},
  cell{2}{4} = {c},
  cell{3}{2} = {c},
  cell{3}{3} = {c},
  cell{3}{4} = {c},
  cell{4}{2} = {c},
  cell{4}{3} = {c},
  cell{4}{4} = {c},
  cell{5}{2} = {c},
  cell{5}{3} = {c},
  cell{5}{4} = {c},
  cell{6}{2} = {c},
  cell{6}{3} = {c},
  cell{6}{4} = {c},
  hline{1-2,5,6} = {-}{},
}
Model                 & Harmony4D      & CHI3D         & \datasetEval-D       \\
Look-Ma*              & \uline{31.45}  & 8.77          & 15.01         \\
CameraHMR~            & 32.84          & 6.30          & 10.21         \\
\net (ours)     & 31.96          & \uline{6.22}  & \uline{9.87}  \\
\net+masks SAM2 (ours) & \textbf{18.33} & \textbf{4.36} & \textbf{7.70}
\end{tblr}
}
\label{tab:network_ablation_double}
\vspace{-0.05in}
\end{table}

\subsection{Contact Prediction from 2D Landmarks}
The previous evaluation shows that our network is capable of differentiate between two people given a noisy mask. However, we cannot obtain contact information from the BEDLAM dataset as it has single people motions. Therefore, we retrain our network with our curated \dataset and added two branches. One to predict the probability of contact between two people and the second one to predict the probability of floor contact.
Note that unlike \cite{wang2025prompthmr, shin2023wham} this is not only foot contact, but any contact that the body has on the floor.
To evaluate contact, we compute the Receiver Operating Characteristic curve (ROC) and report the area under the curve (AUC) where AUC=0.5 is a random guess and higher AUC is better. For floor contact, we use the MOYO test set, since its yoga poses involve complex floor contact. For person--person contact we use Hi4D contact ground truth. For the latter, we retrain our network without the Hi4D sequences. For both methods we subsample the SMPL-X vertices to the number our network predicts. As \cref{fig:contact_roc} reports, the AUC for both predictions is more than $90\%$ which indicates that our model correctly identifies most contact points with few false positives.

\begin{figure}[t]
\centering
\begin{subfigure}[t]{0.48\columnwidth}
  \centering
  \resizebox{\linewidth}{!}{\begin{tikzpicture}
\begin{axis}[
    width=3.8cm,
    height=3.8cm,
    xmin=0, xmax=1,
    ymin=0, ymax=1,
    xlabel={False Positive Rate},
    ylabel={True positive Rate},
    axis lines=left,
    axis line style={black, thick},
    tick style={black, thick},
    xtick={0,1},
    ytick={0,1},
    grid=none,
    legend=false,
    clip=false,
    label style={font=\scriptsize},
    tick label style={font=\scriptsize},
]

    \addplot[
        name path=roc,
        color=violet!70!black,
        dashed,
        ultra thick,
        mark=none,
    ] table [x=FPR, y=TPR, col sep=comma] {figures/csv/roc_contacts_hi4d.csv};

    \path [name path=baseline] (axis cs:1,0) -- (axis cs:1,0);

    \addplot[
        violet!30,
        opacity=0.6
    ] fill between[of=roc and baseline];

    \addplot[
        gray!70,
        dotted,
        thick
    ] coordinates {(0,0) (1,1)};

    \node[font=\scriptsize, text=black]
        at (axis cs:0.6,0.2) {AUC = 0.91};

\end{axis}
\end{tikzpicture}}
  \caption{Hi4D person contact}
\end{subfigure}
\hfill
\begin{subfigure}[t]{0.48\columnwidth}
  \centering
  \resizebox{\linewidth}{!}{\begin{tikzpicture}
\begin{axis}[
    width=3.8cm,
    height=3.8cm,
    xmin=0, xmax=1,
    ymin=0, ymax=1,
    xlabel={False Positive Rate},
    ylabel={True positive Rate},
    axis lines=left,
    axis line style={black, thick},
    tick style={black, thick},
    xtick={0,1},
    ytick={0,1},
    grid=none,
    legend=false,
    clip=false,
    label style={font=\scriptsize},
    tick label style={font=\scriptsize},
]

    \addplot[
        name path=roc,
        color=yellow!70!black,
        dashed,
        ultra thick,
        mark=none,
    ] table [x=FPR, y=TPR, col sep=comma] {figures/csv/roc_contacts_moyo.csv};

    \path [name path=baseline] (axis cs:1,0) -- (axis cs:1,0);

    \addplot[
        yellow!30,
        opacity=0.6
    ] fill between[of=roc and baseline];

    \addplot[
        gray!70,
        dotted,
        thick
    ] coordinates {(0,0) (1,1)};

    \node[font=\scriptsize, text=black]
        at (axis cs:0.6,0.2) {AUC = 0.96};

\end{axis}
\end{tikzpicture}}
  \caption{MOYO floor contact}
\end{subfigure}
\caption{ROC curve evaluation of our contact predictions.}
\label{fig:contact_roc}
\end{figure}

\subsection{Correspondence Evaluation}
We run our correspondence method on all three interaction datasets. The input is the dense landmark and visibility predictions. We obtain a $100\%$ accuracy in matching the identity across views; see \supmat for ablation. This demonstrates that our dense landmarks provide rich and accurate geometric information, removing the need for an extra identity feature computed by external networks similar to classical approaches~\cite{dong2019fast}. Note that the resulting masks from SAM2 can miss some parts of the body, especially when two people are too close, which might lead to sub-optimal landmark predictions. However, our method remains robust.

\subsection{Markerless Motion Capture Evaluation}
We fit SMPL-X parameters to the test sequences with our method, Look-Ma* and CameraHMR~\citep{patel2024camerahmr}, as well as a multi-view implementation~\citep{easymocap} of SMPLify-X~\citep{pavlakos2019expressive}.
For \model, we test two versions, one with the contact optimization stage (\model-C) and another without it. For a fair comparison, we use the ground truth bounding boxes for all the methods, and SAM2 masks for the mask-conditioned network.

\begin{figure}
\centering
\resizebox{.9\linewidth}{!}{%
\tiny
\setlength{\tabcolsep}{1pt}
\begin{tabular}{@{}cccccc@{}}
    Input & GT & LookMa* & CameraHMR  & \model  & \model-C  \\
    \includegraphics[width=.14\linewidth]{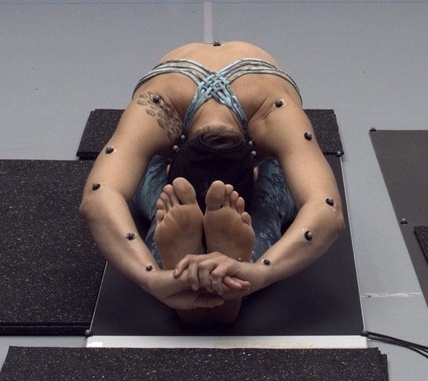} &
    \includegraphics[width=.14\linewidth]{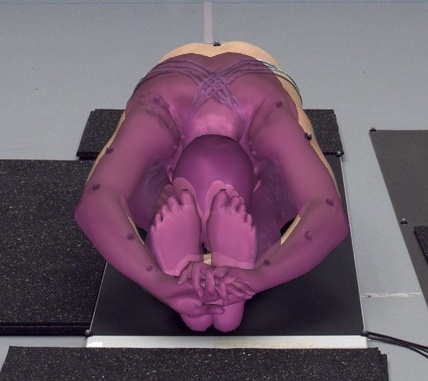} &
    \includegraphics[width=.14\linewidth]{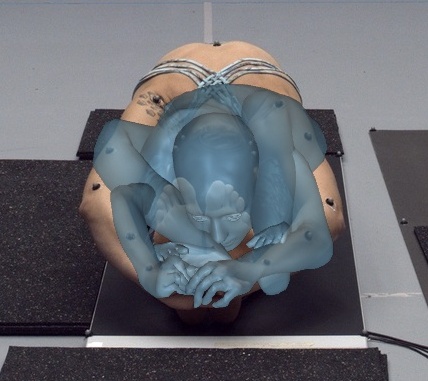} &
    \includegraphics[width=.14\linewidth]{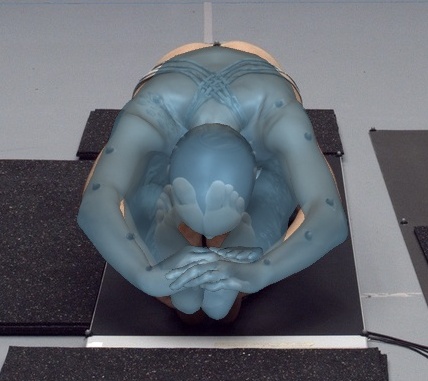} &
    \includegraphics[width=.14\linewidth]  {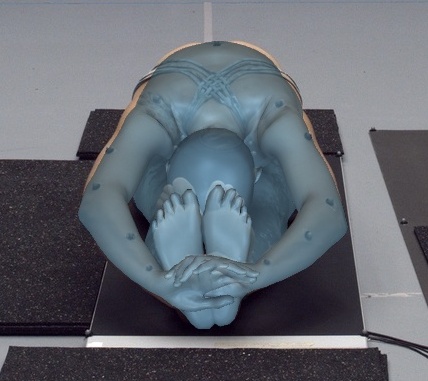} &
    \includegraphics[width=.14\linewidth]{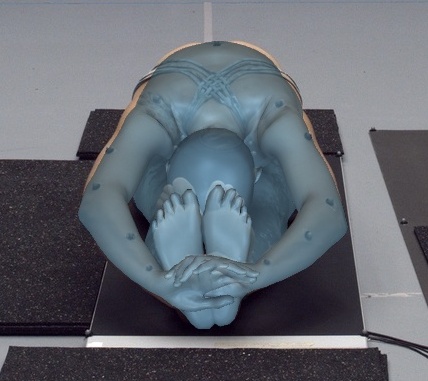}
    \\
    \includegraphics[width=.14\linewidth]{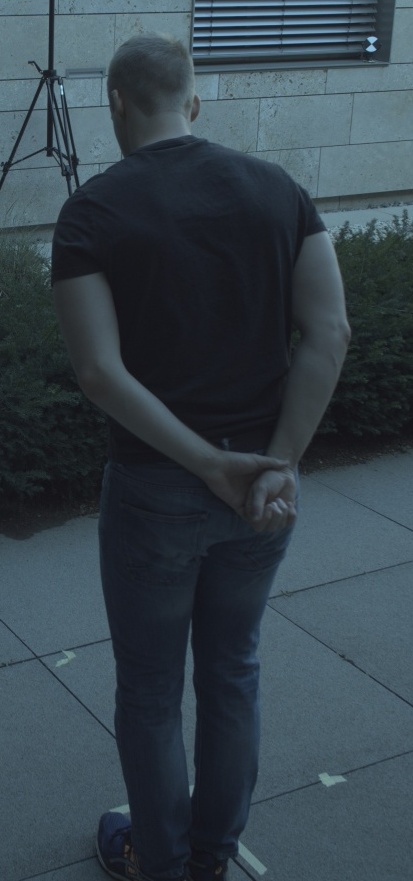} &
    \includegraphics[width=.14\linewidth]{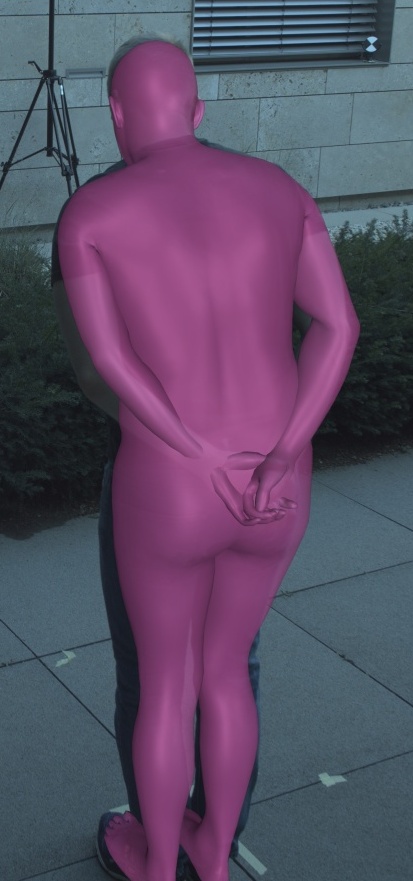} &
    \includegraphics[width=.14\linewidth]{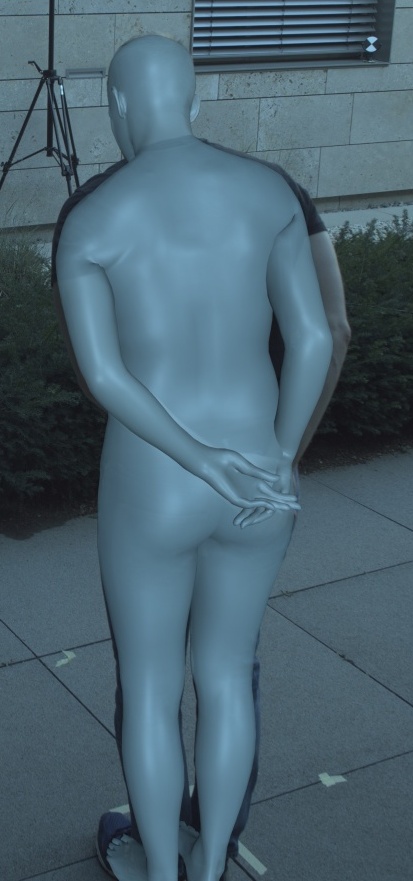} &
    \includegraphics[width=.14\linewidth]{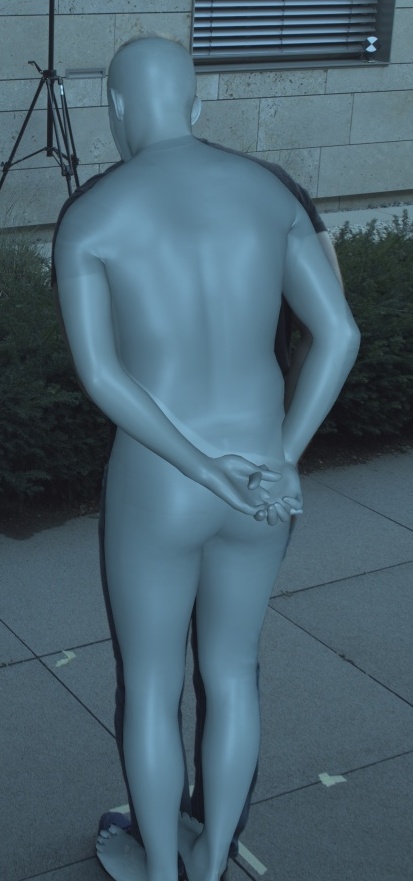} &
    \includegraphics[width=.14\linewidth]{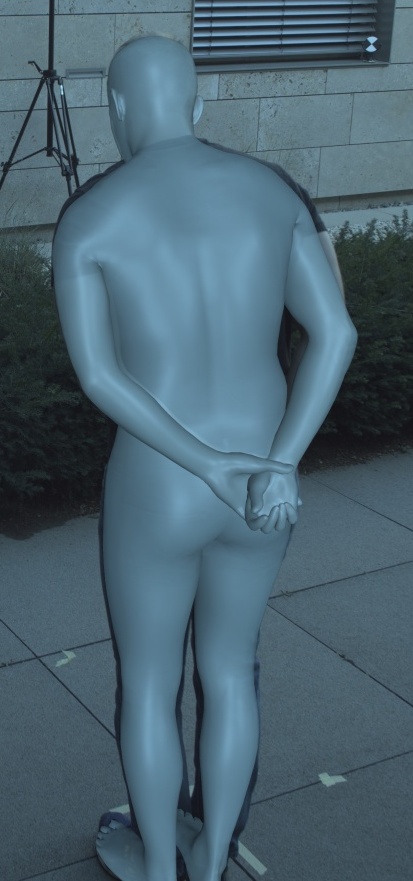} &
    \includegraphics[width=.14\linewidth]{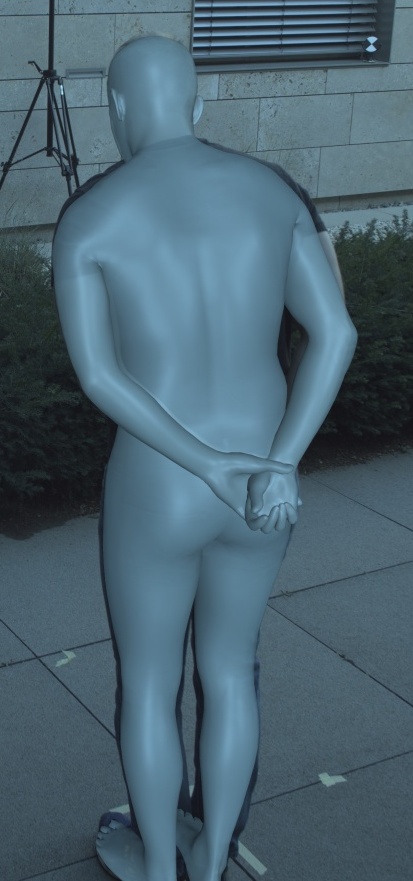}
    \\
    \includegraphics[width=.14\linewidth]{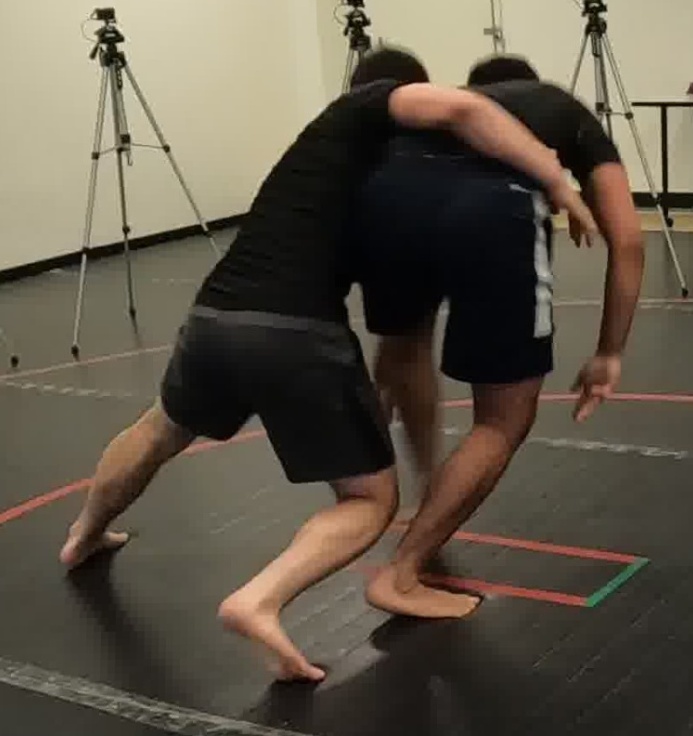} &
    \includegraphics[width=.14\linewidth]{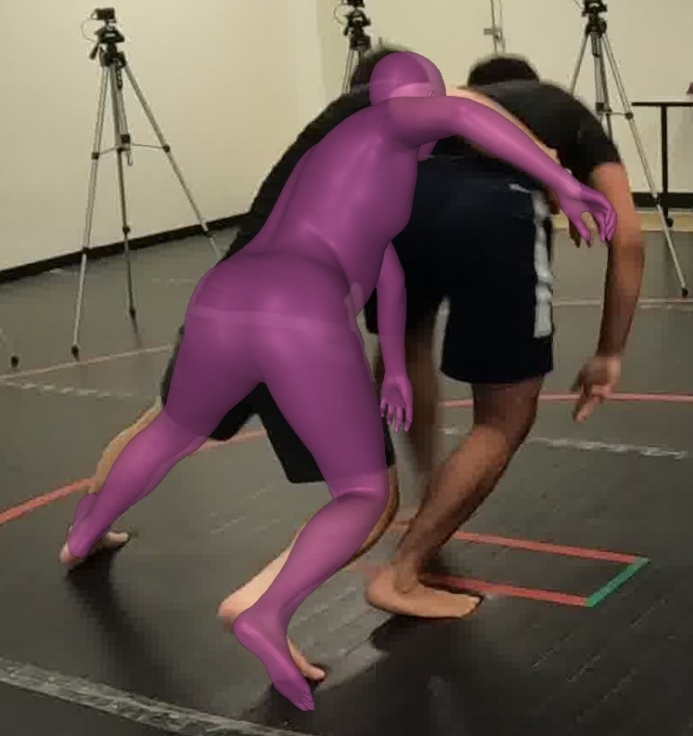} &
    \includegraphics[width=.14\linewidth]{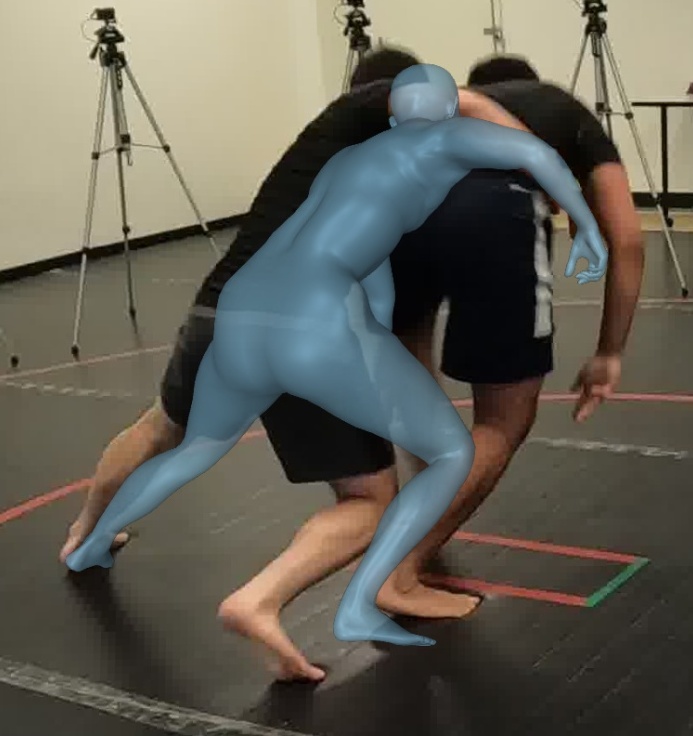} &
    \includegraphics[width=.14\linewidth]{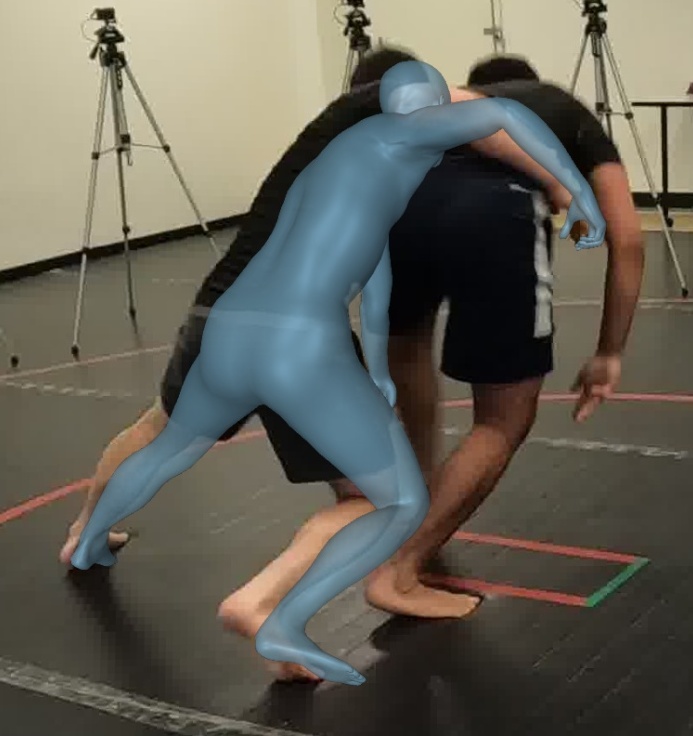} &
    \includegraphics[width=.14\linewidth]{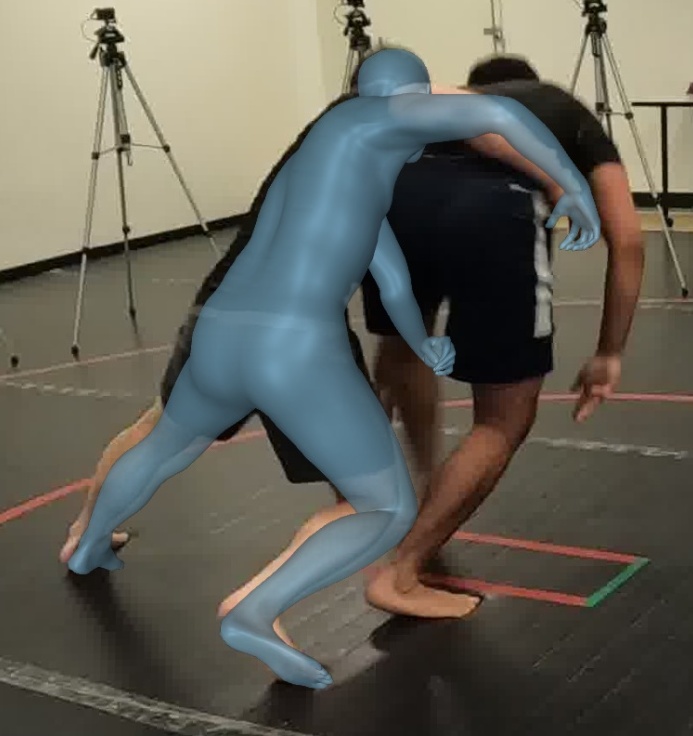} &
    \includegraphics[width=.14\linewidth]{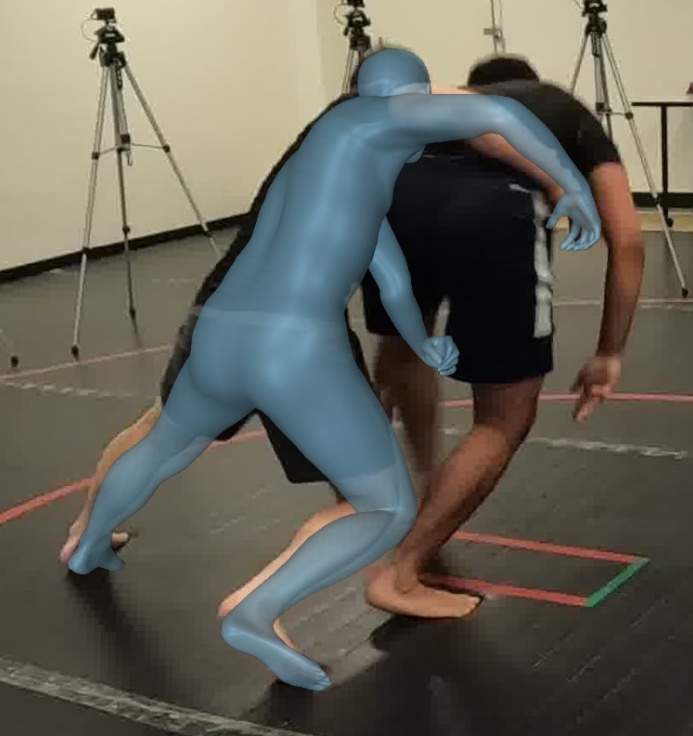}
    \\
    \includegraphics[width=.14\linewidth]{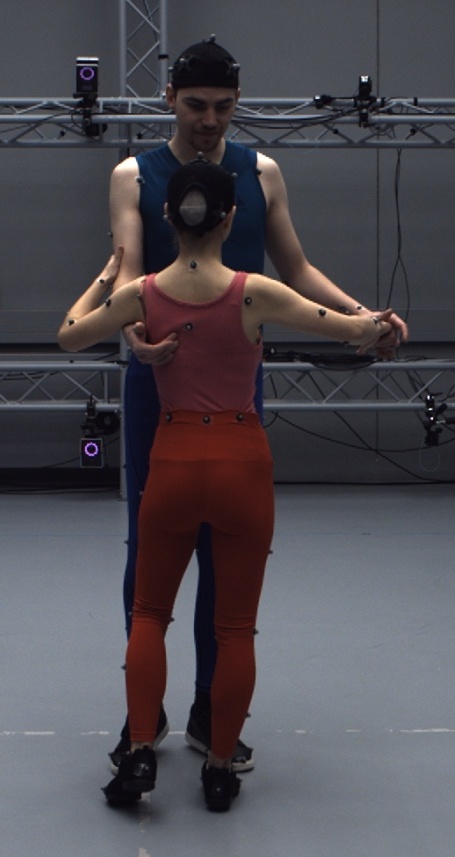} &
    \includegraphics[width=.14\linewidth]{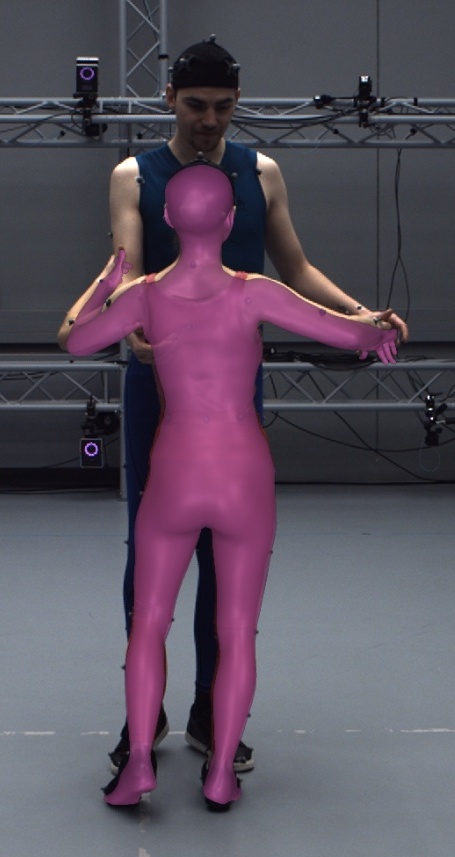} &
    \includegraphics[width=.14\linewidth]{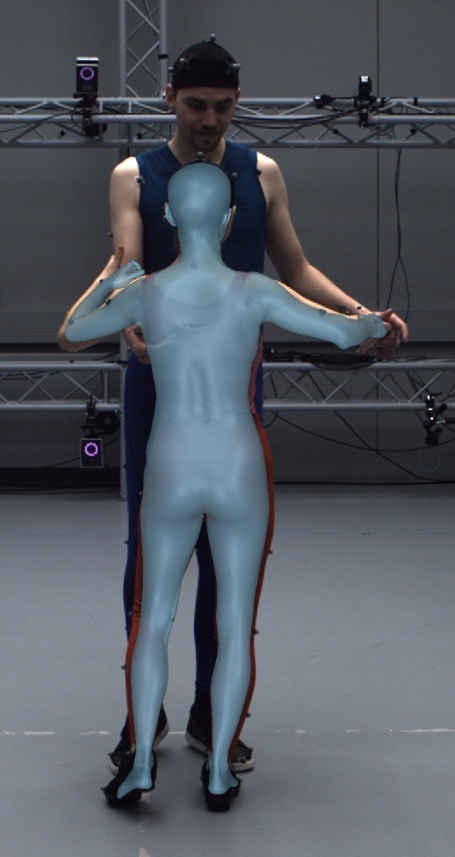} &
    \includegraphics[width=.14\linewidth]{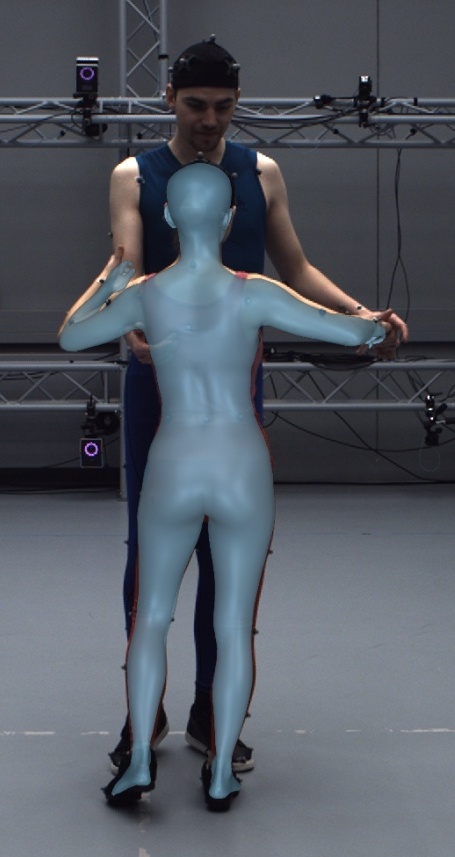} &
    \includegraphics[width=.14\linewidth]{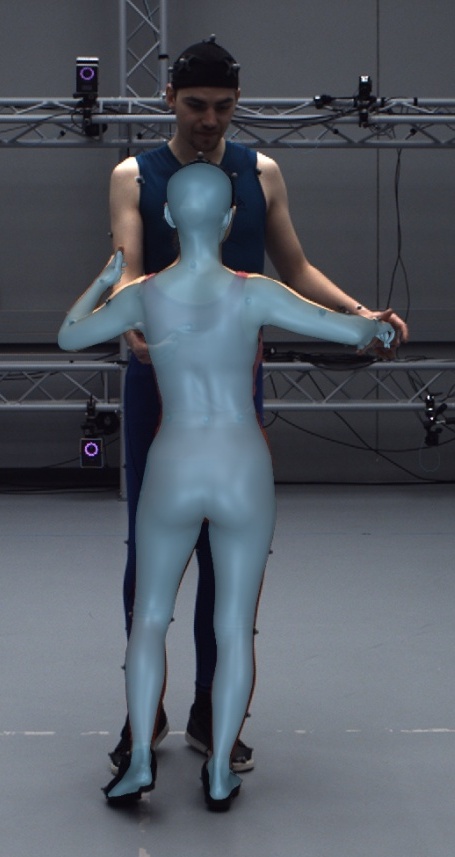} &
    \includegraphics[width=.14\linewidth]{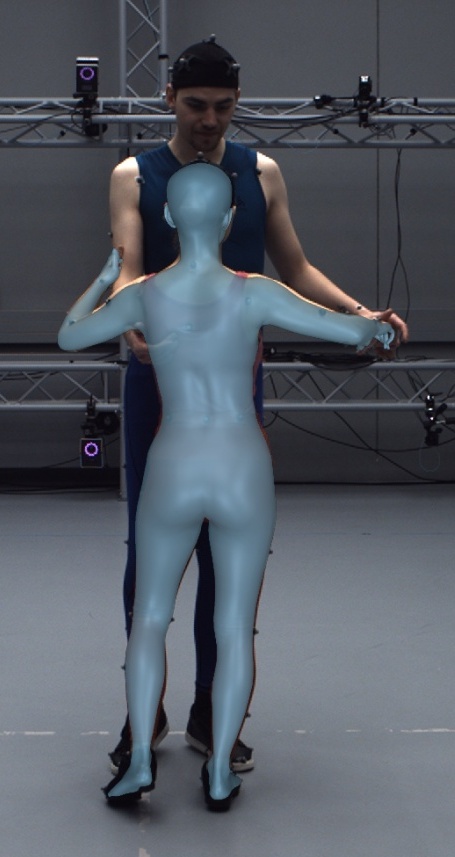}
\end{tabular}
}
\caption{Mesh fitting comparison. Top to bottom, MOYO, RICH, Harmony4D, \datasetEval-D. Notice how \model has better fit to the image than the GT body shape (first, third and fourth row). Also the hand contact on the fourth.}
\label{fig:mesh_comp_cropped}
\end{figure}

As a proxy for ground truth for our \datasetEval datasets, we use Vicon markers fit using MoSh++ \cite{amass}, and for all methods, we predict pose, shape and global translation.
\cref{tab:results_without_vtemplate} reports the MPJPE for each of the 55 SMPL-X joints along with the PVE.
We report the error for the body and hands separately in SupMat.
\model, even without optimizing contact achieves better result than previous methods. However, human reconstruction metrics do not tell the whole story as the ground truth presents some interpenetration which can affect the score. For this, we computed the average depth penetration and number of penetrating vertices across the interaction datasets. In \cref{tab:penetration_eval} we can see that \model results in bodies that have less penetration, indicating that overall \model-C achieves the best results. See \cref{fig:mesh_comp_cropped} and \supmat for more results and performance using different numbers of cameras and optimization stages.
\begin{table}
\centering
\caption{Mean Penetration (M.P.) depth (mm) and vertices on Harmony4D, CHI3D, and MammaEval-D.
}
\resizebox{0.65\linewidth}{!}{%
\begin{tblr}{
  row{1} = {c},
  cell{2}{2} = {c},
  cell{2}{3} = {c},
  cell{3}{2} = {c},
  cell{3}{3} = {c},
  cell{4}{2} = {c},
  cell{4}{3} = {c},
  cell{5}{2} = {c},
  cell{5}{3} = {c},
  cell{6}{2} = {c},
  cell{6}{3} = {c},
  hline{1-2,5,7} = {-}{},
}
Model                     & M.P.Depth & M.P.Verts. \\
Look-Ma*                  & 13.73        & 990.61 \\
CameraHMR                 & 13.41        & 678.10 \\
Ground truth              & 9.84        & 479.94 \\
\model (ours)           & 10.50        & 456.05 \\
\model-C (ours) &\textbf{ 8.46}      & \textbf{378.02}
\end{tblr}
}
\label{tab:penetration_eval}
\end{table}

We also noticed that optimizing for floor contact does not improve the results,
suggesting that our network landmarks are already robust enough without it.

\begin{table*}[!htbp]
\centering
\caption{Benchmark 3D fitting errors (mm).}
\resizebox{0.8\linewidth}{!}{%
\begin{tblr}{
  row{1} = {c},
  row{2} = {c},
  row{8} = {c},
  row{14} = {c},
  cell{1}{1} = {r=2}{},
  cell{1}{2} = {c=2}{},
  cell{1}{4} = {c=2}{},
  cell{1}{6} = {c=2}{},
  cell{1}{8} = {c=2}{},
  cell{1}{10} = {c=2}{},
  cell{1}{12} = {c=2}{},
  cell{3}{2} = {c},
  cell{3}{3} = {c},
  cell{3}{4} = {c},
  cell{3}{5} = {c},
  cell{3}{6} = {c},
  cell{3}{7} = {c},
  cell{3}{8} = {c},
  cell{3}{9} = {c},
  cell{3}{10} = {c},
  cell{3}{11} = {c},
  cell{3}{12} = {c},
  cell{3}{13} = {c},
  cell{4}{2} = {c},
  cell{4}{3} = {c},
  cell{4}{4} = {c},
  cell{4}{5} = {c},
  cell{4}{6} = {c},
  cell{4}{7} = {c},
  cell{4}{8} = {c},
  cell{4}{9} = {c},
  cell{4}{10} = {c},
  cell{4}{11} = {c},
  cell{4}{12} = {c},
  cell{4}{13} = {c},
  cell{5}{2} = {c},
  cell{5}{3} = {c},
  cell{5}{4} = {c},
  cell{5}{5} = {c},
  cell{5}{6} = {c},
  cell{5}{7} = {c},
  cell{5}{8} = {c},
  cell{5}{9} = {c},
  cell{5}{10} = {c},
  cell{5}{11} = {c},
  cell{5}{12} = {c},
  cell{5}{13} = {c},
  cell{6}{2} = {c},
  cell{6}{3} = {c},
  cell{6}{4} = {c},
  cell{6}{5} = {c},
  cell{6}{6} = {c},
  cell{6}{7} = {c},
  cell{6}{8} = {c},
  cell{6}{9} = {c},
  cell{6}{10} = {c},
  cell{6}{11} = {c},
  cell{6}{12} = {c},
  cell{6}{13} = {c},
  cell{7}{2} = {c},
  cell{7}{3} = {c},
  cell{7}{4} = {c},
  cell{7}{5} = {c},
  cell{7}{6} = {c},
  cell{7}{7} = {c},
  cell{7}{8} = {c},
  cell{7}{9} = {c},
  cell{7}{10} = {c},
  cell{7}{11} = {c},
  cell{7}{12} = {c},
  cell{7}{13} = {c},
  hline{1,3,8} = {-}{},
  hline{2} = {2-13}{},
}
Model       & RICH           &                & Harmony4D      &                & CHI3D          &                & \datasetEval-S    &                & \datasetEval-D      &                & MOYO           &                \\
            & MPJPE          & PVE            & MPJPE          & PVE            & MPJPE          & PVE            & MPJPE          & PVE            & MPJPE          & PVE            & MPJPE          & PVE            \\
SMPLfiy     & 96.18          & 71.42          & -              & -              & 67.68          & 51.79          & 47.15          & 35.42          & 53.92          & 43.08          & 62.15          & 44.68          \\
LookMa*     & 39.52          & 30.29          & 59.37          & 45.6           & 46.47          & 39.36          & 25.97          & 23.94          & 27.98          & 24.89          & 60.15          & 53.82          \\
CameraHMR~  & 25.61          & 21.36          & 58.59          & 42.0           & 40.8           & 34.61          & 15.25          & 18.43          & 20.41          & 21.06          & 33.75          & 33.74          \\
\model & 22.20          & 19.76          & \textbf{45.26} & \textbf{34.02} & 38.01          & 32.84          & 12.96          & 17.18          & \textbf{17.71} & 19.80          & 22.95          & 25.48          \\
\model-C     & \textbf{22.20} & \textbf{19.76} & 45.35          & 34.05          & \textbf{37.96} & \textbf{32.82} & \textbf{12.96} & \textbf{17.18} & 17.73          & \textbf{19.78} & \textbf{22.95} & \textbf{25.48}
\end{tblr}
}
\vspace{-0.12in}
\label{tab:results_without_vtemplate}
\end{table*}

Additionally, we compare \model on the Hi4D dataset.
For a fair evaluation, we retrained our network and removed the sequences that used Hi4D.
The dataset is in SMPL format, so we transform our SMPL-X mesh to SMPL, and regress 19 joints following \citet{lu2024avatarpose}.
\cref{tab:hi4d_comp} shows that our method outperforms previous methods by a large margin, even though our network was not trained on any sequence of Hi4D. In \supmat we compare against Harmony4D and show that our method produces better fits.
\begin{table}[t]
\centering
\caption{MPJPE on the Hi4D dataset for 19 SMPL joints.}
\resizebox{0.65\linewidth}{!}{%
\begin{tabular}{l c}
\hline
Method & MPJPE (mm) \\
\hline
MvP~\cite{zhang2021direct}              & 92.77 \\
Graph~\cite{wu2021graph}          & 89.62 \\
Faster VoxelPose~\cite{ye2022faster} & 68.40 \\
MVPose*~\cite{dong2019fast}  & 53.05 \\
MVPose~\cite{dong2021fast}        & 42.63 \\
4DAssociation~\cite{zhang20204d} & 41.29 \\
AvatarPose \cite{lu2024avatarpose} & 32.10 \\
\hline
\model-C (Ours) & \textbf{12.44} \\
\hline
\end{tabular}
}
\label{tab:hi4d_comp}
\end{table}

We do not compare with pure feed-forward approaches because they need to be retrained for each multi-view configuration.
We also do not compare with single-frame camera methods because they can not recover the global translation and rotation of the mesh.

\subsection{VICON Markers Comparison}
\label{sec:heldout}

But how do the results match up against those of Vicon?
Vicon claims sub-millimeter accuracy in locating markers.
However, the estimated body is subject to noise from soft-tissue artifacts and marker misplacement, and it is the accuracy of this inferred model that is important.
To evaluate this, we capture \datasetEval-Extra with a marker-based system and fit SMPL-X to a standard Vicon marker set (FrontWaist10Fingers) using MoSh++~\citep{amass}.
The dataset consist of 3 subject, 4 sequences with 37 additional markers which are not used by MoSh++.
Using the \mbox{SMPL-X} bodies estimated by MoSh++ and \model, we predict the held-out marker locations and compute the average 3D Euclidean distance error.
The Vicon+MoSh++ method has an error of 21.619mm, while our method has an error of 22.481mm; see the website for a video with side-by-side comparisons.

Our takeaway is that there is only a minimal, 0.862mm difference between our method and the process used to construct the AMASS dataset, which has been widely used for many tasks.
This suggests that our markerless approach is sufficiently accurate to replace the traditional pipeline. Our method reduces the full capture-to-SMPL-X pipeline from $\sim72$ hours to $~26$ hours on a consumer GPU—making it nearly three times faster than the manual marker-based workflow. See \supmat for details. Our method also generalizes to consumer-grade capture setups; see the \supmat for results using four iPhones.

\section{Conclusions}
We have demonstrated a novel architecture for multi-view markerless capture of human motion in SMPL-X format.
We address the challenging case of close multi-person interaction through algorithm design choices and novel datasets.
\model is able to predict dense 2D landmarks robustly even with extreme poses and occlusion using novel per-landmark learnable queries.
We provide a unique quantitative comparison between \model and the gold-standard marker-based solution
and find that our approach achieves competitive reconstruction quality with significantly lower cost.
\newpage
{\small\noindent\textbf{Acknowledgments.} 
We thank Claudia Gallatz for trial coordination, Senya Polikovsky and Gökce Ergün for technical support with the captures, Benjamin Pellkofer for IT support and Peter Kulits for thoughtful feedback and support. We also thank Tomasz Niewiadomski, Valerian Fourel, Arina Kuznetcova, Tithi Rakshit, Suraj Bhor, Alpar Cseke and Florentin Doll for cleaning the Vicon marker data. Finally, we thank Berna Kabadayi, Shrisha Bharadwaj, and Francesca Romana Blanda for their assistance with data capture.
While Michael J.~Black is an employee of Epic Games, his research in this project was performed solely at, and funded solely by, the Max Planck Society.}
{
    \small
    \bibliographystyle{ieeenat_fullname}
    \bibliography{main}
}

\clearpage

\section*{Supplementary Material}
The \supvid shows a side-by-side comparison of our method vs the ground truth data.
In the video we intentionally did not label which result is which.
The answer is at the end of this document in Section \ref{sec:reveal}.
We ask the reader to view the video before checking the answer.  Please refer to \url{https://mamma.is.tue.mpg.de/} for the \supvid.

\section{Landmark Networks Training Details}
Our network, CameraHMR~\cite{patel2024camerahmr}, and LookMa*\cite{hewitt2024look} are trained for 300K iterations using 4 NVIDIA A100 GPUs, with a per-GPU batch size of 24 and gradient accumulation set to 2 steps. The training time took  around 3 days. We use the Adam optimizer with 500 warm-up iterations.
For \net and CameraHMR, we initialize the transformer backbone with ViTPose-B weights pretrained at a resolution of $256\times192$.
Following \citet{zhang2022dino}, we interpolate the positional embeddings to increase the effective input resolution to $512\times384$.
\citet{hewitt2024look} use HRNet-W48 as the backbone.
We follow their configuration, except we initialize the network using pose HRNet weights trained on the COCO dataset.
The input resolution is set to $512\times384$, consistent with the transformer-based models. As the other models do not predict contact, we trained a version of our model that only predicts uncertainty and visibility.

\section{\dataset Dataset}
Each rendered sequence contains 2--6 synthetic subjects, placed at random positions within the capture volume.
They are randomly assigned one of 100 skin textures which are overlaid with one of 1.7K garment textures.

In total, we render 2.8k sequences (5.5 hours). Each sequence is rendered from 8 views, where the first view is randomly selected and the rest via the Farthest Point Sampling (FPS) algorithm on our 32-camera setup, maximizing spatial coverage and projected-pose diversity.
For lighting and background variety, we use a pool of 95 HDR images.
We render at 6 fps, 2056$\times$1504px, roughly twice the resolution of BEDLAM, to ensure much finer detail in the hand regions. in \cref{tab:mammasyn_composition} we show the dataset composition by number of images.

\begin{figure}[htp]
\centering
\resizebox{\linewidth}{!}{%
\begin{tabular}{ccc}
    \includegraphics[width=0.32\linewidth]{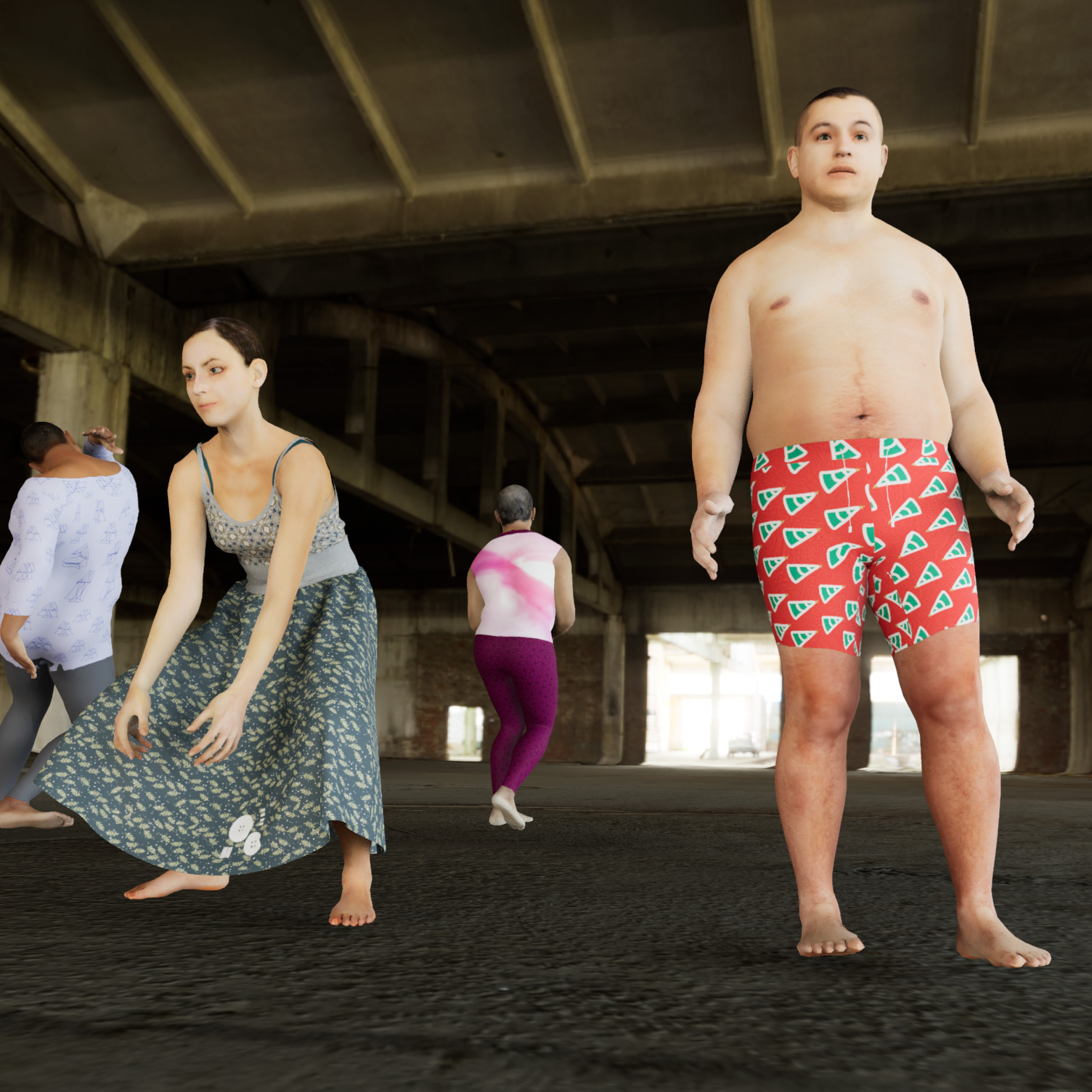} &
    \includegraphics[width=0.32\linewidth]{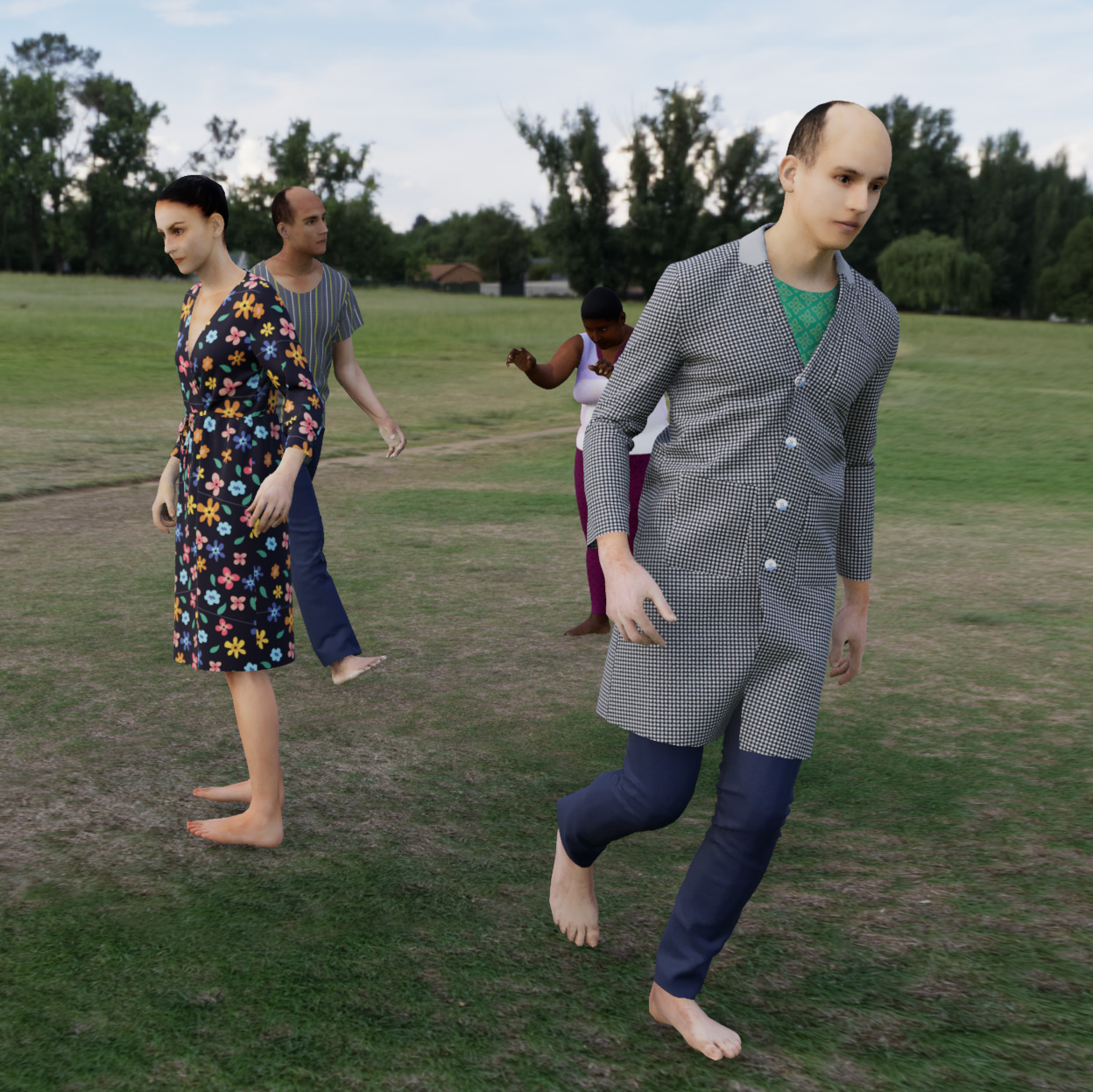} &
    \includegraphics[width=0.32\linewidth]{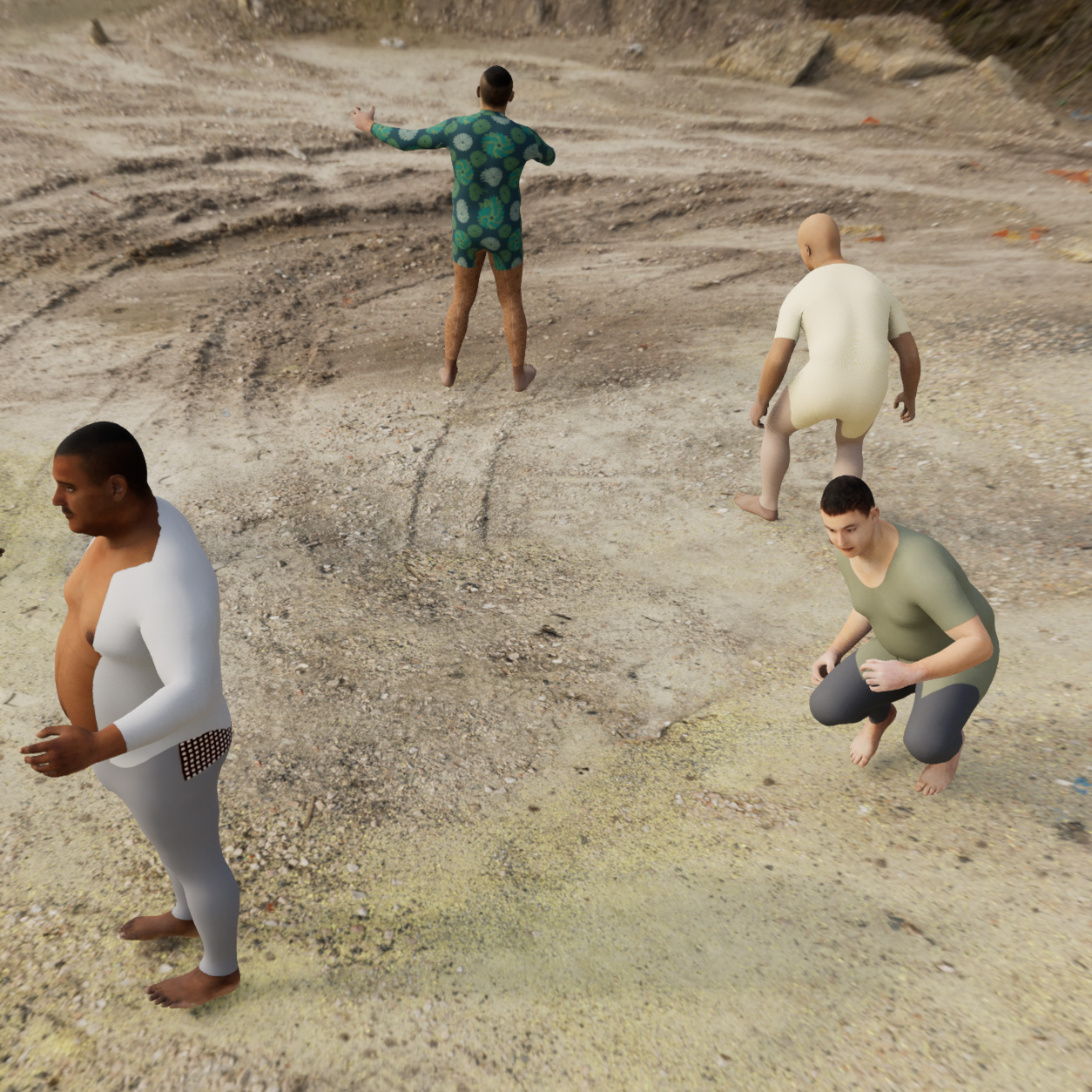} \\[0.03cm]
    \includegraphics[width=0.32\linewidth]{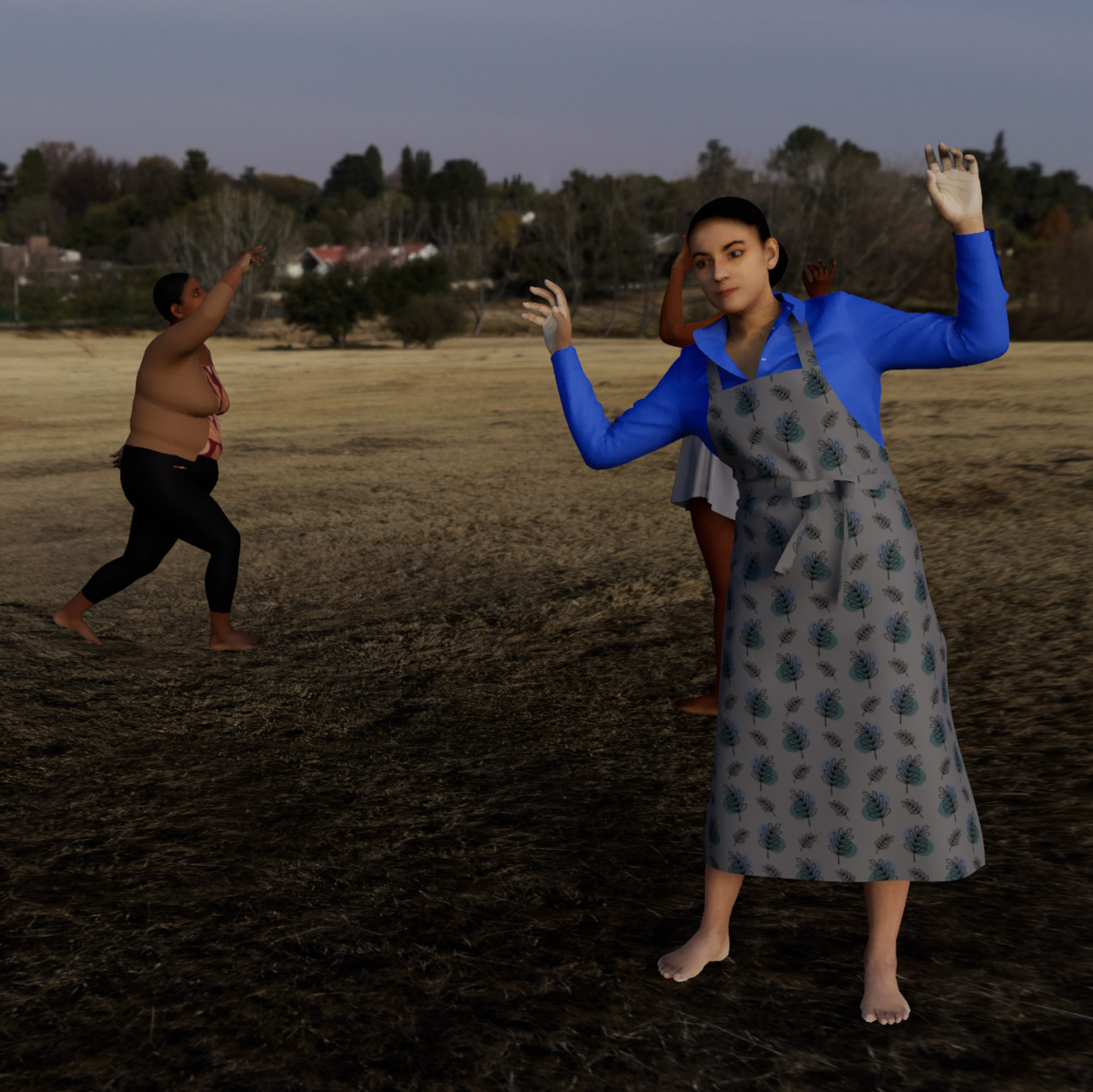} &
    \includegraphics[width=0.32\linewidth]{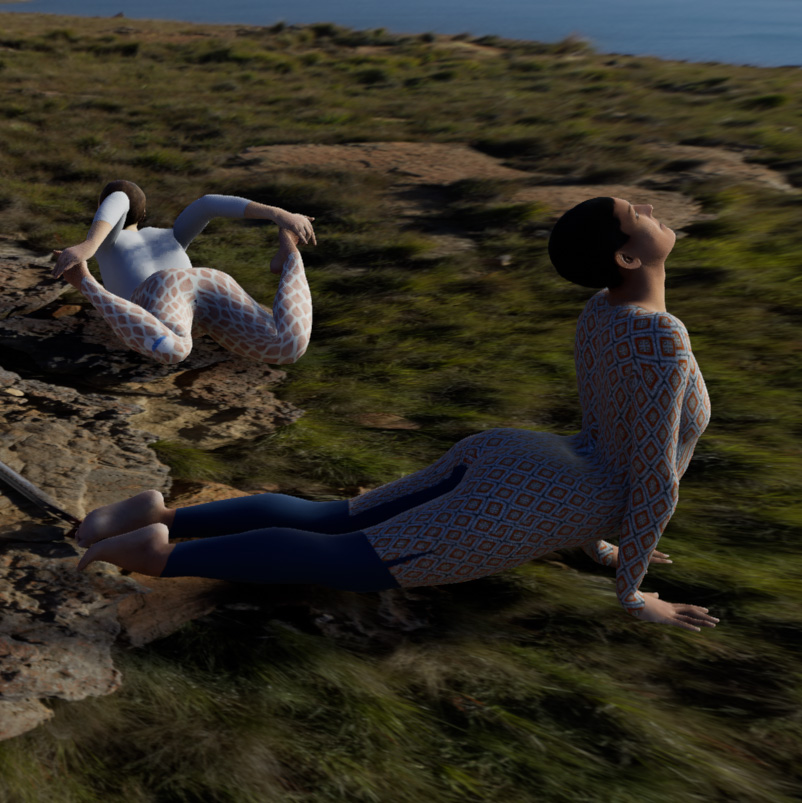} &
    \includegraphics[width=0.32\linewidth]{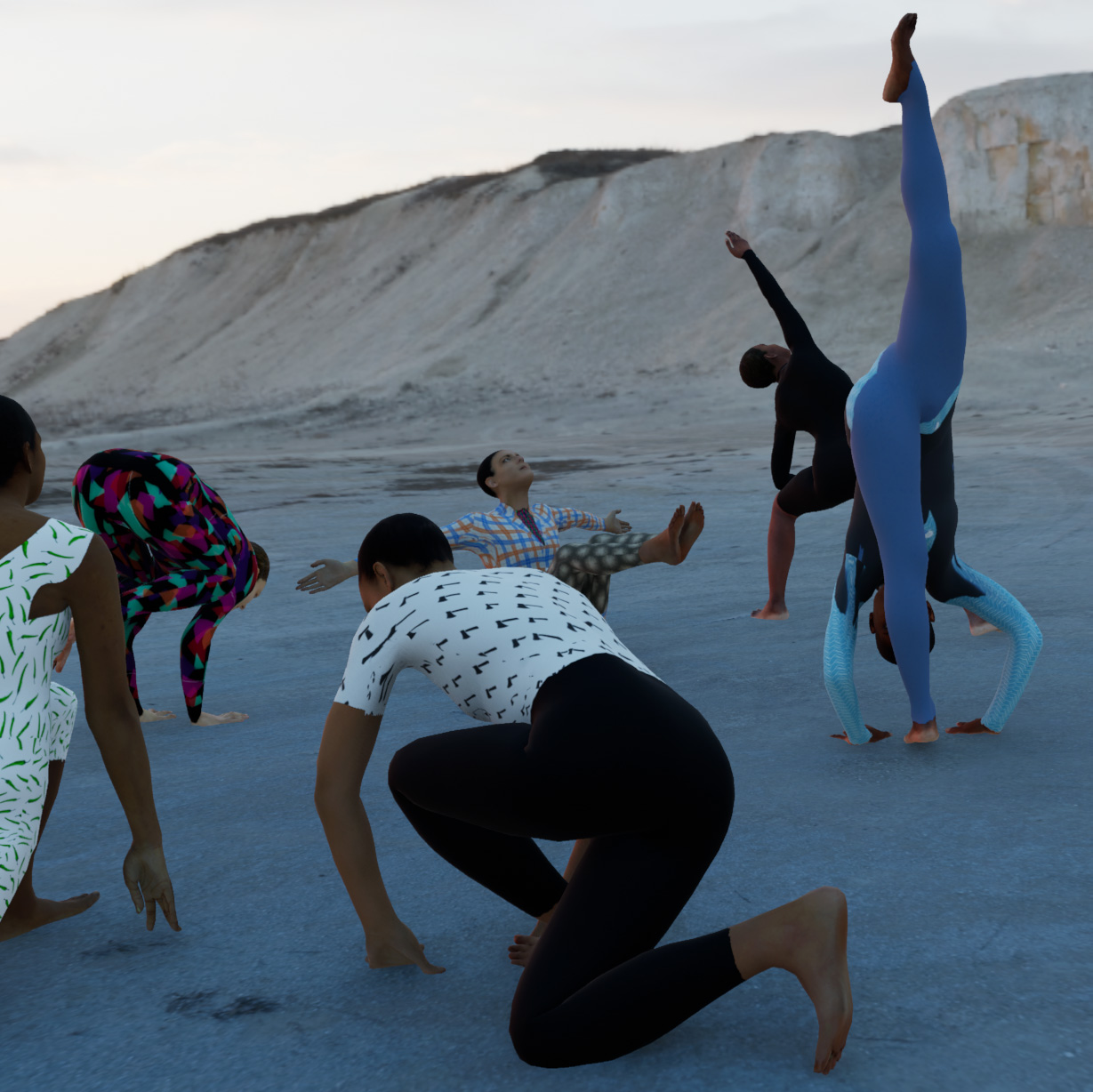} \\[0.03cm]
    \includegraphics[width=0.32\linewidth]{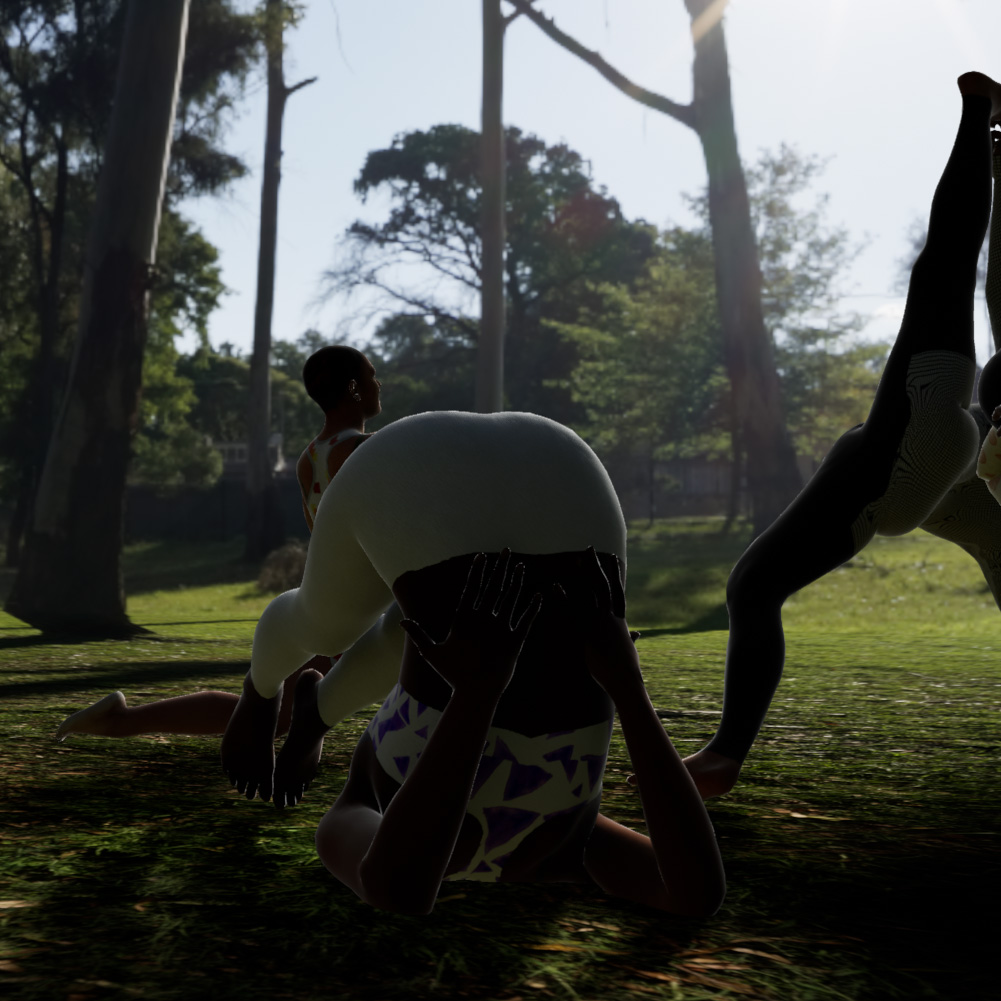} &
    \includegraphics[width=0.32\linewidth]{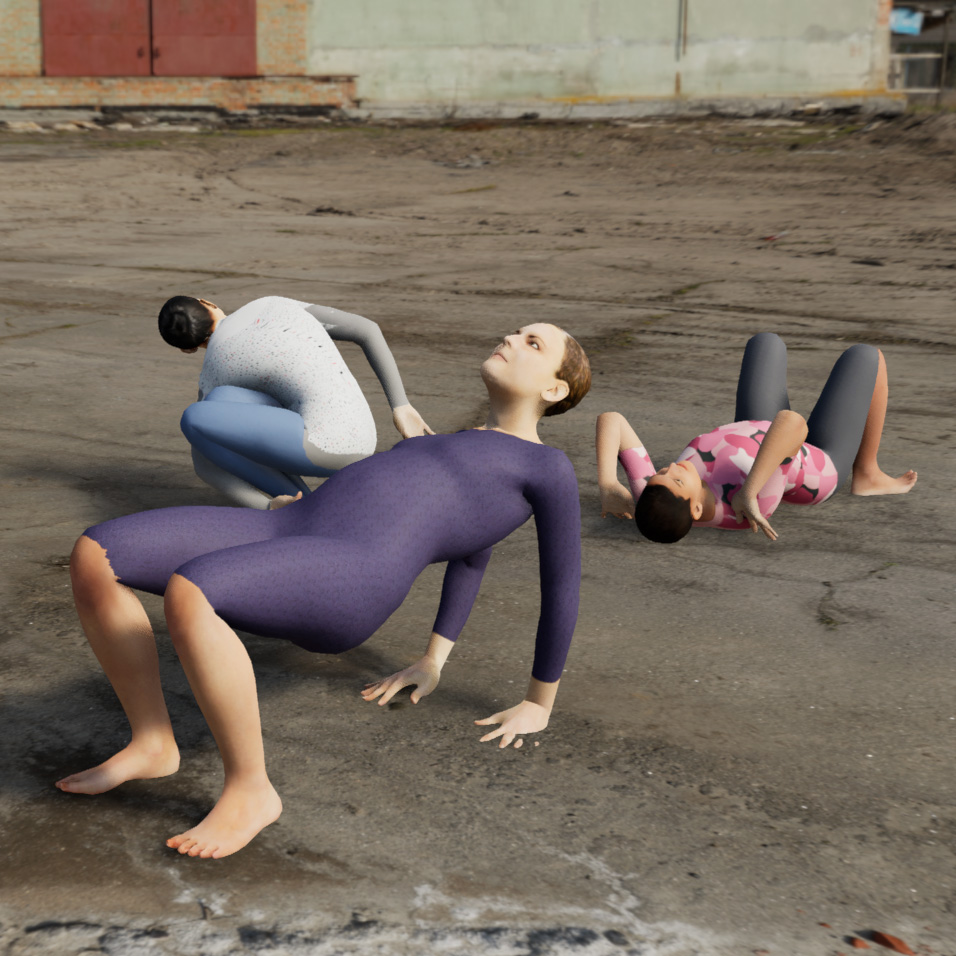} &
    \includegraphics[width=0.32\linewidth]{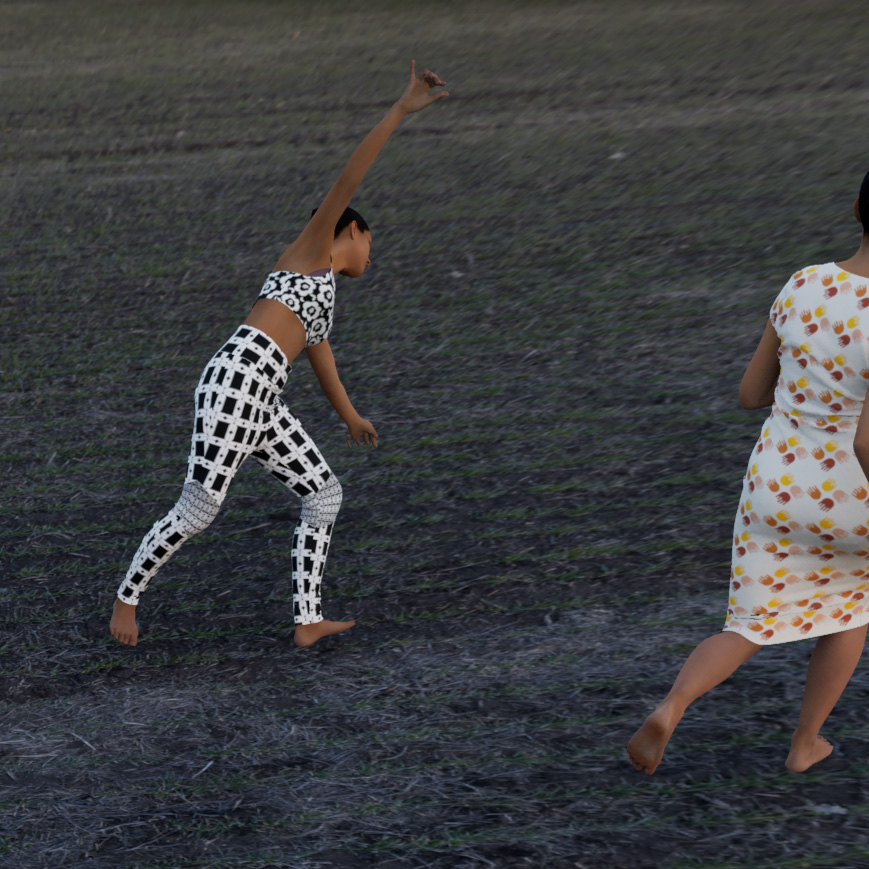} 
\end{tabular}}

\caption{Crops from \dataset-S (Single person).}
\label{fig:extra_vis}
\end{figure}

\begin{figure}[htp]
\centering
\resizebox{\linewidth}{!}{
\begin{tabular}{ccc}
    \includegraphics[width=0.32\linewidth]{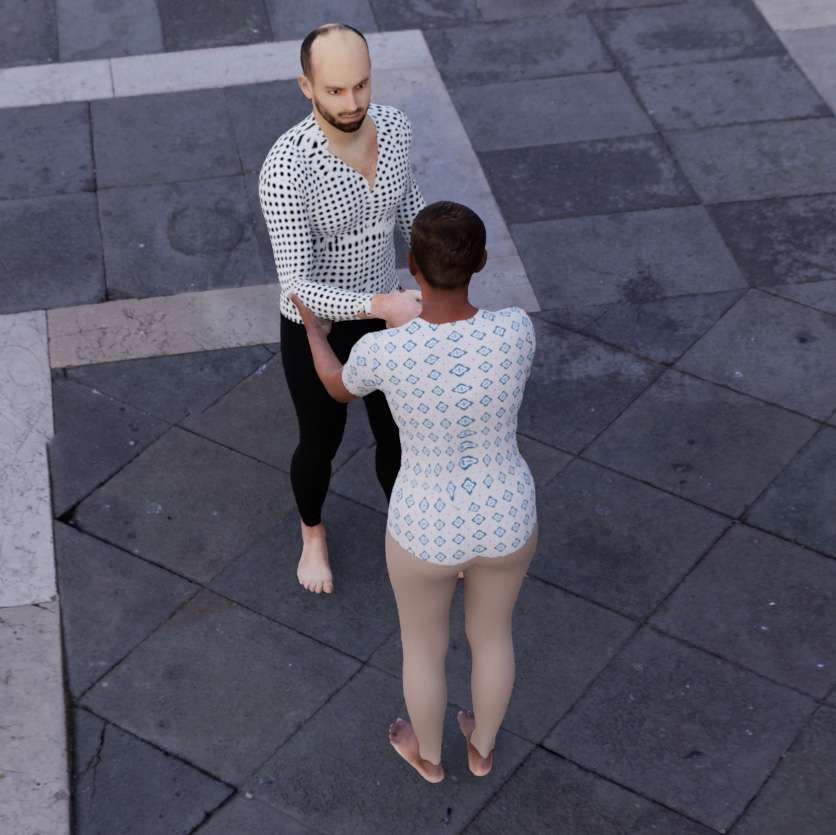} &
    \includegraphics[width=0.32\linewidth]{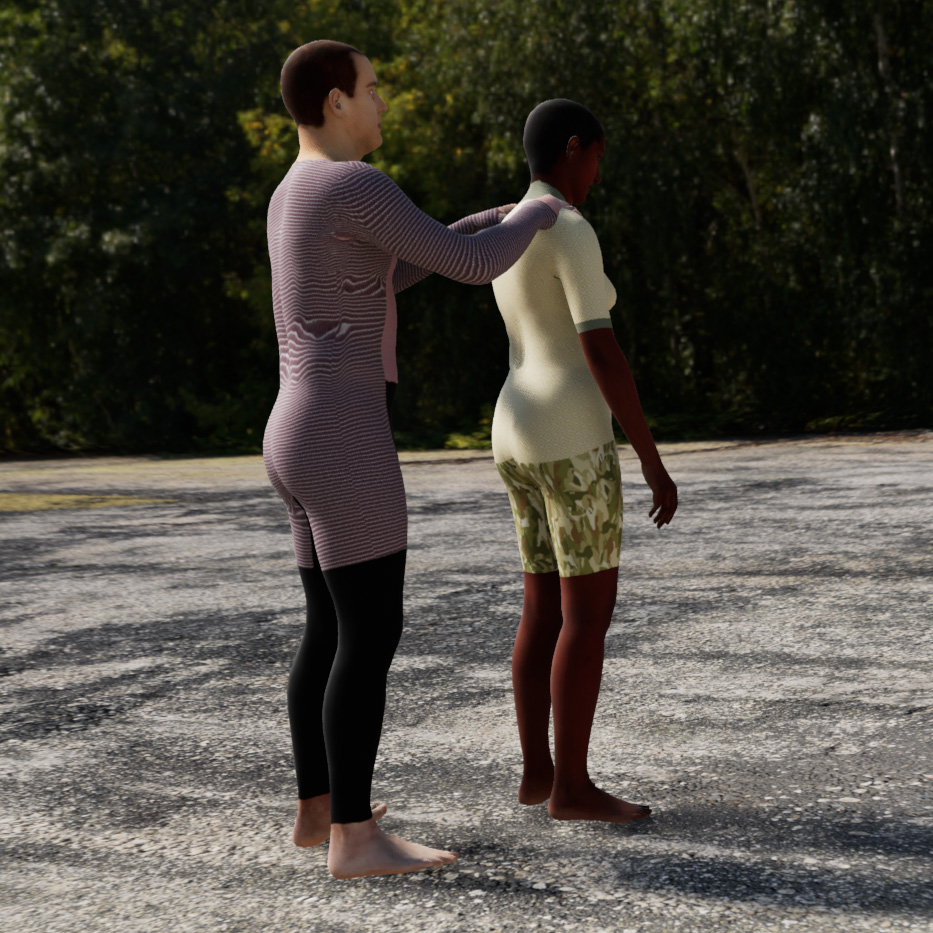} &
    \includegraphics[width=0.32\linewidth]{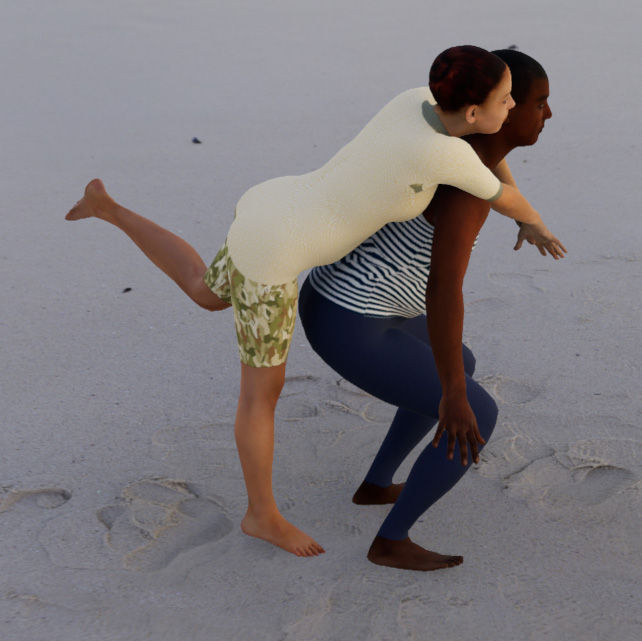} \\[0.03cm]
    \includegraphics[width=0.32\linewidth]{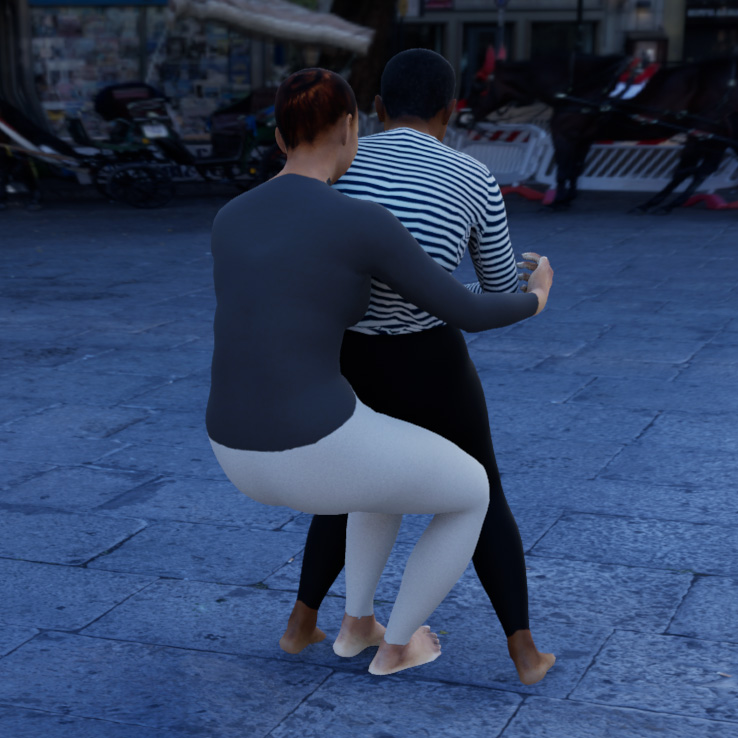} &
    \includegraphics[width=0.32\linewidth]{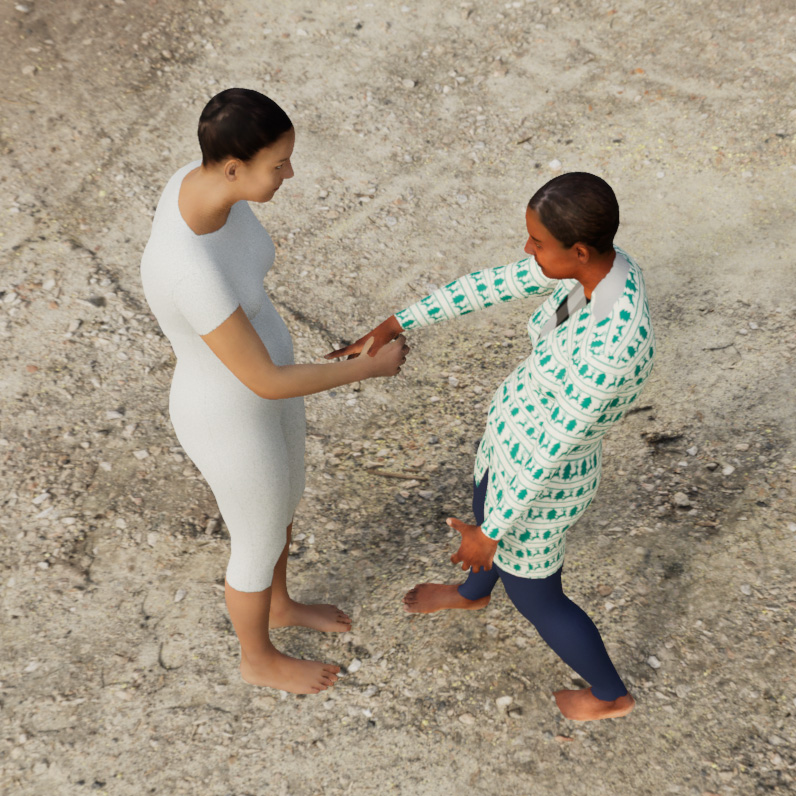} &
    \includegraphics[width=0.32\linewidth]{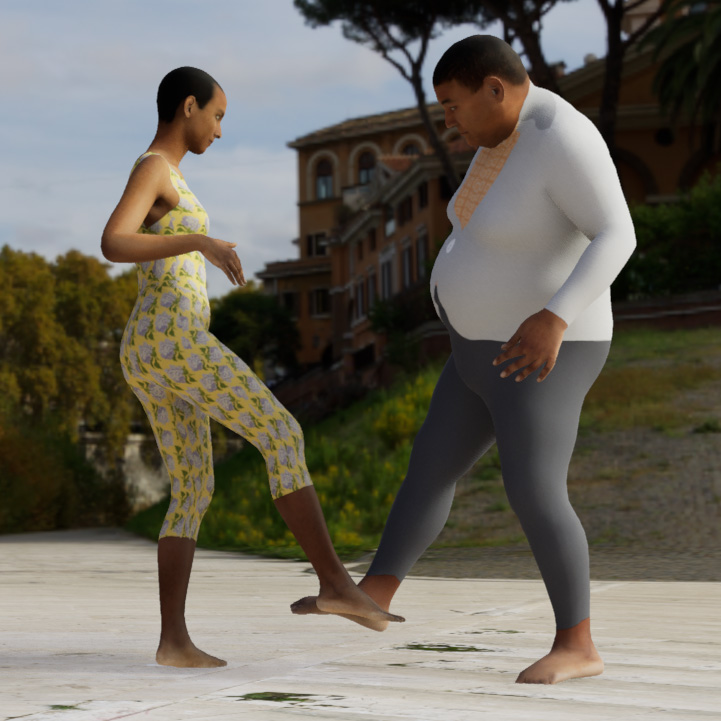} \\[0.03cm]
    \includegraphics[width=0.32\linewidth]{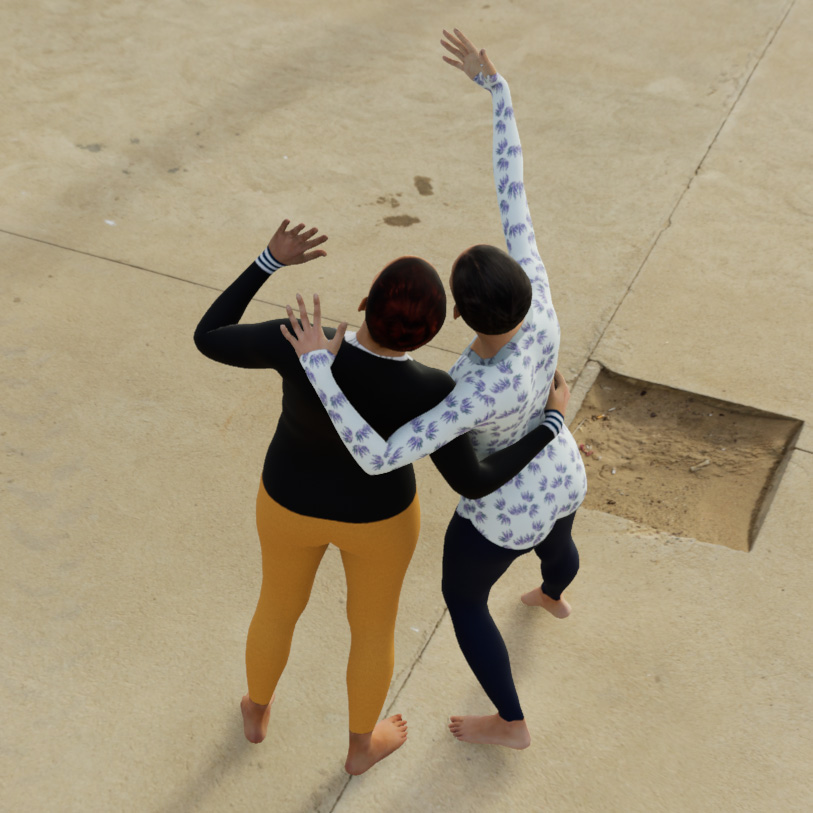} &
    \includegraphics[width=0.32\linewidth]{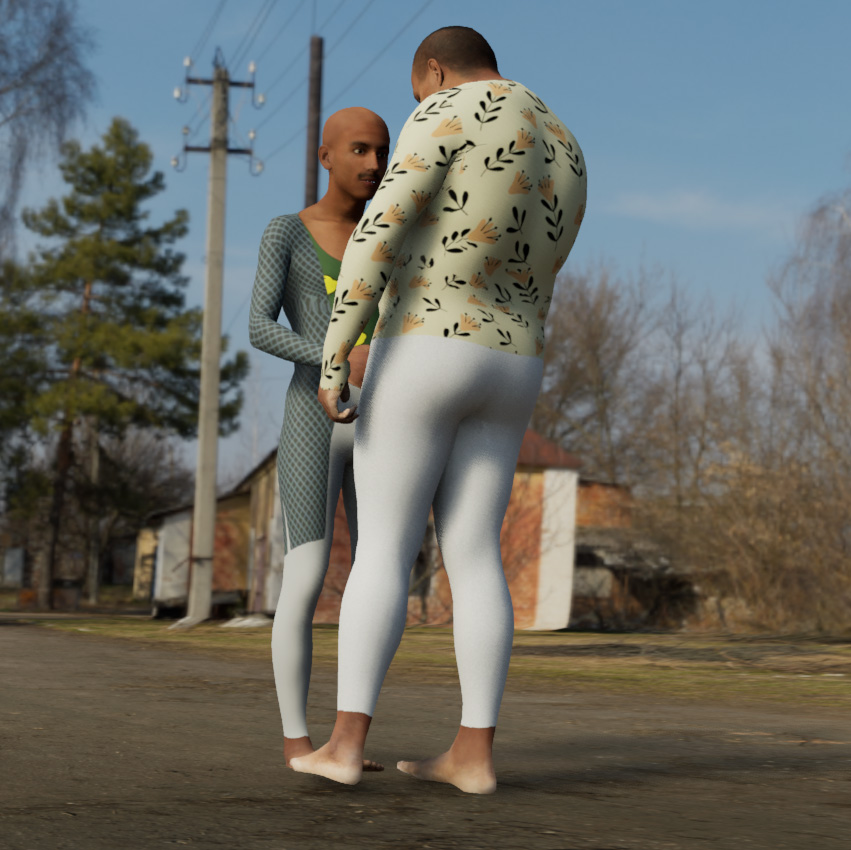} &
    \includegraphics[width=0.32\linewidth]{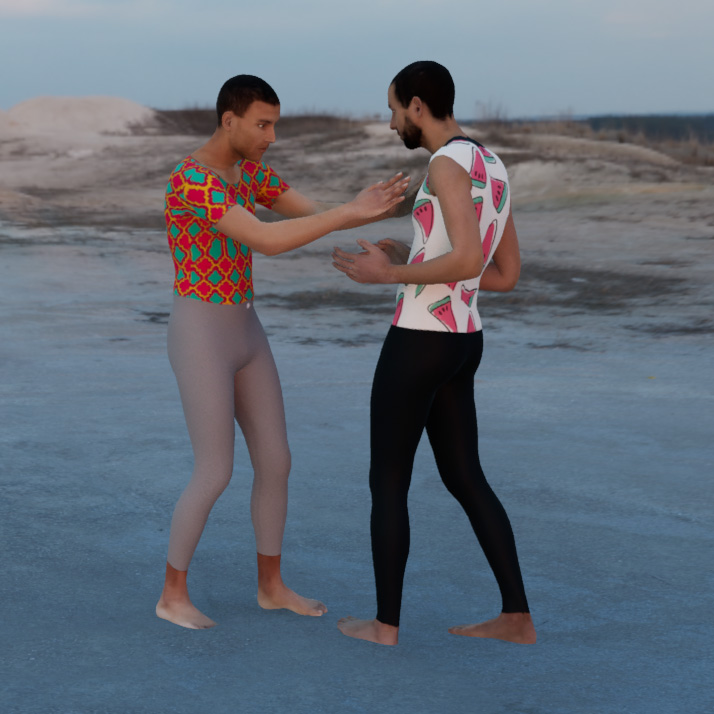} 
\end{tabular}}

\caption{Crops from \dataset-I (Interactions).}
\label{fig:extra_vis_interaction}
\end{figure}

\begin{figure}[htbp]
    \centering
    \begin{minipage}[b]{0.49\linewidth}
        \centering
        \includegraphics[width=\linewidth]{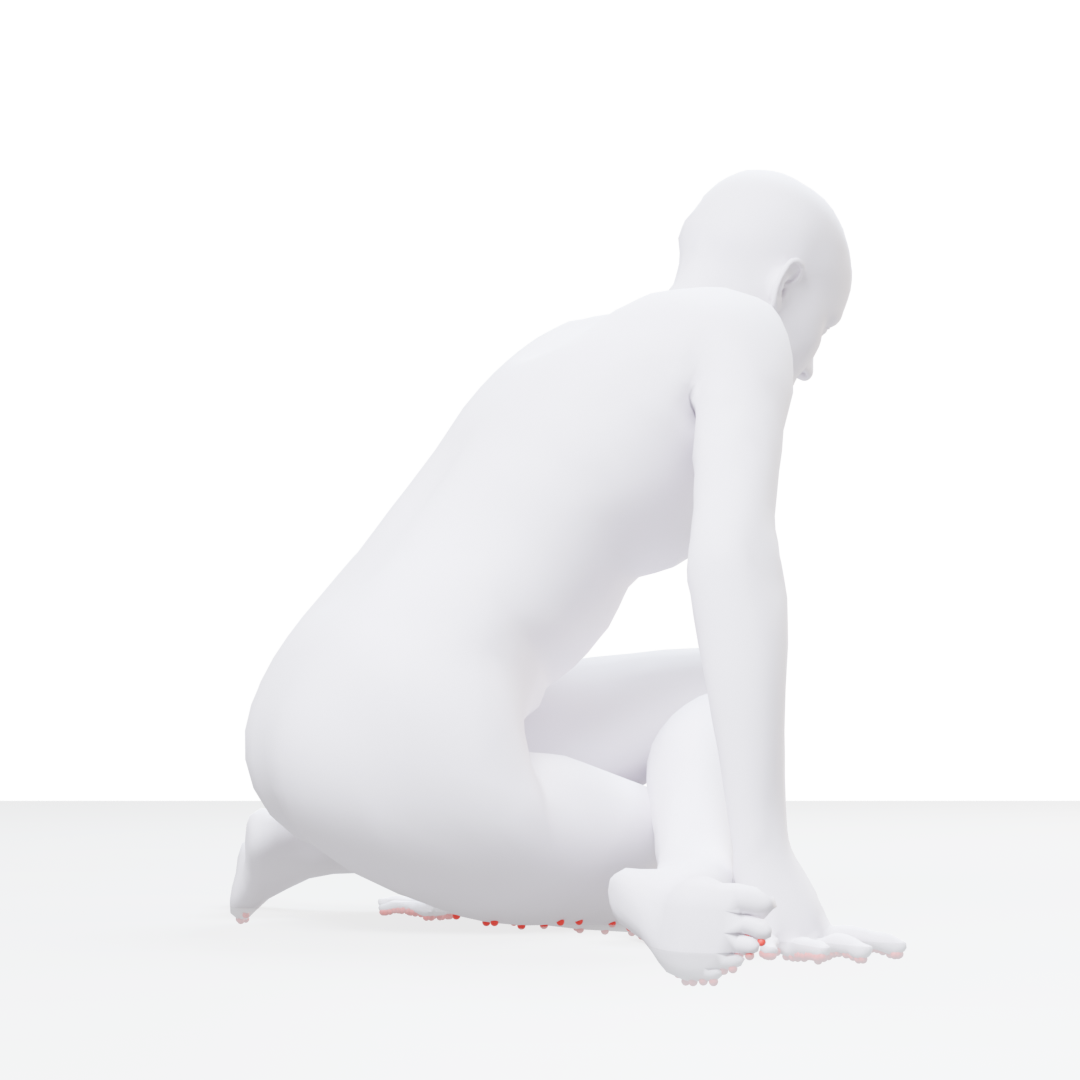}
    \end{minipage}
    \hfill
    \begin{minipage}[b]{0.49\linewidth}
        \centering
        \includegraphics[width=\linewidth]{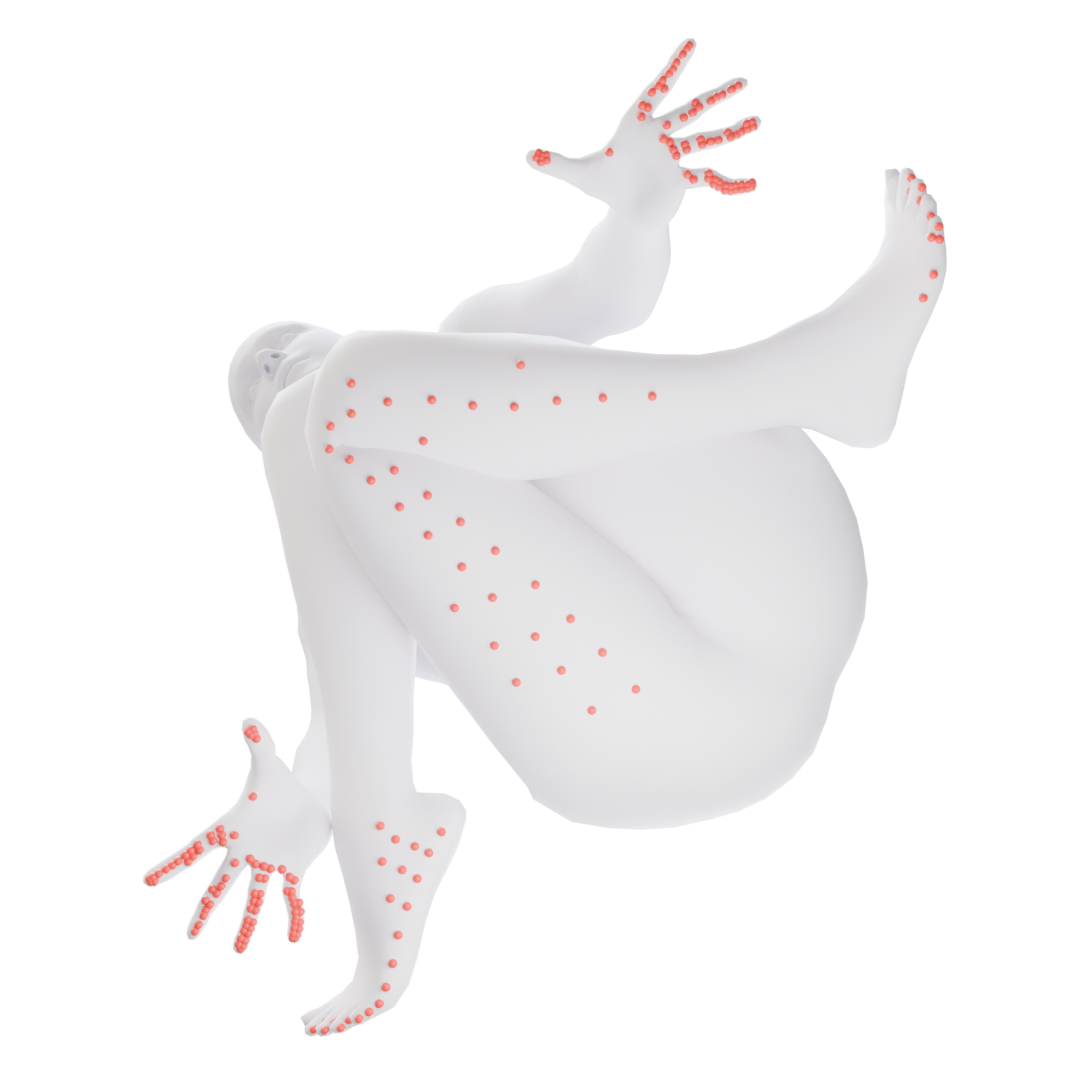}
    \end{minipage}
    \caption{Computed floor contact from geometric cues, used in \dataset. We show the pose from a side-view and an under-view. SMPL-X vertices that are in contact are shown in red.}
    \label{fig:moyo_floor_contact}
\end{figure}

\begin{figure}[t]
    \centering
    \includegraphics[width=\linewidth]{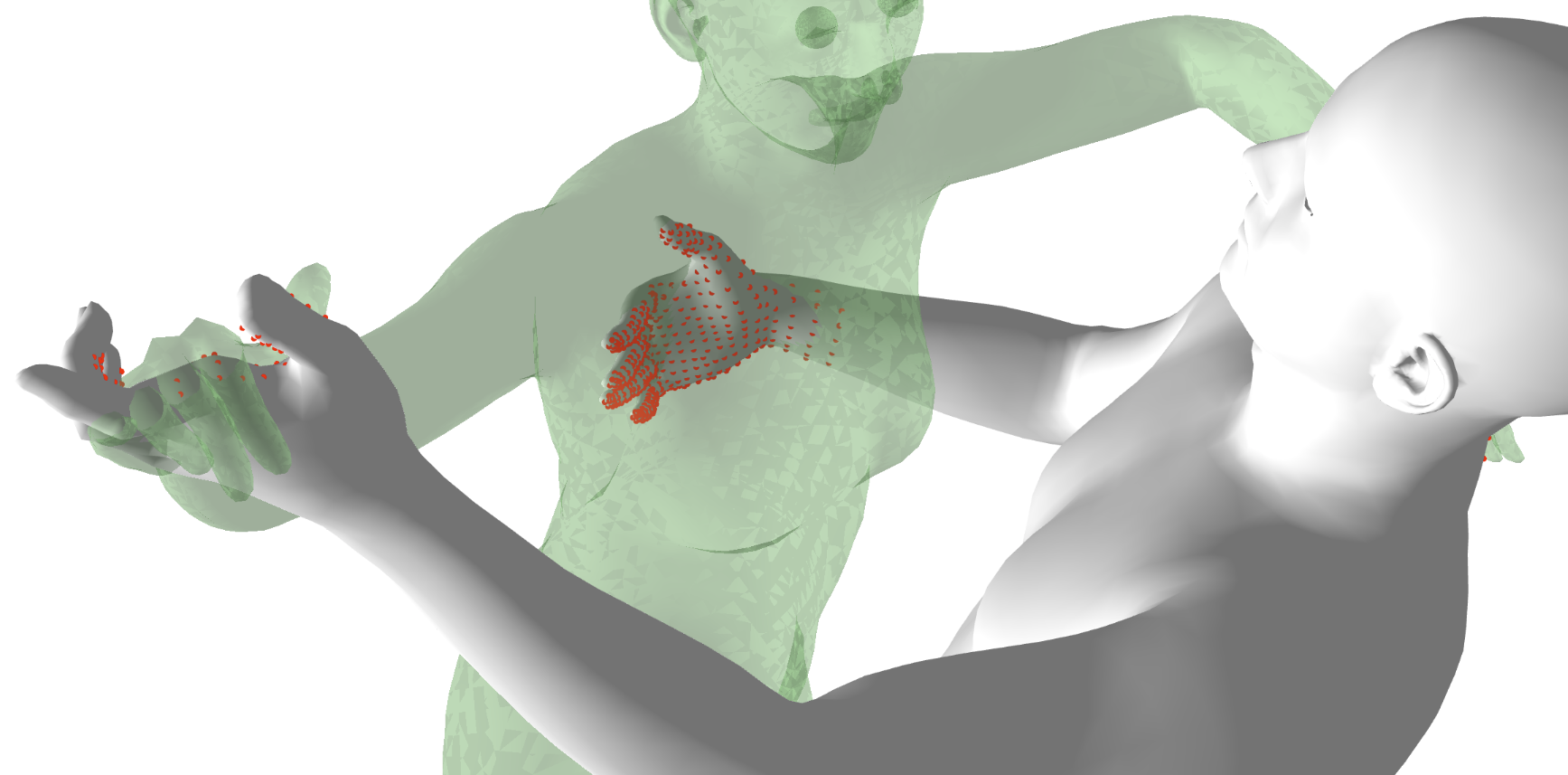}

    \caption{Computed contact between two subjects from geometric cues, used in \dataset. SMPL-X vertices of gray subject that are in contact with the green transparent subject are shown in red.}
    \label{fig:mammasyn_interactions_contact}
\end{figure}

In \cref{fig:extra_vis} and \cref{fig:extra_vis_interaction}, we show more cropped samples of our dataset, and in \cref{fig:moyo_floor_contact} and \cref{fig:mammasyn_interactions_contact}, we show the contact labels.

\begin{table}[htb]
    \centering
    \resizebox{0.9\linewidth}{!}{%
    \begin{tblr}{
      colspec={c c l c}, 
      row{1} = {c},
      column{1,2,4} = {c},
      hline{1-2,4,9,Z} = {-}{}
    }
    Subset      & \# minutes & Datasets                 & \% of images \\
    \dataset-S  & 86         & BEDLAM                   & 9.9\%        \\
                &            & MOYO                     & 15.9\%       \\
    \dataset-I  & 211        & Harmony4D                & 17.7\%       \\
                &            & Hi4D                     & 8.0\%        \\
                &            & Inter-X                  & 14.6\%       \\
                &            & InteractionsCouple (ours)& 19.4\%      \\
                &            & LatinDance (ours)        & 3.4\%        \\
    \dataset-H  & 36         & InterHand2.6M            & 10.0\%       \\
                &            & SignAvatars              & 1.0\%        \\
    \end{tblr}
    }
    \caption{Composition of the \dataset dataset. Percentages are computed over all \dataset images.}
    \label{tab:mammasyn_composition}
\end{table}

\section{Dataset Evaluation}
\label{sec:dataset_eval}
We evaluate our proposed dataset, \dataset, on landmark prediction accuracy, particularly in cases involving occlusions caused by interactions.
We use the best-performing network, \net~+masks SAM2, pre-trained on BEDLAM (B), and train it for another 300K iterations on our dataset. We compared against the network trained only on BEDLAM but with an additional 300K iterations (BEDLAM*). We also train our network combining our dataset and BEDLAM (ALL+B).

\begin{table}[t]
\centering
\caption{\textbf{Evaluation of datasets.} Mean 2D Euclidean distance error (in pixels) between ground truth and predicted landmarks; includes visible and invisible landmarks.}
\resizebox{\linewidth}{!}{%
\begin{tblr}{
  column{even} = {c},
  column{3} = {c},
  column{5} = {c},
  column{7} = {c},
  hline{1-2,6} = {-}{},
}
Dataset                   & RICH  & Harmony4D & CHI3D & \datasetEval-S & \datasetEval-D & MOYO  \\
BEDLAM                    & 8.83  & 18.33     &   4.36  & 6.16 &7.70 &  11.04 \\
BEDLAM*                & 9.35  & 19.11     & 4.40  & 6.49     & 7.60   & 11.92  \\
\dataset            & 8.68  & 19.09     & 4.74  & 6.53     & 7.71   & 7.05  \\
\dataset+B          & \textbf{8.09}  & \textbf{17.34}     & \textbf{4.09}  & \textbf{5.62}     & \textbf{6.66}   & \textbf{6.95}  \\
\end{tblr}
}
\label{tab:dataset_ablation}
\end{table}

\cref{tab:dataset_ablation} shows that training longer with BEDLAM does not improve performance (in fact, it degrades slightly).
However, when we use our datasets along with BEDLAM, the performance of the network improves, especially on challenging poses such as those in MOYO and the interaction sequences.

\section{\datasetEval Dataset}
\begin{figure*}[!htbp]
\centering
    \begin{subfigure}{0.49\linewidth}
        \centering
        \includegraphics[width=\linewidth]{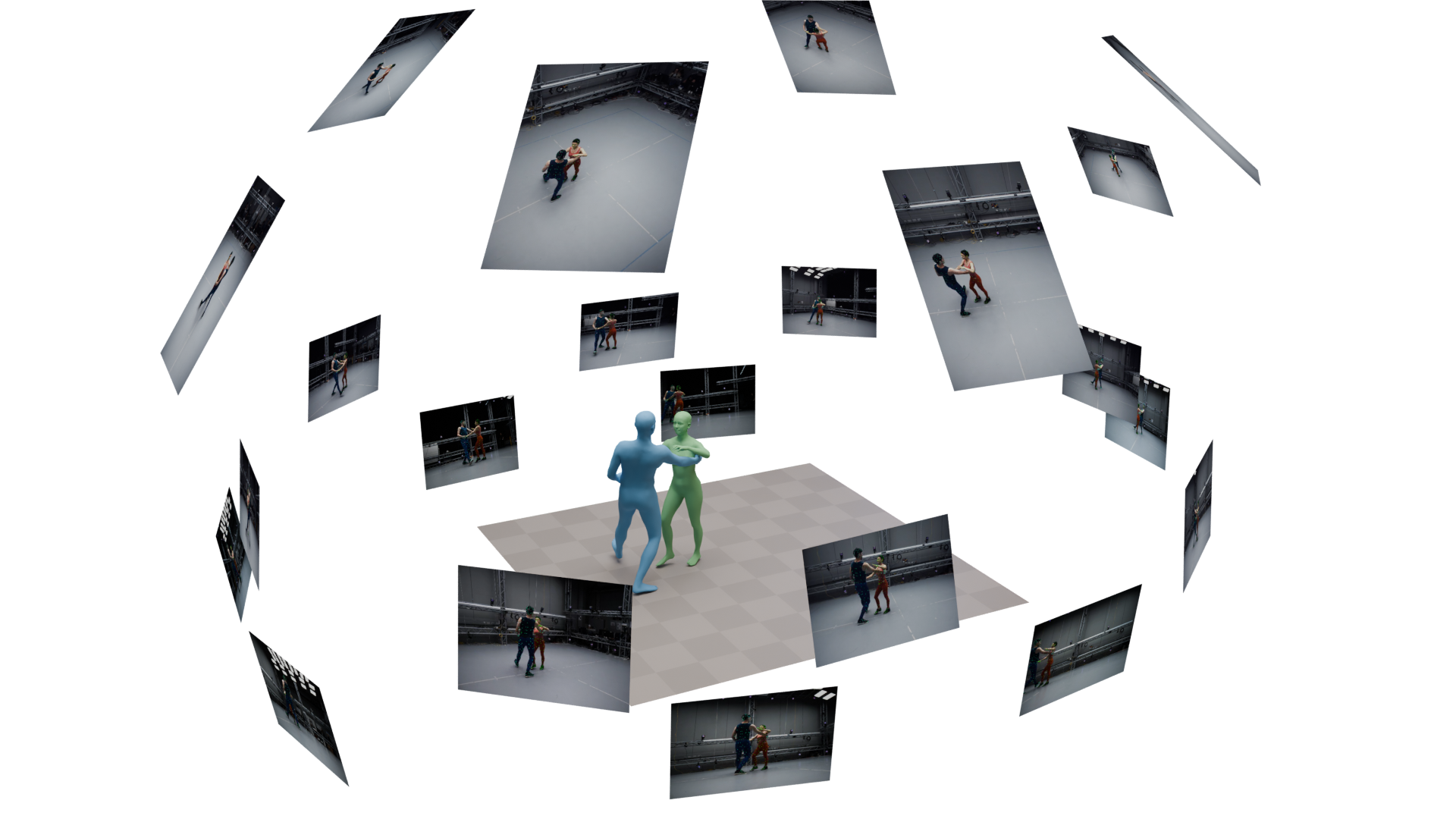}
        \caption{\datasetEval-Dance}
        \label{fig:ioi-A}
    \end{subfigure}
    \hfill
    \begin{subfigure}{0.49\linewidth}
        \centering
        \includegraphics[width=\linewidth]{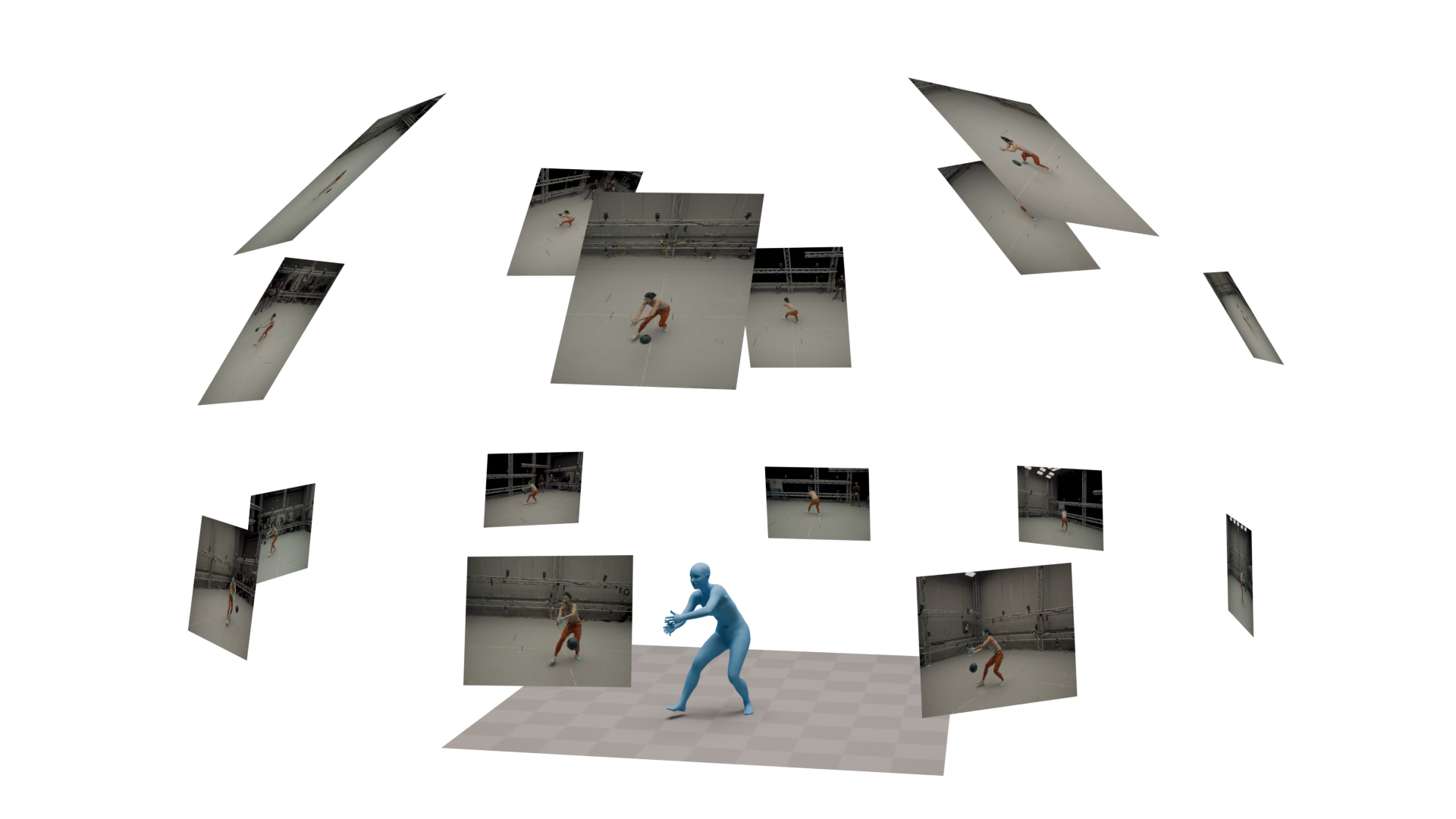}
        \caption{\datasetEval-Singles}
        \label{fig:ioi-B}
    \end{subfigure}

    \vspace{0.5em}

    \begin{subfigure}{1.\linewidth}
        \centering
        \includegraphics[width=\linewidth]{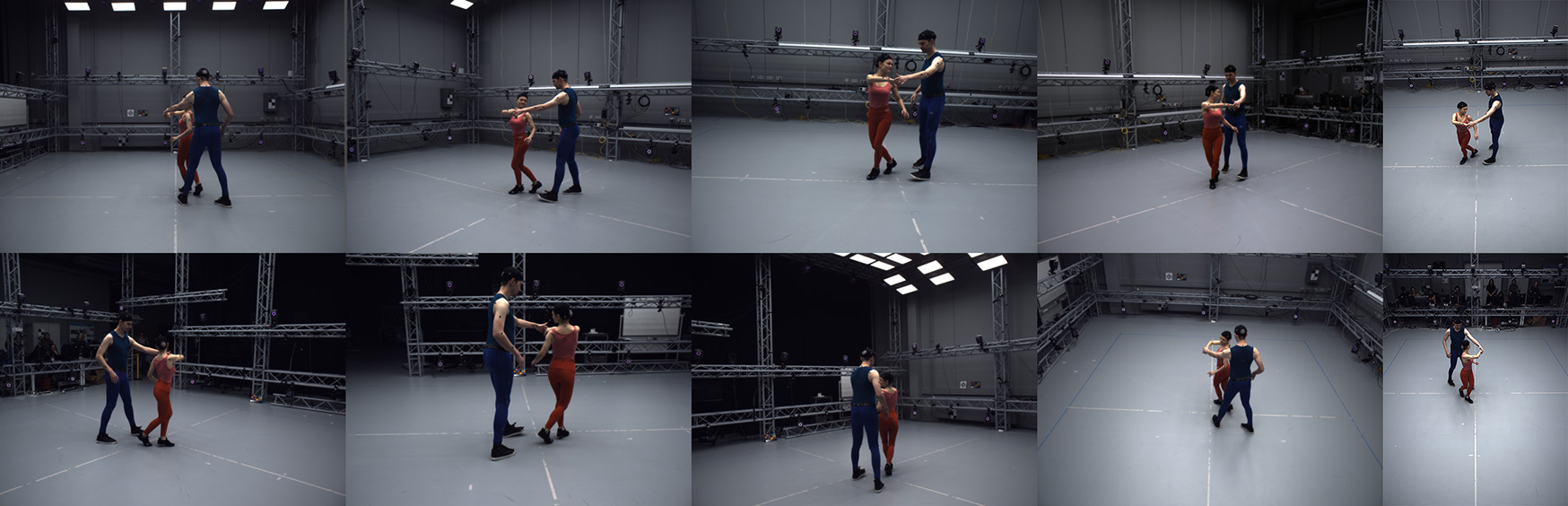}
        \caption{Sample views from the IOI cameras.}
        \label{fig:ioi-C}
    \end{subfigure}
\caption{Our \datasetEval capture setup.}
\label{fig:ioi-eval-Dance}
\end{figure*}

We captured our evaluation datasets (\datasetEval-Singles, \datasetEval-Dance, and \datasetEval-Extra) using a Vicon marker-based motion capture system, synchronized with a multi-view RGB camera system (IO Industries (IOI)). The motion capture setup consists of 54 VICON Vantage V16 cameras. The IOI setup includes 16 Victorem and 17 Volucam cameras. Both systems are mounted on a truss structure with the dimensions of 9m by 7m with a height of 5m. Within that space, the IOI cameras are positioned to cover a 3m$\times$3m$\times$3m space with the subject's full body visible in all views (see \cref{fig:ioi-eval-Dance}). Custom 12K Lux lighting by NORKA provides enough light to avoid shadows and motion blur at high capture frame rates. The lights  are not constantly powered but are triggered to only turn on together with the IOI system when the shutter is open during each capture frame. This protects the subjects' vision from being permanently exposed to high-intensity light and prevents overheating while providing sufficient lighting for the captured images. All systems are synchronized using transistor-transistor Logic (TTL), precision time protocol (PTP), and linear timecode (LTC) to ensure accurate calibration and a frame-to-frame correspondence of captured data. A key factor in creating high-accuracy data across both camera systems is our custom calibration procedure. The VICON calibration wand is utilized to capture a calibration sequence that is processed to determine camera parameters in a joint space with a common origin. All sequences are captured at 30 fps for the IOI, with the framerate for VICON varying from 60 fps for \datasetEval-Dance and \datasetEval-Extra to 120 fps for \datasetEval-Single.

For our evaluation comparing Vicon with \model, we need to know ground truth marker locations on the surface of the SMPL-X body.
To that end, we place markers on a subject and then capture their 3D body shape using a 3dMD body scanner.
We fit the SMPL-X mesh to this scan to get the true body shape in SMPL-X topology.
We then capture the subject moving in the motion capture system and run MoSh++ with the ground 3D shape template.
This lets us establish the marker locations on the 3D surface of the body.
We use these known locations when evaluating held-out marker error in the comparison section in the main document.
Note that we do not use the ground truth body shape when in \model or when running MoSh++ for the comparison with \model. In both cases, the methods estimate the body shape directly from the images or markers, respectively.

\section{Comparison with Harmony4D}
\begin{figure*}[htbp]
    \centering
    \includegraphics[width=0.98\textwidth]{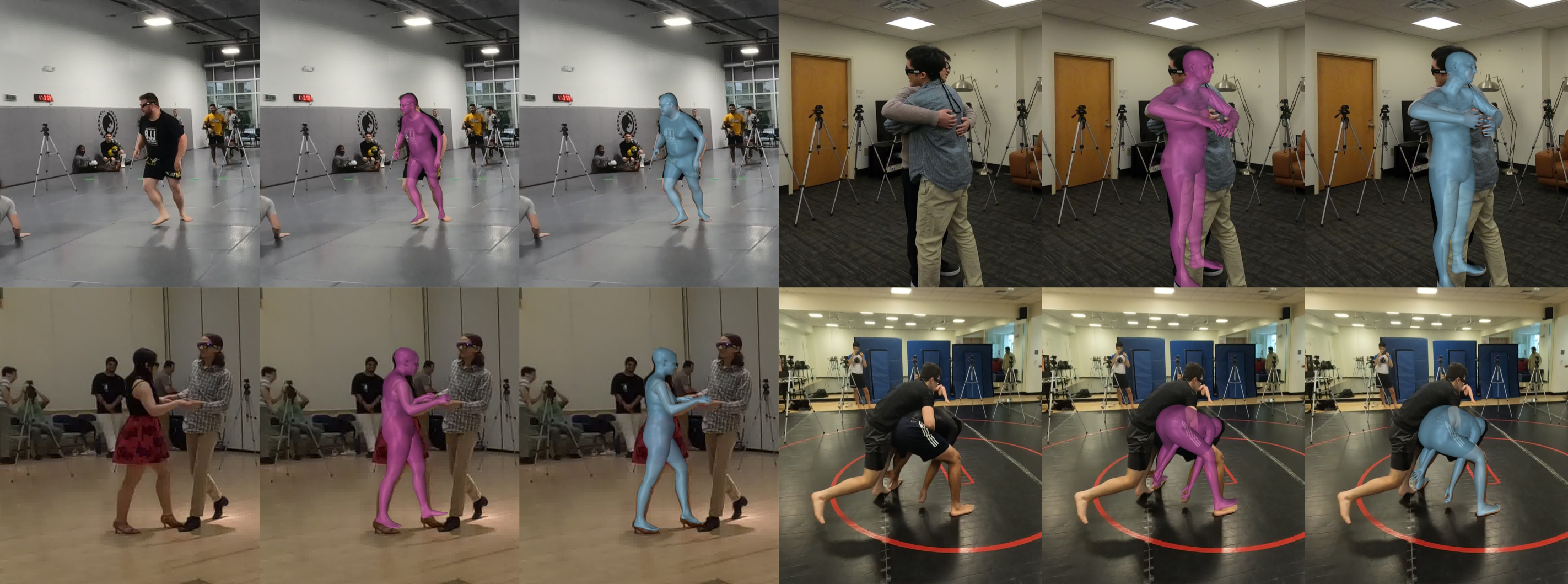}
    \caption{Mesh reprojection comparision between Harmony4D (red) and ours (blue).}
    \label{fig:comp_harmony4d}
\end{figure*}

The closest work to ours is Harmony4D. However, we cannot directly compare with that method since the test set was generated using their own method. Instead, we evaluate the silhouette reprojections against SAM2~\cite{sam2} mask predictions.
For this comparison, we reproject the Harmony4D meshes into each view and use the resulting silhouettes as masks for our network. However, as illustrated in the main paper and \cref{fig:comp_harmony4d}, the provided ground-truth meshes are often misaligned or do not perfectly conform to the subject's body, leading to noisy and imperfect input for our network. We achieve greater mean IoU (mIoU) regardless, 74.28\% vs 69.80\%. \cref{fig:comp_harmony4d} illustrates that our method overlaps better with the silhouette of the people. We also highlight that Harmony4D uses an off-the-shelf method to obtain the body shapes, meanwhile our network does not have any special initialization or body shape and pose prior. Note that although SAM2 is generally good in extracting  masks of the desired person, it sometimes misses parts of the body during video segmentation. This limitation is the main reason why the average IoU is around $70\%$ rather than higher.

\begin{table*} 
\centering
\caption{Full Benchmark 3D fitting errors (mm). We evaluate the error for the full body, only for the body, and only for the hands.}
\resizebox{0.8\linewidth}{!}{%
\begin{tblr}{
  row{1} = {c},
  row{2} = {c},
  row{8} = {c},
  row{14} = {c},
  cell{1}{1} = {r=2}{},
  cell{1}{2} = {c=2}{},
  cell{1}{4} = {c=2}{},
  cell{1}{6} = {c=2}{},
  cell{1}{8} = {c=2}{},
  cell{1}{10} = {c=2}{},
  cell{1}{12} = {c=2}{},
  cell{3}{2} = {c},
  cell{3}{3} = {c},
  cell{3}{4} = {c},
  cell{3}{5} = {c},
  cell{3}{6} = {c},
  cell{3}{7} = {c},
  cell{3}{8} = {c},
  cell{3}{9} = {c},
  cell{3}{10} = {c},
  cell{3}{11} = {c},
  cell{3}{12} = {c},
  cell{3}{13} = {c},
  cell{4}{2} = {c},
  cell{4}{3} = {c},
  cell{4}{4} = {c},
  cell{4}{5} = {c},
  cell{4}{6} = {c},
  cell{4}{7} = {c},
  cell{4}{8} = {c},
  cell{4}{9} = {c},
  cell{4}{10} = {c},
  cell{4}{11} = {c},
  cell{4}{12} = {c},
  cell{4}{13} = {c},
  cell{5}{2} = {c},
  cell{5}{3} = {c},
  cell{5}{4} = {c},
  cell{5}{5} = {c},
  cell{5}{6} = {c},
  cell{5}{7} = {c},
  cell{5}{8} = {c},
  cell{5}{9} = {c},
  cell{5}{10} = {c},
  cell{5}{11} = {c},
  cell{5}{12} = {c},
  cell{5}{13} = {c},
  cell{6}{2} = {c},
  cell{6}{3} = {c},
  cell{6}{4} = {c},
  cell{6}{5} = {c},
  cell{6}{6} = {c},
  cell{6}{7} = {c},
  cell{6}{8} = {c},
  cell{6}{9} = {c},
  cell{6}{10} = {c},
  cell{6}{11} = {c},
  cell{6}{12} = {c},
  cell{6}{13} = {c},
  cell{7}{2} = {c},
  cell{7}{3} = {c},
  cell{7}{4} = {c},
  cell{7}{5} = {c},
  cell{7}{6} = {c},
  cell{7}{7} = {c},
  cell{7}{8} = {c},
  cell{7}{9} = {c},
  cell{7}{10} = {c},
  cell{7}{11} = {c},
  cell{7}{12} = {c},
  cell{7}{13} = {c},
  cell{8}{1} = {c=13}{},
  cell{9}{2} = {c},
  cell{9}{3} = {c},
  cell{9}{4} = {c},
  cell{9}{5} = {c},
  cell{9}{6} = {c},
  cell{9}{7} = {c},
  cell{9}{8} = {c},
  cell{9}{9} = {c},
  cell{9}{10} = {c},
  cell{9}{11} = {c},
  cell{9}{12} = {c},
  cell{9}{13} = {c},
  cell{10}{2} = {c},
  cell{10}{3} = {c},
  cell{10}{4} = {c},
  cell{10}{5} = {c},
  cell{10}{6} = {c},
  cell{10}{7} = {c},
  cell{10}{8} = {c},
  cell{10}{9} = {c},
  cell{10}{10} = {c},
  cell{10}{11} = {c},
  cell{10}{12} = {c},
  cell{10}{13} = {c},
  cell{11}{2} = {c},
  cell{11}{3} = {c},
  cell{11}{4} = {c},
  cell{11}{5} = {c},
  cell{11}{6} = {c},
  cell{11}{7} = {c},
  cell{11}{8} = {c},
  cell{11}{9} = {c},
  cell{11}{10} = {c},
  cell{11}{11} = {c},
  cell{11}{12} = {c},
  cell{11}{13} = {c},
  cell{12}{2} = {c},
  cell{12}{3} = {c},
  cell{12}{4} = {c},
  cell{12}{5} = {c},
  cell{12}{6} = {c},
  cell{12}{7} = {c},
  cell{12}{8} = {c},
  cell{12}{9} = {c},
  cell{12}{10} = {c},
  cell{12}{11} = {c},
  cell{12}{12} = {c},
  cell{12}{13} = {c},
  cell{13}{2} = {c},
  cell{13}{3} = {c},
  cell{13}{4} = {c},
  cell{13}{5} = {c},
  cell{13}{6} = {c},
  cell{13}{7} = {c},
  cell{13}{8} = {c},
  cell{13}{9} = {c},
  cell{13}{10} = {c},
  cell{13}{11} = {c},
  cell{13}{12} = {c},
  cell{13}{13} = {c},
  cell{14}{1} = {c=13}{},
  cell{15}{2} = {c},
  cell{15}{3} = {c},
  cell{15}{4} = {c},
  cell{15}{5} = {c},
  cell{15}{6} = {c},
  cell{15}{7} = {c},
  cell{15}{8} = {c},
  cell{15}{9} = {c},
  cell{15}{10} = {c},
  cell{15}{11} = {c},
  cell{15}{12} = {c},
  cell{15}{13} = {c},
  cell{16}{2} = {c},
  cell{16}{3} = {c},
  cell{16}{4} = {c},
  cell{16}{5} = {c},
  cell{16}{6} = {c},
  cell{16}{7} = {c},
  cell{16}{8} = {c},
  cell{16}{9} = {c},
  cell{16}{10} = {c},
  cell{16}{11} = {c},
  cell{16}{12} = {c},
  cell{16}{13} = {c},
  cell{17}{2} = {c},
  cell{17}{3} = {c},
  cell{17}{4} = {c},
  cell{17}{5} = {c},
  cell{17}{6} = {c},
  cell{17}{7} = {c},
  cell{17}{8} = {c},
  cell{17}{9} = {c},
  cell{17}{10} = {c},
  cell{17}{11} = {c},
  cell{17}{12} = {c},
  cell{17}{13} = {c},
  cell{18}{2} = {c},
  cell{18}{3} = {c},
  cell{18}{4} = {c},
  cell{18}{5} = {c},
  cell{18}{6} = {c},
  cell{18}{7} = {c},
  cell{18}{8} = {c},
  cell{18}{9} = {c},
  cell{18}{10} = {c},
  cell{18}{11} = {c},
  cell{18}{12} = {c},
  cell{18}{13} = {c},
  cell{19}{2} = {c},
  cell{19}{3} = {c},
  cell{19}{4} = {c},
  cell{19}{5} = {c},
  cell{19}{6} = {c},
  cell{19}{7} = {c},
  cell{19}{8} = {c},
  cell{19}{9} = {c},
  cell{19}{10} = {c},
  cell{19}{11} = {c},
  cell{19}{12} = {c},
  cell{19}{13} = {c},
  hline{1,3,8-9,14-15,20} = {-}{},
  hline{2} = {2-13}{},
}
Model       & RICH           &                & Harmony4D      &                & CHI3D          &                & \datasetEval-S    &                & \datasetEval-D      &                & MOYO           &                \\
            & MPJPE          & PVE            & MPJPE          & PVE            & MPJPE          & PVE            & MPJPE          & PVE            & MPJPE          & PVE            & MPJPE          & PVE            \\
SMPLfiy     & 96.18          & 71.42          & -              & -              & 67.68          & 51.79          & 47.15          & 35.42          & 53.92          & 43.08          & 62.15          & 44.68          \\
LookMa*     & 39.52          & 30.29          & 59.37          & 45.6           & 46.47          & 39.36          & 25.97          & 23.94          & 27.98          & 24.89          & 60.15          & 53.82          \\
CameraHMR~  & 25.61          & 21.36          & 58.59          & 42.0           & 40.8           & 34.61          & 15.25          & 18.43          & 20.41          & 21.06          & 33.75          & 33.74          \\
\model & 22.20          & 19.76          & \textbf{45.26} & \textbf{34.02} & 38.01          & 32.84          & 12.96          & 17.18          & \textbf{17.71} & 19.80          & 22.95          & 25.48          \\
\model-C     & \textbf{22.20} & \textbf{19.76} & 45.35          & 34.05          & \textbf{37.96} & \textbf{32.82} & \textbf{12.96} & \textbf{17.18} & 17.73          & \textbf{19.78} & \textbf{22.95} & \textbf{25.48} \\
Only body   &                &                &                &                &                &                &                &                &                &                &                &                \\
SMPLfiy     & 71.34          & 63.99          & -              & -              & 40.05          & 44.21          & 31.4           & 30.81          & 37.69          & 38.92          & 47.81          & 39.66          \\
LookMa*     & 37.78          & 28.89          & 49.09          & 42.69          & 33.48          & 35.18          & 25.34          & 22.7           & 23.81          & 23.41          & 62.02          & 53.57          \\
CameraHMR~  & 28.18          & 21.28          & 46.74          & 38.67          & 30.58          & 30.83          & 18.27          & 18.5           & 19.17          & 20.43          & 33.0           & 33.37          \\
\model & 27.32          & 20.44          & \textbf{40.84} & \textbf{32.58} & 28.36          & 29.25          & 17.61          & 17.88          & 18.86          & 19.95          & 24.72          & 25.89          \\
\model-C     & \textbf{27.32} & \textbf{20.44} & 40.87          & 32.6           & \textbf{28.35} & \textbf{29.25} & \textbf{17.61} & \textbf{17.88} & \textbf{18.82} & \textbf{19.93} & \textbf{24.72} & \textbf{25.89} \\
Only hands  &                &                &                &                &                &                &                &                &                &                &                &                \\
SMPLfiy     & 112.73         & 93.83          & -              & -              & 86.1           & 74.64          & 57.65          & 49.29          & 64.73          & 55.62          & 71.71          & 59.80          \\
LookMa*     & 40.67          & 34.5           & 66.23          & 54.37          & 55.14          & 51.95          & 26.38          & 27.68          & 30.76          & 29.34          & 58.9           & 54.56          \\
CameraHMR~  & 23.91          & 21.61          & 66.5           & 52.05          & 47.62          & 45.98          & 13.24          & 18.21          & 21.24          & 22.97          & 34.24          & 34.87          \\
\model & 18.79          & 17.72          & \textbf{48.2}  & \textbf{38.33} & 44.44          & 43.65          & 9.85           & 15.06          & \textbf{16.94} & \textbf{19.33} & 21.77          & 24.26          \\
\model-C     & \textbf{18.79} & \textbf{17.72} & 48.34          & 38.41          & \textbf{44.36} & \textbf{43.6}  & \textbf{9.85}  & \textbf{15.06} & 17.01          & 19.34          & \textbf{21.77} & \textbf{24.26}
\end{tblr}
}

\label{tab:results_without_vtemplate_body_hands_only}
\end{table*}

\begin{table}

\centering
\caption{IoU comparison between our method and Harmony4D on the Harmony4D test set.}
\resizebox{0.7\linewidth}{!}{%
\begin{tblr}{
  row{1} = {c},
  cell{2}{2} = {c},
  cell{2}{3} = {c},
  cell{3}{2} = {c},
  cell{3}{3} = {c},
  cell{4}{2} = {c},
  cell{4}{3} = {c},
  cell{5}{2} = {c},
  cell{5}{3} = {c},
  cell{6}{2} = {c},
  cell{6}{3} = {c},
  cell{7}{2} = {c},
  cell{7}{3} = {c},
  cell{8}{2} = {c},
  cell{8}{3} = {c},
  cell{9}{2} = {c},
  cell{9}{3} = {c},
  cell{10}{2} = {c},
  cell{10}{3} = {c},
  cell{11}{2} = {c},
  cell{11}{3} = {c},
  cell{12}{2} = {c},
  cell{12}{3} = {c},
  cell{13}{2} = {c},
  cell{13}{3} = {c},
  cell{14}{2} = {c},
  cell{14}{3} = {c},
  cell{15}{2} = {c},
  cell{15}{3} = {c},
  cell{16}{2} = {c},
  cell{16}{3} = {c},
  cell{17}{2} = {c},
  cell{17}{3} = {c},
  cell{18}{2} = {c},
  cell{18}{3} = {c},
  cell{19}{2} = {c},
  cell{19}{3} = {c},
  cell{20}{2} = {c},
  cell{20}{3} = {c},
  cell{21}{2} = {c},
  cell{21}{3} = {c},
  cell{22}{2} = {c},
  cell{22}{3} = {c},
  cell{23}{2} = {c},
  cell{23}{3} = {c},
  cell{24}{2} = {c},
  cell{24}{3} = {c},
  cell{25}{2} = {c},
  cell{25}{3} = {c},
  cell{26}{2} = {c},
  cell{26}{3} = {c},
  cell{27}{2} = {c},
  cell{27}{3} = {c},
  cell{28}{2} = {c},
  cell{28}{3} = {c},
  cell{29}{2} = {c},
  cell{29}{3} = {c},
  cell{30}{2} = {c},
  cell{30}{3} = {c},
  cell{31}{2} = {c},
  cell{31}{3} = {c},
  cell{32}{2} = {c},
  cell{32}{3} = {c},
  cell{33}{2} = {c},
  cell{33}{3} = {c},
  cell{34}{2} = {c},
  cell{34}{3} = {c},
  cell{35}{2} = {c},
  cell{35}{3} = {c},
  cell{36}{2} = {c},
  cell{36}{3} = {c},
  cell{37}{2} = {c},
  cell{37}{3} = {c},
  cell{38}{2} = {c},
  cell{38}{3} = {c},
  cell{39}{2} = {c},
  cell{39}{3} = {c},
  cell{40}{2} = {c},
  cell{40}{3} = {c},
  cell{41}{2} = {c},
  cell{41}{3} = {c},
  hline{1-2,41-42} = {-}{},
}
Sequence        & Harmony4D     & Ours w/ mask    \\
002\_hugging    & 75.72         & \textbf{77.96}  \\
025\_grappling2 & 66.26         & \textbf{71.94}  \\
028\_grappling2 & 72.26         & \textbf{78.34}  \\
030\_grappling2 & 70.26         & \textbf{77.18}  \\
031\_grappling2 & 74.35         & \textbf{80.44}  \\
032\_grappling2 & 54.79         & \textbf{60.29}  \\
033\_grappling2 & 68.47         & \textbf{74.95}           \\
034\_grappling2 & 64.16         & \textbf{71.48}  \\
035\_grappling2 & 69.42         & \textbf{76.16}  \\
036\_grappling2 & 67.82         & \textbf{74.00}  \\
037\_grappling2 & 65.71         & \textbf{72.25}  \\
038\_grappling2 & 71.93         & \textbf{78.71}  \\
039\_grappling2 & 66.44         & \textbf{72.78}  \\
040\_grappling2 & 73.41         & \textbf{78.62}  \\
041\_grappling2 & 71.59         & \textbf{77.72}  \\
042\_grappling2 & 71.62         & \textbf{77.47}  \\
043\_grappling2 & 71.94         & \textbf{78.61}  \\
009\_sword2     & 68.25         & \textbf{69.19}           \\
010\_sword2     & 65.16         & \textbf{66.03}  \\
001\_sword3     & \textbf{70.70}& 70.36           \\
002\_sword3     & 71.38         & \textbf{72.52}  \\
003\_sword3     & 71.57         & \textbf{72.52}  \\
004\_sword3     & 71.39         & \textbf{72.02}  \\
005\_sword3     & 71.84         & \textbf{72.30}  \\
006\_sword3     & 70.71         & \textbf{71.41}  \\
007\_ballroom2  & 67.72         & \textbf{72.11}  \\
008\_ballroom2  & 69.51         & \textbf{73.29}  \\
009\_ballroom2  & 69.52         & \textbf{72.84}  \\
010\_ballroom2  & 69.41         & \textbf{73.11}  \\
016\_mma4       & 65.36         & \textbf{76.18}  \\
001\_mma5       & 70.69         & \textbf{74.86}  \\
002\_mma5       & 72.19         & \textbf{77.27}  \\
003\_mma5       & 71.19         & \textbf{75.85}  \\
004\_mma5       & 71.86         & \textbf{76.57}  \\
005\_mma5       & 71.79         & \textbf{76.77}  \\
009\_mma5       & 71.40         & \textbf{75.69}  \\
011\_mma5       & 71.40         & \textbf{76.07}  \\
013\_mma5       & 72.11         & \textbf{76.09}  \\
016\_mma5       & 71.02         & \textbf{74.95}  \\
mIoU            & 69.80         & \textbf{74.28}
\end{tblr}
}

\label{tab:iou_comparison_full}
\end{table}

\section{Multiview Matching Evaluation}
In the main paper we show that our Multiview algorithm is able to match people across views with a success rate of $100\%$ in the evaluation datasets. The datasets Harmony4D and \datasetEval-D have around 20 cameras, whereas CHI3D has only 4.

Here, we evaluate the robustness of our algorithm by reducing the number of cameras from 16 down to 2 in powers of two (16, 8, 4, 2). We used \datasetEval-D dataset. After the evaluation we observe that even with 2 cameras our method is capable to match correctly the people in the scene $100\%$ of the times. This shows that our dense landmarks plus temporal information (from SAM2) provide rich and accurate geometric information that is useful for matching the identity of a person across views. Note that for the single-frame case, our method can fail, as wrong mask predictions or heavy occlusions can confuse the network and lead to incorrect landmark predictions. This is particularly true when the limbs are heavily occluded. In those cases, the network has to "guess'' the location of those parts, which can make them not view-consistent.

\section{Optimization: Number of Cameras}
We evaluated the effect of camera count on single and two-person cases, \datasetEval-(S)ingles and \datasetEval-(D)ance respectively. To ensure uniform spatial coverage, we sample [2, 4, 8, 12, 16] cameras using Farthest Point Sampling (FPS).
\cref{fig:cam_variation_acc} shows that for both single and two-person cases, our method achieves strong performance with as few as 4 cameras, and reaches optimal accuracy at around 12 cameras after which improvements saturate.
The strong performance with as few as 4-8 cameras suggests that our method can be used with low-cost capture setups.
\begin{figure}
    \centering
        \begin{tikzpicture}
        \begin{axis}[
            width=0.95\linewidth,
            height=3.8cm,
            xlabel={Number of cameras},
            ylabel={MPJPE (mm)},
            xlabel style={font=\footnotesize},
            ylabel style={font=\footnotesize},
            tick label style={font=\footnotesize},
            legend style={font=\footnotesize},
            xtick={2,4,8,12,16,20,22},
            grid=both,
            grid style={dashed, line width=0.8pt, black!35},
            legend pos=north east,
            legend cell align=left,
            axis on top=false,
            cycle list={
                {blue, dashed, mark=*, mark size=1.5pt},
                {orange, dashed, mark=square*, mark size=1.5pt}
            },
        ]

        \addplot
        coordinates {
            (2,24.13)
            (4,13.66)
            (8,13.29)
            (12,12.89)
            (16,12.96)
        };
        \addlegendentry{MAMMA-Singles}

        \addplot
        coordinates {
            (2,47.74)
            (4,20.18)
            (8,17.65)
            (12,17.33)
            (16,17.42)
        };
        \addlegendentry{MAMMA-Dance}

        \end{axis}
        \end{tikzpicture}

    \caption{\textbf{Camera variation accuracy.}}
    \label{fig:cam_variation_acc}
\end{figure}

\section{Optimization: Stages Evaluation}
We measured the average runtime of each stage S on \datasetEval-D in \cref{fig:optim_stages_runtime}:
S1: global translation and rotation estimation.
S2: optimization of pose, shape, and translation.
S3: update of uncertainty weights based on reprojection error.
S4: incorporation of contact constraints.
S2 already provides a good trade-off between accuracy and runtime. S3 and S4 further refine local details and substantially reduce penetrations.
\begin{figure}
    \centering
        \begin{tikzpicture}
        \begin{axis}[
            ytick={0, 25, 50, 100, 200, 300},
            width=0.95\linewidth,
            height=4.3cm,
            xlabel={Average cumulative time (seconds)},
            ylabel={MPJPE (mm)},
            xlabel style={font=\footnotesize},
            ylabel style={font=\footnotesize},
            tick label style={font=\footnotesize},
            legend style={font=\footnotesize},
            grid=both,
            major grid style={dashed, gray!40},
            minor grid style={dashed, gray!20},
            legend pos=north east,
            enlarge x limits=0.08,
            enlarge y limits=0.2,
        ]
        \addplot[
            color=blue!70!black,
            dashed,
            mark=*,
            mark size=1.5pt,
            line width=1pt,
        ]
        coordinates {
            (35.83, 282.29)
            (347.56, 20.95)
            (559.08, 17.71)
            (680.23, 17.73)
        };
        \addlegendentry{MAMMA-Dance}
        
        \node[anchor=south, font=\footnotesize] at (axis cs:35.83, 282.29) {S1};
        \node[anchor=south, font=\footnotesize] at (axis cs:347.56, 20.95) {S2};
        \node[anchor=south, font=\footnotesize] at (axis cs:559.08, 17.71) {S3};
        \node[anchor=south, font=\footnotesize] at (axis cs:680.23, 17.73) {S4};
        
        \end{axis}
        \end{tikzpicture}
    \caption{\textbf{Optimization (S)tages runtime.}}
    \label{fig:optim_stages_runtime}
\end{figure}

\section{Perceptual Study}
To further evaluate the perceptual quality of our reconstructions, we conducted an Amazon Mechanical Turk study in which 33 participants rated the realism of rendered motion from GT and \model-C results using a five-point Likert scale. A one-sided Wilcoxon signed-rank test revealed that our reconstructions were perceived as significantly more realistic than the ground truth on CHI3D ($p = 2\times10^{-6}$, $\Delta = 1.36$) and Harmony4D ($p = 2.4\times10^{-4}$, $\Delta = 0.48$). For MOYO, the reconstruction also slightly outperformed the GT ($p \approx 0.0010$, $\Delta = 0.15$), although the effect size is smaller given that both versions are already highly realistic (means $\approx 4.2$–$4.4$). In contrast, no significant advantage was observed for RICH, \datasetEval-D, or \datasetEval-S, where GT and \model-C results received similarly high ratings (all $p > 0.3$, $|\Delta| \le 0.11$).

\section{Visualizing the Contact Probabilities}
\begin{figure}[tbp]
\centering
\includegraphics[width=\linewidth]{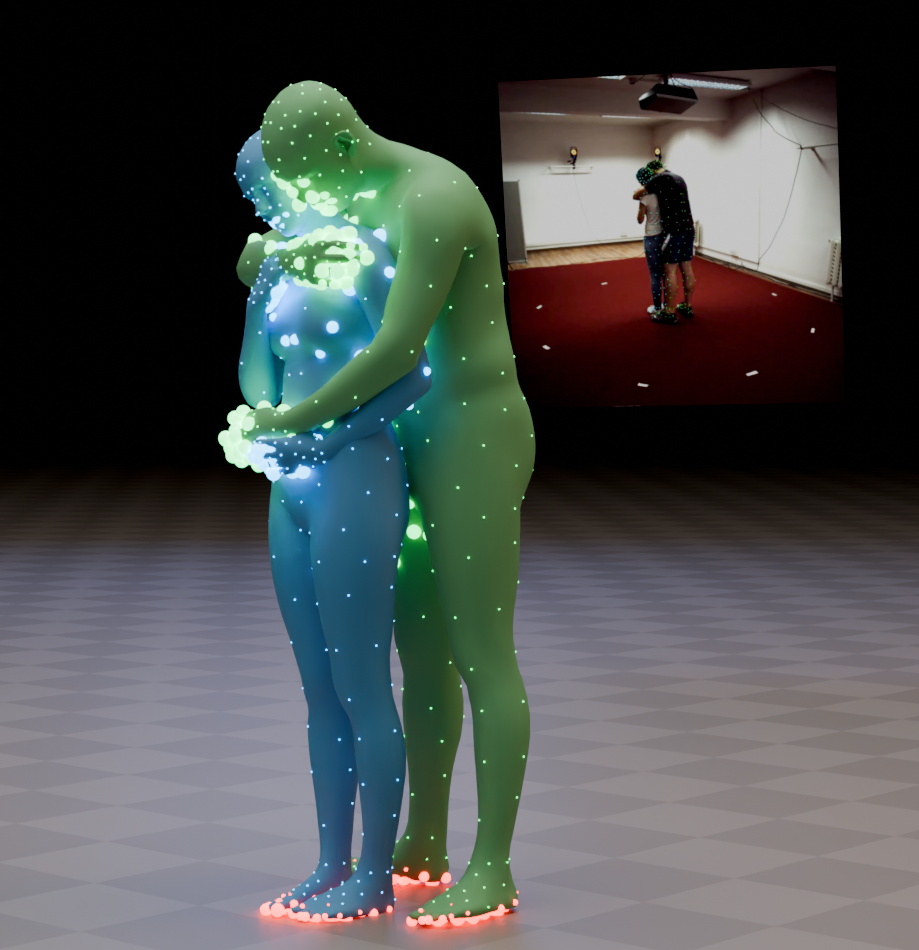}

\caption{Contact probabilities for the 512 landmarks, averaged over per-view predictions. Brighter and larger points indicate higher contact probability.}

\label{fig:contact_prob_visual}
\end{figure}

During optimization, we use contact probabilities averaged over all views. \Cref{fig:contact_prob_visual} shows a visual example of these averaged values.

\section{Capture Protocol}
Participants in the \datasetEval-Single and \datasetEval-Extra were selected to have a variety of body shapes and ethnicities, and gender balance. The \datasetEval-Dance subjects were chosen based on their proficiency as dancers. All subjects were informed about the details of the capture protocol and the use of their data in advance and gave written informed consent.

The capture procedure and data processing steps have been approved by an ethics committee. All data collection, storage, and processing activities are compliant with privacy regulations.
Participants wore tight-fitted clothing in distinctive colors to enable good quality body scanner data and the processing of the multi-view captures. In the body scanner, each subject was recorded performing several sequences and still poses.

For all captures, the \textit{FrontWaist10Fingers} marker set from Vicon was used with a total of 73 markers distributed on the body and hands of the participant. In line with the Vicon data capture process, subjects were calibrated by performing a range-of-motion sequence.

\begin{figure}[tbp]
\centering
\includegraphics[width=\linewidth]{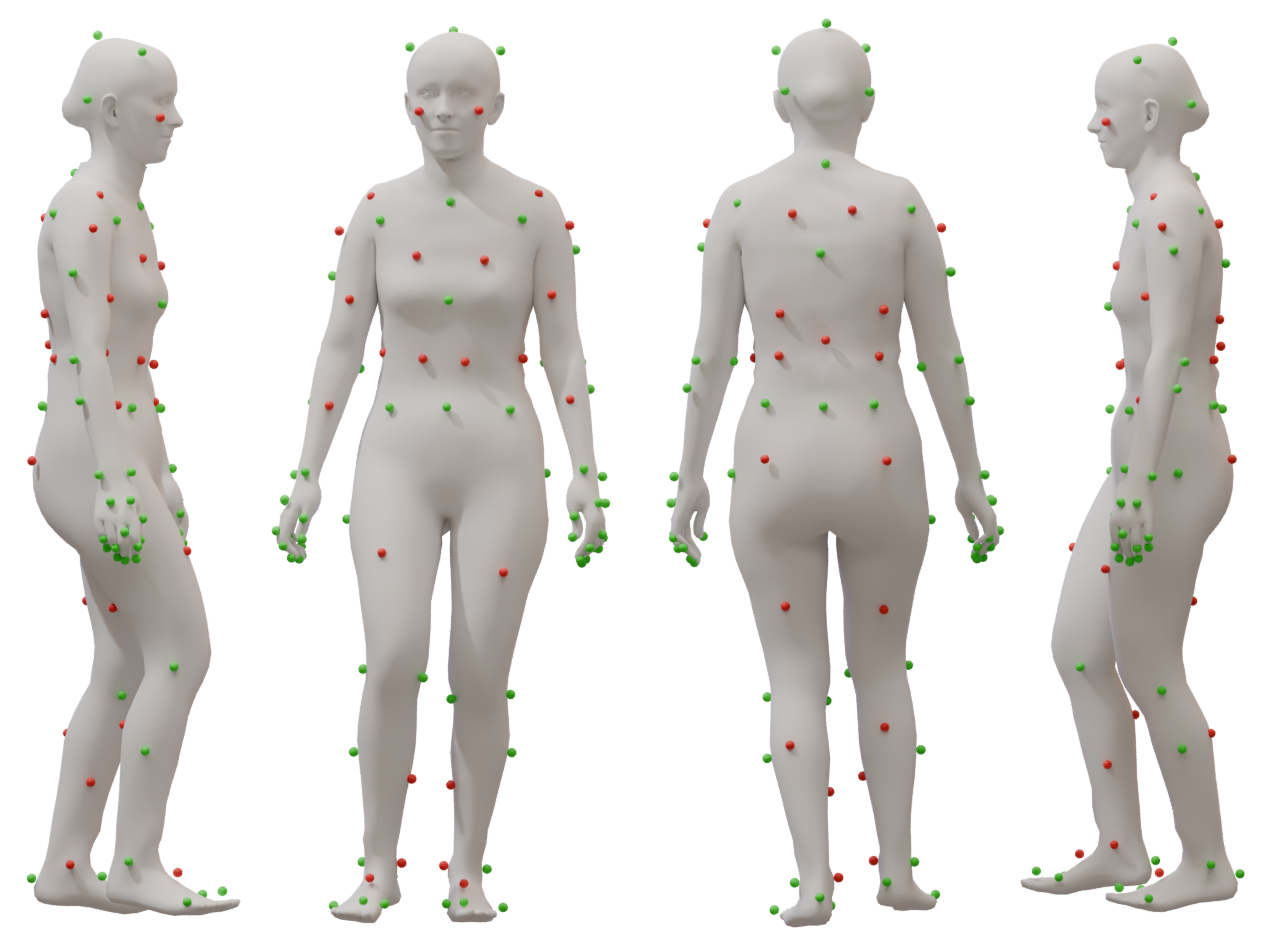}

\caption{To evaluate \model predictions on the Vicon ground truth, we use the additional held-out 37-marker layout (shown in red) alongside the default 73 FrontWaist10Fingers marker set from Vicon (green).}

\label{fig:extra_markers}
\end{figure}

For the \datasetEval-Extra, the additional 37 markers were attached after subject calibration to enable both the standard Vicon post-processing and the tracking of added markers that were labelled manually. The positions of the 37 extra markers were selected to be distinct from the standard marker positions, to avoid occlusion, and to include soft-tissue areas that are excluded from standard motion capture templates (\cref{fig:extra_markers}). Reference images of the subjects after marker placement were captured to allow for an accurate replacement of fallen markers and support further processing steps.

For the Vicon Marker-based comparison experiment, we excluded one subject due to calibration errors of the Vicon-IOI system and discarded the first 5 frames to account for trigger-light delay.

\subsection{Computing marker positions} \label{section:moshpp}

We use MoSh++ to create the GT SMPL-X fits of our \datasetEval datasets and to compute the marker positions of the methods used in our Vicon Markers comparison experiment. MoSh++ requires an initialization step that assigns each motion-capture marker to a vertex on the SMPL-X body model. In its first optimization stage, the marker positions are allowed to vary slightly to better align with the input data. While standard marker layouts often come with predefined marker-to-vertex mappings, real-world usage introduces significant variability. Due to human error and the diverse range of body shapes and clothing conditions, the actual marker placements for the same layout can differ substantially across captures, sometimes by several tens of centimeters. MoSh++ is highly sensitive to initialization quality, and an accurate marker-to-vertex mapping is critical for achieving optimal results. To ensure reliable initialization, we capture 3D scans of subjects wearing markers, fit the SMPL-X model to the scans, and manually select the most appropriate vertex for each marker.

During the first optimization stage, MoSh++ refines marker positions using a local coordinate system defined as follows: (a) the closest SMPL-X vertex to the marker serves as the origin; (b) the coordinate axes are constructed from two edges of the triangle incident to this vertex, along with their cross product to form an orthonormal basis. The final marker position is stored as a combination of the vertex index and three displacement values, each corresponding to an offset along one of the local coordinate axes. The origin and local coordinate axes obtained during this optimization stage enable us to regress marker positions for any SMPL-X body shape.

\subsection{\datasetEval-Extra Protocol}
\begin{table}[htb]
    \centering
        \caption{Vicon Markers comparison experiment: Mean per-marker distance (mm) of MoSh and \model on the Vicon held-out 37 markers of our \datasetEval-Extra dataset.}
    \resizebox{0.7\linewidth}{!}{%
    \begin{tblr}{
      colspec={c l c c},
      row{1} = {c},
      column{3} = {c},
      column{4} = {c},
      hline{1-2,14,15} = {-}{}
    }
    Subject & Action          & MoSh  & \model  \\
    00202   & solo\_dancing   & 24.368 & 25.464 \\
    00202   & calib\_routine  & 20.081 & 21.259 \\
    00202   & walking         & 23.384 & 25.841 \\
    00202   & warmup          & 21.888 & 24.108 \\
    00219   & solo\_dancing   & 23.967 & 22.889 \\
    00219   & calib\_routine  & 23.338 & 22.670 \\
    00219   & walking         & 23.507 & 24.503 \\
    00219   & warmup          & 23.898 & 21.864 \\
    00236   & solo\_dancing   & 19.864 & 21.876 \\
    00236   & calib\_routine  & 17.832 & 19.050 \\
    00236   & walking         & 18.791 & 21.002 \\
    00236   & warmup          & 18.514 & 19.248 \\
            &                 & 21.619 & 22.481 \\
    \end{tblr}
    }
    \label{tab:vicon_markers_eval}
\end{table}

The process described in Section \ref{section:moshpp}, allows us to compute the 37 additional held-out Vicon marker positions from SMPL-X vertices (\cref{fig:extra_markers}). This enables us to quantitatively compare markerless to marker-based methods. As a baseline, we use MoSh++, which takes as input only the 73 markers (\textit{FrontWaist10Fingers} marker set). In \cref{tab:vicon_markers_eval}, we report the mean per-marker distance error of MoSh++ (baseline) and \model, per-sequence, with both methods evaluated without using a GT body shape (i.e.~without using the 3D scan).

Similarly, we additionally tested our protocol on the \datasetEval-Dance sequences, but this time we held-out 9 markers (from the 73 FrontWaist10Fingers marker set) distributed across the body and ran the same protocol. As a baseline, we use MoSh++, which takes as input the remaining 64 markers. The average error of our \model is 27.590mm and 26.150mm for MoSh++ (baseline) with a difference of 1.44mm.

\section{Usability and Cost}
In addition to accuracy, the cost of setup and processing is important.
We compare the time cost of using our automatic method vs.\ the marker-based pipeline which requires manually cleaning and labelling data.
To estimate the cost of marker labelling,
we recorded the performance of 3 marker post-processing technicians who had prior experience cleaning mocap frames, with experience ranging from 118,972 to 149,454 frames, with an average of 134,500 frames.
Each technician cleaned 3 mocap sequences of 2 people dancing (West Coast Swing), while timing and documenting their workflow.
Using the Vicon Shogun Post tool, the data was inspected frame-by-frame and all missing markers (gaps) and marker swaps were manually corrected.
They cleaned 9 sequences with a total of 29,244 frames and a capture duration of 24 minutes and 31 seconds.
Cleaning required 46 hours and 51 minutes. The most time-intensive step of the pipeline was fixing the swapped markers located on the fingers, with an average time of 4.95 hours per sequence.
While marker swaps on body locations are typically less problematic than those on the fingers, the time to fix them took an average of 15 minutes.
Gap filling was performed in Shogun Post and, while manual, the time is minimal compared with the labeling.

After cleaning we ran MoSh++ on the result, which took approximately 25 hours of computation.
Ignoring the time to put on and take off markers, the time from capture to SMPL-X fits was approximately 72 hours.
In contrast, running \model takes approximately $26$ hours on a single GTX-4090. SAM2 runs at 2 fps, \net at 12 fps and the optimization on average takes 65 secs. for a sequence of 100 frames.

\section{\model Dance Release}

\begin{figure}[tbp]
\centerline{\includegraphics[width=1.0\linewidth]{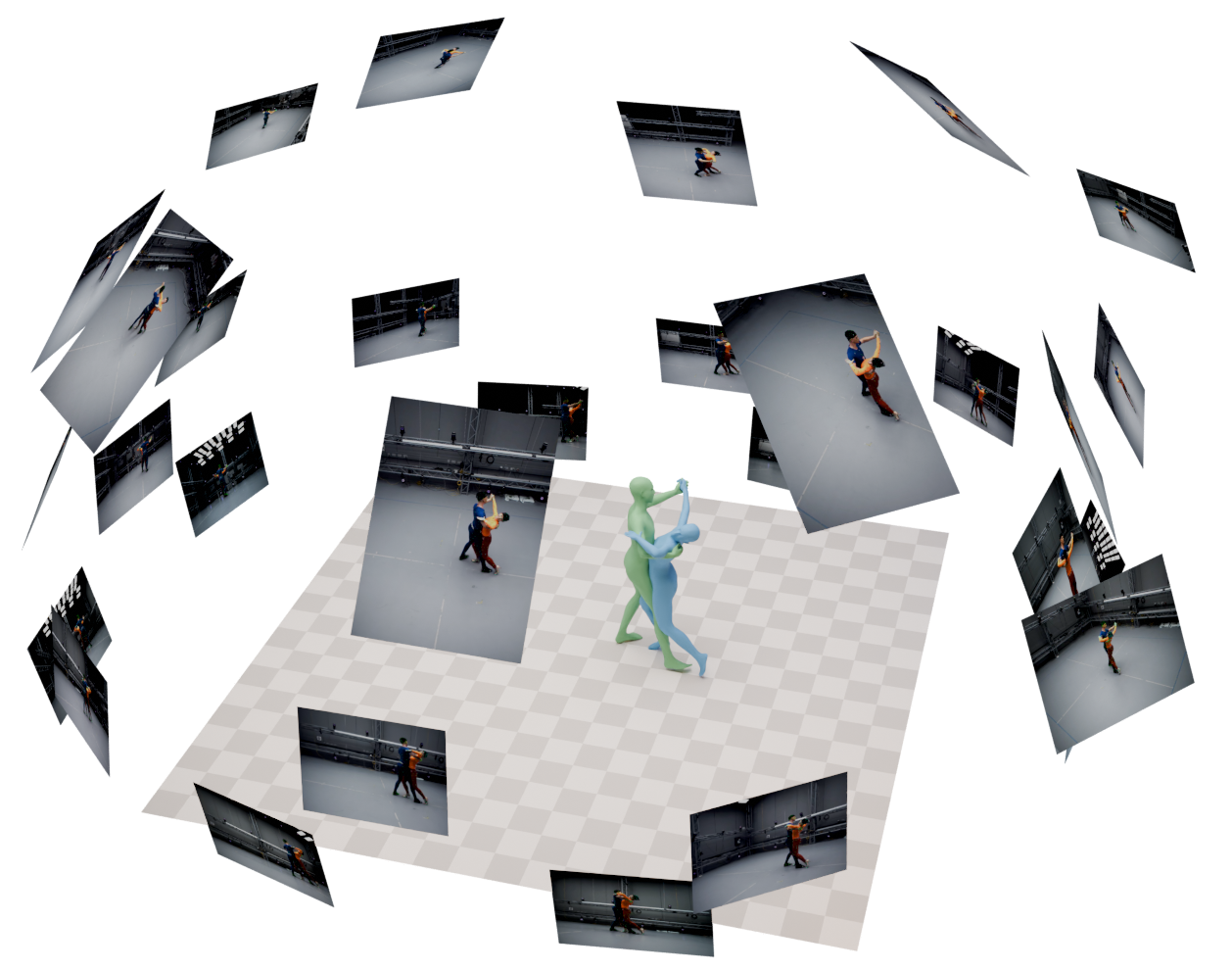}}
\caption{Multi-view setup of the \model Dance sequences captured solely with \model.}
\label{fig:dance_release}
\end{figure}

In addition to \dataset, \datasetEval-Singles, \datasetEval-Dance and \datasetEval-Extra, we release a new dataset captured using our \model pipeline. The dataset includes Bachata, West Coast Swing, Breakdance and Ballroom, performed by 9 subjects for approximately 1 hour in total. We capture at 30 fps, from 32 synchronized cameras, using the multi-view setup shown in \cref{fig:dance_release}. For the released dataset  we used the masks predicted by SAM3.

\section{More than Two People}
\begin{figure}[tbp]
\centerline{\includegraphics[width=1.0\linewidth]{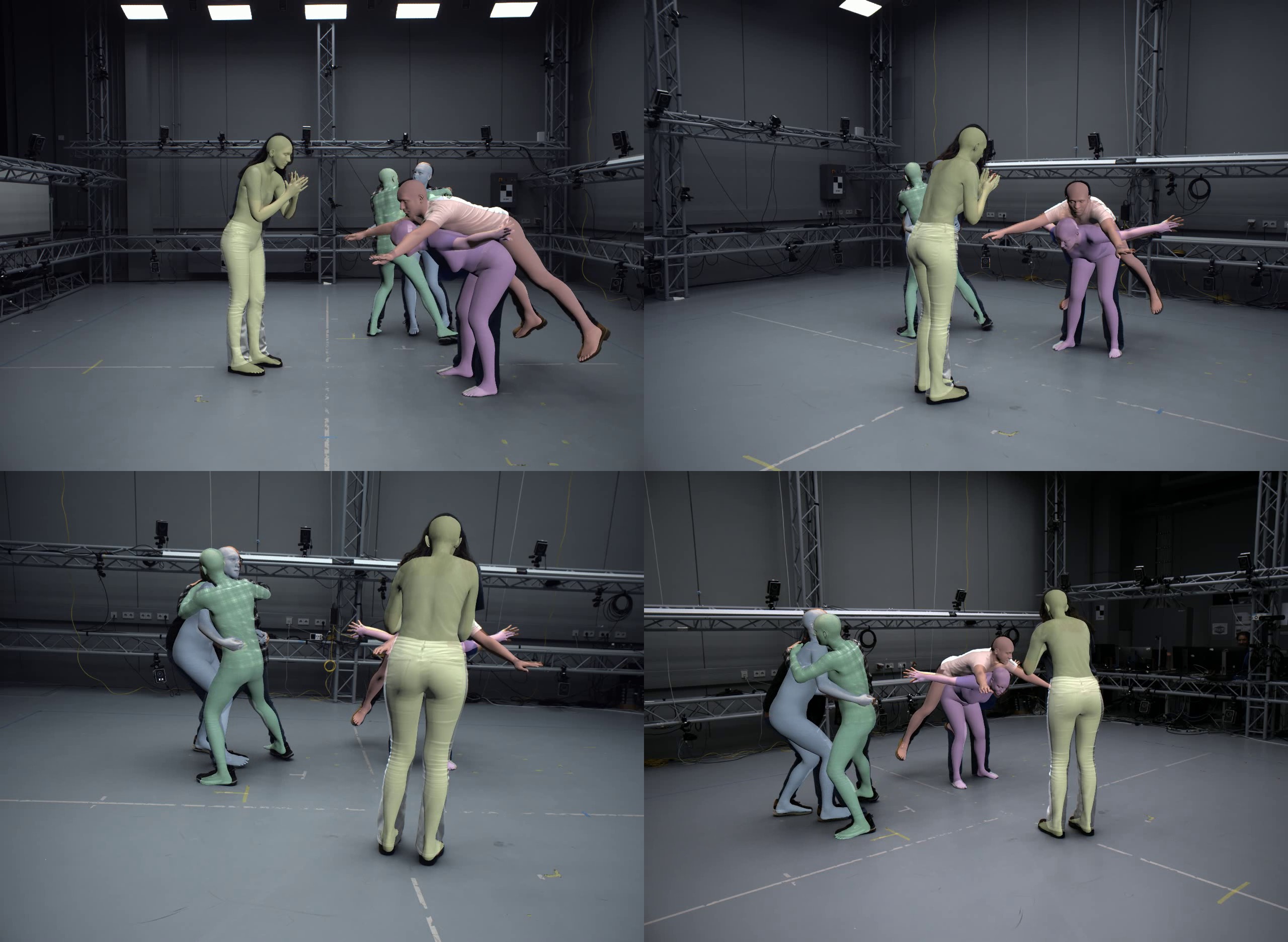}}
\caption{Motion captured by \model with five people interacting in the scene.}
\label{fig:multiview_five_people}
\end{figure}

\begin{figure}[tbp]
\centerline{\includegraphics[width=1.0\linewidth]{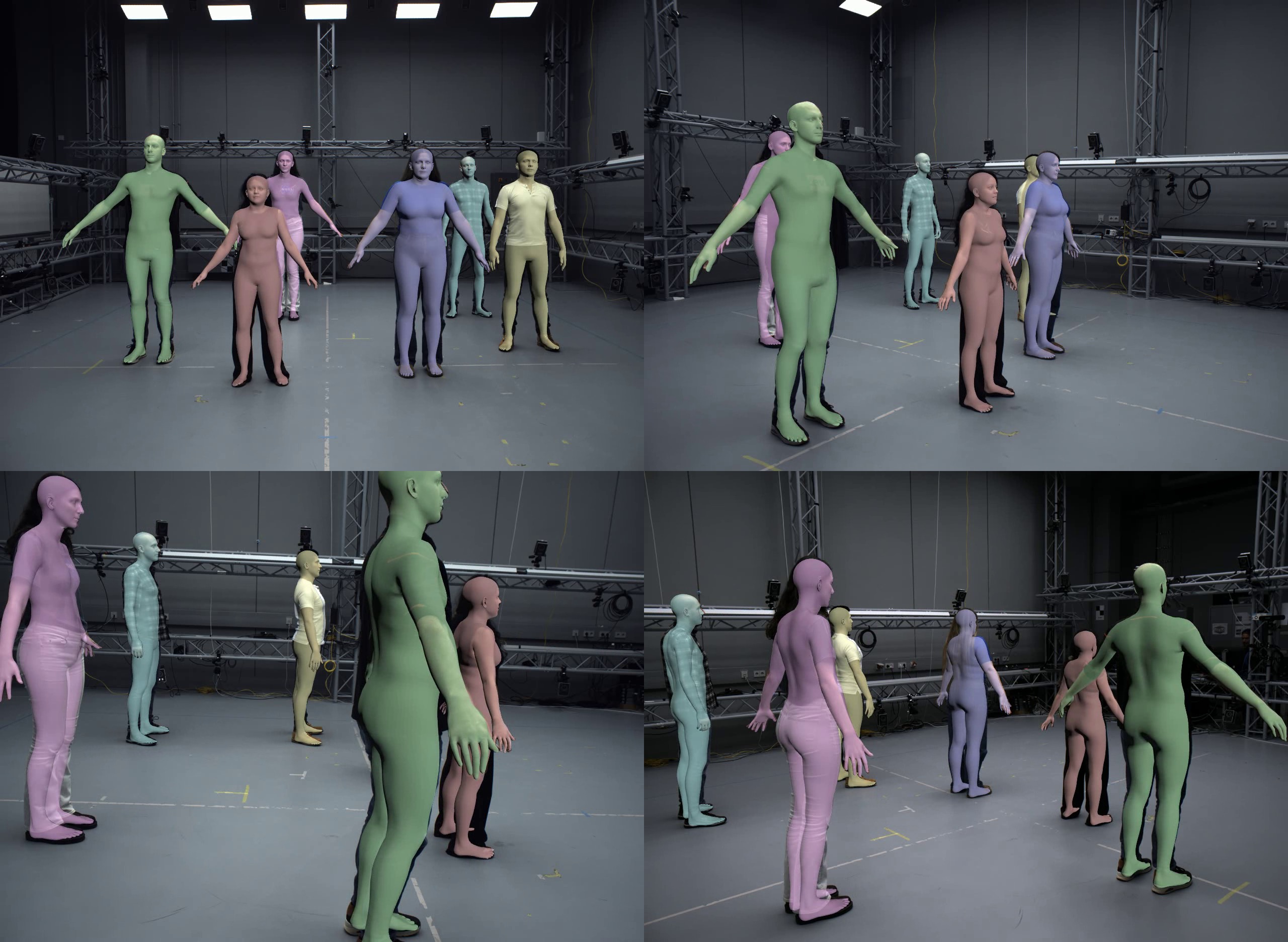}}
\caption{Sample showing the diversity in pose, shape and clothes of the six people captured by \model.}
\label{fig:multiview_six_people}
\end{figure}

We evaluate the performance of our method in scenes involving more than two individuals following the Dance Release configuration. In total, we record 48 sequences: 15, 12, 10, and 11 scenes containing 3, 4, 5, and 6 interacting subjects, respectively; see \cref{fig:multiview_five_people}. The subjects have a diversity in height, body shape, and gender, and wear a wide range of clothing, see \cref{fig:multiview_six_people}.

\\
Among these sequences, 30 exhibit visually accurate results, with physically plausible joint motions throughout the entire sequence. An additional 12 sequences remain largely plausible, with minor artifacts such as occasional joint jitter in a small number of frames. The remaining 6 sequences display unrealistic poses throughout the sequence. These failures are more frequent in scenes with more people, including 3 sequences with 6 subjects and 1 sequence in each of the other groups. We will release the motions we captured.
\\
We observe that failure cases are primarily caused by inaccurate limb predictions, typically due to severe occlusions or errors in the segmentation masks.

\section{Consumer-based Cameras}
\begin{figure}[tbp]
\centerline{\includegraphics[width=1.0\linewidth]{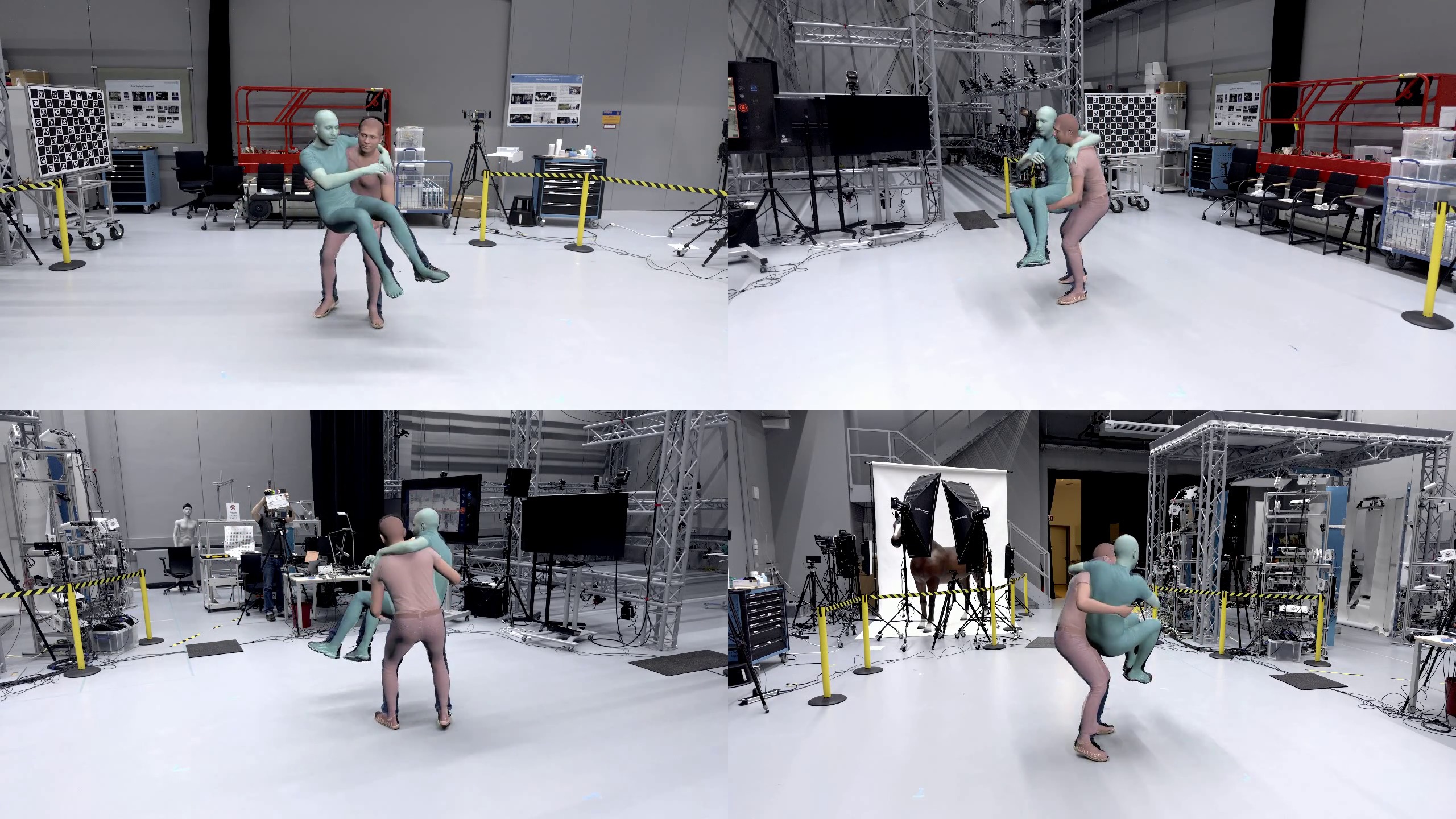}}
\caption{iPhone indoor capture with recovered mesh overlaid on the image.}
\label{fig:iphone_indoor}
\end{figure}

\begin{figure}[tbp]
\centerline{\includegraphics[width=1.0\linewidth]{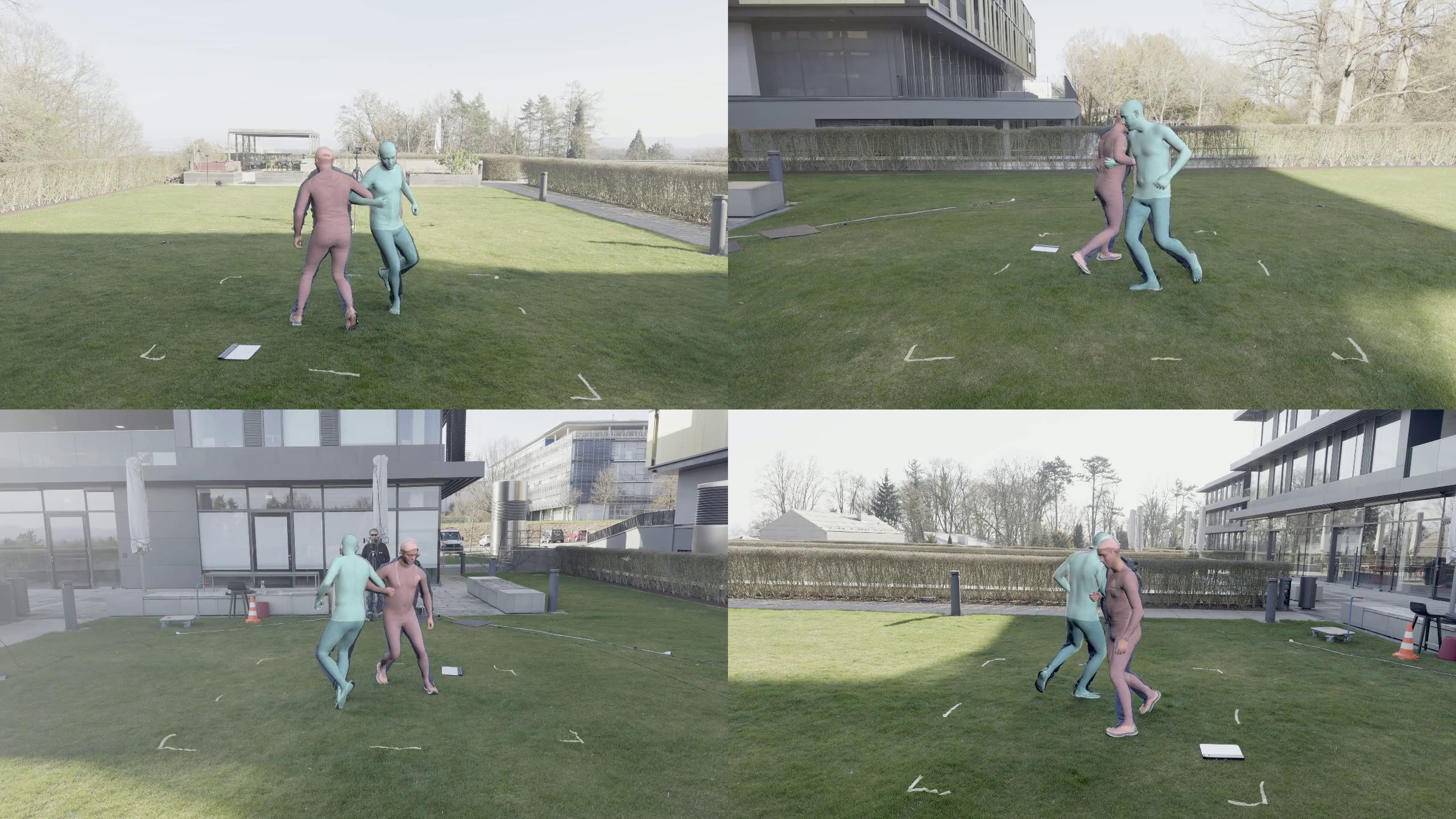}}
\caption{iPhone outdoor capture with recovered mesh overlaid on the image.}
\label{fig:iphone_outdoor}
\end{figure}

We use four iPhone 17 Pro Max devices synchronized via Blackmagic Genlock hardware. Specifically, each device is connected to a Blackmagic Camera Pro Dock to receive LTC timecode and Genlock signals from an external Ambient LockIt generator.  On-device control is handled using the Blackmagic Camera app, enabling one master device to coordinate three slave devices. During recording at 60\,fps, we observe occasional synchronization deviations of up to two frames. While our setup relies on synchronization hardware, it can be replaced by simpler alternatives, such as aligning video streams using audio cues (e.g., a clap at the start of recording). Camera calibration is performed using a ChArUco (chessboard–ArUco) board and the software provided by calib.io. We also tested the OpenCV calibration tool and the calibration difference between both tools is minimal.
This setup demonstrates that a practical, low-cost, and portable capture system can be realized with our technology.
\\
We record 18 indoor (\cref{fig:iphone_indoor}) and 28 outdoor (\cref{fig:iphone_outdoor}) sequences. The indoor subset contains 13 single-person and 8 two-person sequences, while the outdoor subset includes 19 single-person and 9 two-person sequences. The captured actions range from simple walking and running to object interactions, as well as light close-range interactions between subjects.
All recorded sequences exhibit plausible human poses and temporally smooth motions. However, 5 sequences show occasional joint flickering in a small number of frames. Consistent with the previous section, these artifacts are primarily caused by inaccurate limb predictions under occlusion. We will release these motions as well.

\section{Limitations and Future Work}

Our system recovers motion sequences of subjects interacting with each other from multiview cameras.
Despite its competitive performance, \model still has limitations.
The predicted landmark uncertainties and visibility probabilities allow the optimizer to reduce the influence of low-confidence or occluded landmarks.
As a result, the optimizer may ignore landmarks that are heavily occluded and uncertain across views, leading to flickering artifacts or over-smoothing, especially when smoothing regularization is heavily weighted.

The contact probability prediction of our network is conservative, the highest probability is around $60\%$. This is due to the single view ambiguity. If two people are interacting, and one occludes the other person, there is a chance that they are really close but not in contact. One fix to make the network more confident is weighting the loss for parts that are in contact, however, this comes at the cost of failing in the previous example.

Foot contact prediction is generally good. However, our network often predicts the foot landmarks based on the height of the shoes. If we then penalize floor contact errors too strongly during optimization, the optimizer may incorrectly pull the person downward to satisfy the contact constraint. In other situations, when the body is already touching the floor, the dense landmarks alone provide sufficient cues, so the additional contact term has little effect. Therefore, we recommend using the floor contact signal primarily for single-view cases or as a post-processing step to correct foot motion near the floor.

The accuracy of hand-motion recovery can still be improved.
It is worth noting that most marker-based captures ignore the hands completely since they are too costly to capture and clean.
We believe that image-based methods can be further improved with better training data.

We plan to extend the network to predict landmarks across multiple views jointly to improve inter-view consistency.
Similarly, temporal modeling would enhance landmark stability over time.
Another potential direction is the integration of diffusion-based human motion priors to refine predictions.

\section{\supvid Reveal}
\label{sec:reveal}
The videos on the left are the ground truth and \model predictions are on the right.

\end{document}